\documentclass[acmtog]{acmart}
\acmSubmissionID{213}

\usepackage{booktabs} %

\definecolor{ForestGreen}{rgb}{0.13, 0.55, 0.13}
\usepackage{pifont}%
\newcommand{\cmark}{{\color{ForestGreen} \ding{51}}}%
\newcommand{\xmark}{{\color{red} \ding{55}}}%

\newcommand{\rom}[1]{\uppercase\expandafter{\romannumeral #1\relax}}

\usepackage{calc}
\newcommand{\rotatedCentering}[3]{\rotatebox{#1}{\hspace{0.2cm}#3}}

\usepackage{colortbl}
\usepackage{tabulary}
\usepackage{tabularx}
\usepackage{etoolbox}
\usepackage{multirow}
\usepackage{cuted}
\usepackage[percentage]{overpic} 
\usepackage{pgfplots}
\pgfplotsset{compat=newest}%

\definecolor{mygray}{HTML}{EFEFEF}

\usepackage[ruled]{algorithm2e} %

\SetAlFnt{\small}
\SetAlCapFnt{\small}
\SetAlCapNameFnt{\small}
\SetAlCapHSkip{0pt}

\setcopyright{acmlicensed}
\acmJournal{TOG}
\acmYear{2023} \acmVolume{42} \acmNumber{4} \acmArticle{} \acmMonth{8} \acmPrice{15.00}\acmDOI{10.1145/3592107}

\begin{document}
\title{Unsupervised Learning of Robust Spectral Shape Matching}

\author{Dongliang Cao}
\email{dcao@uni-bonn.de}
\orcid{0000-0002-6505-6465}
\author{Paul Roetzer}
\email{paul.roetzer@uni-bonn.de}
\orcid{0009-0005-6698-6663}
\author{Florian Bernard}
\email{fb@uni-bonn.de}
\orcid{0009-0008-1137-0003}
\affiliation{%
  \institution{University of Bonn}
  \streetaddress{Friedrich-Hirzebruch-Allee 5}
  \city{Bonn}
  \state{North Rhine-Westphalia}
  \country{Germany}
  \postcode{D-53115}
}
\setkeys{Gin}{keepaspectratio}

\def\pathOurs{figures/ours/}
\def\pathDiscOp{figures/dscrtopt/}
\def\pathAttFMap{figures/afmaps/}
\def\pathAttFMapFast{figures/afmapsFst/}
\def\pathGeoFMap{figures/gmaps/}
\def\pathDpfm{figures/dpfm/}
\def\pathDpfmUn{figures/dpfmUn/}
\def\srcEnd{_M}
\def\trgtEnd{_N}

\begin{strip}%
\centerline{%
\footnotesize%
\newcommand{\teaserheight}[0]{2.8cm}
\newcommand{\maxteaserwidth}[0]{2.5cm}
\newcommand{\morphbetweenheight}[0]{2cm}
\newcommand{\distbetweenmorph}{-0.4cm}
\begin{tabular}{c}%
    \setlength{\tabcolsep}{1pt} 
    \begin{tabular}{ccccc}%
    \begin{tabular}{cc}
         \includegraphics[width=1.8cm,height=\teaserheight]{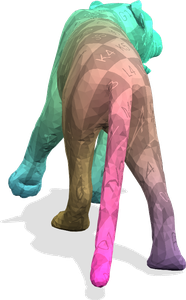}
         \includegraphics[width=1.8cm,height=\teaserheight]{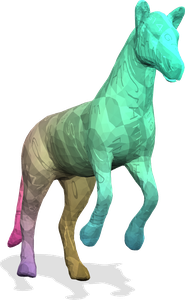}\\
         \includegraphics[width=2.2cm,height=\teaserheight]{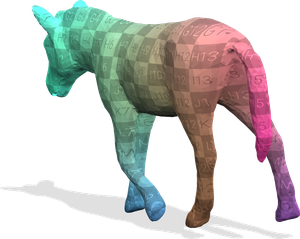}
         \includegraphics[width=1cm,height=\teaserheight]{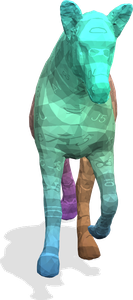}
    \end{tabular} &

    \begin{tabular}{cc}
        \includegraphics[width=\maxteaserwidth,height=\teaserheight]{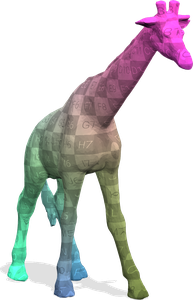} &
        \hspace{-0.4cm}
        \includegraphics[width=1.4cm,height=\teaserheight]{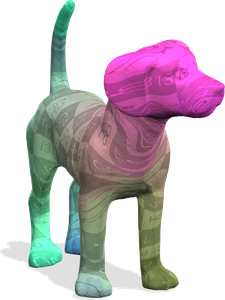}\\
        \includegraphics[width=\maxteaserwidth,height=\teaserheight]{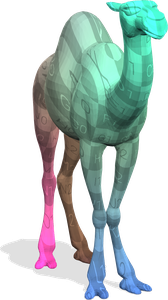} &
        \hspace{-0.4cm}
        \includegraphics[width=1.4cm,height=\teaserheight]{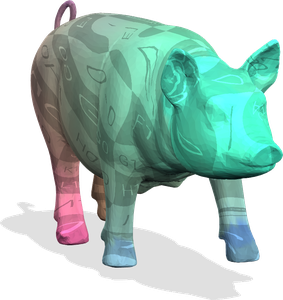}\\
    \end{tabular}&
    
    \begin{tabular}{cc}
        \includegraphics[width=\maxteaserwidth,height=\teaserheight]{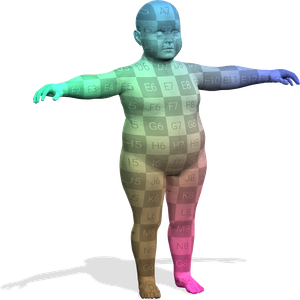} &
        \hspace{-0.4cm}
        \begin{overpic}[width=\maxteaserwidth,height=\teaserheight]{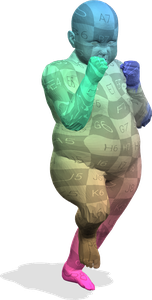}
        \put(12,50){\includegraphics[height=0.8cm]{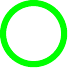}}
        \end{overpic} \\ 
        \includegraphics[width=\maxteaserwidth,height=\teaserheight]{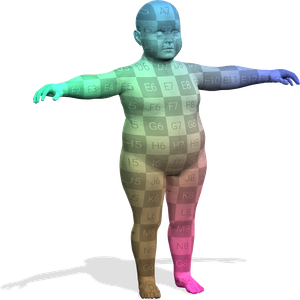} &
        \hspace{-0.4cm}
        \begin{overpic}[width=\maxteaserwidth,height=\teaserheight]{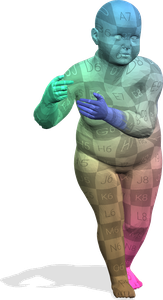}
        \put(30,50){\includegraphics[height=0.8cm]{figures/overlays/green_circle.pdf}}
        \end{overpic}
    \end{tabular}&
    
    \hspace{0.4cm}
    \begin{tabular}{cc}
        \begin{overpic}[width=\maxteaserwidth,height=\teaserheight]{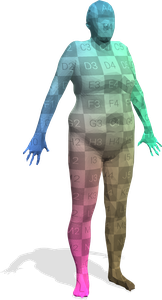}
        \end{overpic}&
         \hspace{-0.3cm}
         \begin{overpic}[width=\maxteaserwidth,height=\teaserheight]{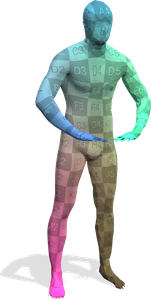}
         \put(-5,10){\includegraphics[height=1.8cm]{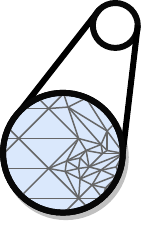}}
         \end{overpic}\\
         \begin{overpic}[width=\maxteaserwidth,height=\teaserheight]{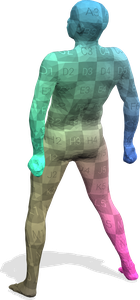}%
        \end{overpic}&
         \hspace{-0.3cm}
         \begin{overpic}[width=1.5cm,height=\teaserheight]{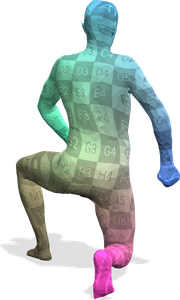}%
         \end{overpic}
    \end{tabular} &
    
    \begin{tabular}{cc}
         \includegraphics[width=2cm,height=\teaserheight]{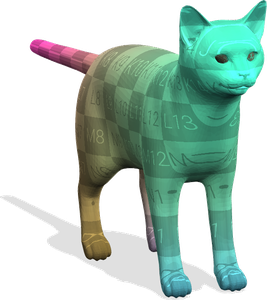}&
         \hspace{-0.3cm}
         \includegraphics[width=1.8cm,height=\teaserheight]{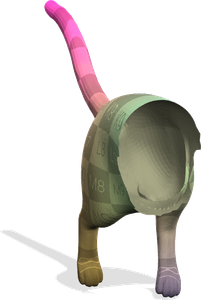}\\
         \includegraphics[width=2cm,height=\teaserheight]{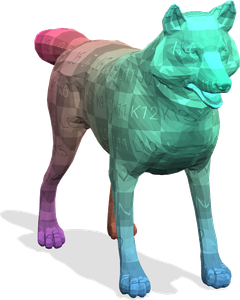}&
         \includegraphics[width=2cm,height=\teaserheight]{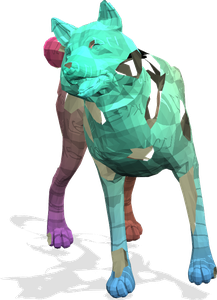}
    \end{tabular}\\
    
    Alignment Invariant&
    Non-Isometry&
    Topological Noise&
    Anisotropic Meshing&
    Partial Shape
    \end{tabular}%
    
\end{tabular}
}
\captionof{figure}{We propose the first \textbf{unsupervised spectral shape matching} approach that is \textbf{robust across a broad range of challenging settings}:  shape matching without initial alignment due to the intrinsic formulation, shape matching of non-isometric shape pairs, shape matching with topological noise, shape matching with anisotropic meshing, and partial shape matching.}
\label{fig:teaser}
\end{strip}

\begin{abstract}
We propose a novel learning-based approach for robust 3D shape matching. Our method builds upon deep functional maps and can be trained in a fully unsupervised manner. Previous deep functional map methods mainly focus on predicting optimised functional maps alone, and then rely on off-the-shelf post-processing to obtain accurate point-wise maps during inference. However, this two-stage procedure for obtaining point-wise maps often yields sub-optimal performance. In contrast, building upon recent insights about the relation between functional maps and point-wise maps, we propose a novel unsupervised loss to couple the functional maps and point-wise maps, and thereby directly obtain point-wise maps without any post-processing. Our approach obtains accurate correspondences not only for near-isometric shapes, but also for more challenging non-isometric shapes and partial shapes, as well as shapes with different discretisation or topological noise. Using a total of nine diverse datasets, we extensively evaluate the performance and demonstrate that our method substantially outperforms previous state-of-the-art methods, even compared to recent supervised methods. {Our code is available at \url{https://github.com/dongliangcao/Unsupervised-Learning-of-Robust-Spectral-Shape-Matching}}.
\end{abstract}



\maketitle

\section{Introduction}
\label{sec:intro}
Shape matching is a long-standing problem in computer graphics and shape analysis with a diverse range of applications, including texture transfer~\cite{dinh2005texture}, deformation transfer~\cite{sumner2004deformation}, and parametric model construction~\cite{loper2015smpl,egger20203d}. Even though shape matching has been studied for decades~\cite{van2011survey,tam2012registration}, finding correspondences between non-rigidly deformed shapes still remains highly challenging, especially in the presence of large non-isometric deformation, partiality, or topological inconsistency. 

\begin{table}[tbh!]
\centering
\small
\caption{\textbf{Method comparison.} Our method is the first unsupervised learning approach that operates in the intrinsic/spectral domain and achieves accurate matching performance for both non-isometric and partial shapes.}
\label{tab:comparison}
\begin{tabular}{@{}lcccc@{}}
\toprule
Methods & Unsup. & Fully intrinsic & Non-iso. & Part. \\ \midrule
\cite{litany2017deep} & \xmark & \cmark & \xmark  & \xmark \\
\cite{groueix20183d} & \xmark & \xmark & \xmark  & \xmark \\
\cite{halimi2019unsupervised} & \cmark & \cmark & \xmark  & \xmark\\
\cite{roufosse2019unsupervised} & \cmark & \cmark & \xmark  & \xmark\\
\cite{wiersma2020cnns} & \xmark & \xmark & \xmark  & \xmark \\
\cite{li2020shape} & \xmark & \xmark & \xmark  & \xmark \\
\cite{donati2020deep} & \xmark & \cmark & \xmark  & \xmark \\
\cite{eisenberger2020deep} & \cmark & \xmark & \cmark & \xmark \\
\cite{attaiki2021dpfm} & \xmark & \cmark & \xmark & \cmark \\
\cite{eisenberger2021neuromorph} & \cmark & \xmark & \cmark & \xmark \\
\cite{trappolini2021shape} & \xmark & \xmark & \xmark & \xmark \\
\cite{cao2022unsupervised} & \cmark & \cmark & \xmark & \cmark \\
\cite{donati2022deep} & \cmark & \cmark & \cmark & \xmark \\
\cite{li2022learning} & \cmark & \cmark & \cmark & \xmark \\
Ours & \cmark & \cmark & \cmark & \cmark \\ \bottomrule
\end{tabular}
\end{table}

With recent developments in deep learning, numerous learning-based approaches were proposed for non-rigid 3D shape matching, including both supervised~\cite{litany2017deep,groueix20183d,wiersma2020cnns,donati2020deep,trappolini2021shape} and unsupervised methods~\cite{halimi2019unsupervised,roufosse2019unsupervised,eisenberger2020deep,sharma2020weakly}. While most of them mainly focus on near-isometric shape matching (such as the recent unsupervised work by \citet{eisenberger2020deep} which demonstrated near-perfect performance), only few works explicitly consider more challenging scenarios, such as non-isometric shape matching~\cite{donati2022deep,li2022learning} and partial shape matching~\cite{attaiki2021dpfm,cao2022unsupervised}. For such settings,  the performance gap between the state-of-the-art supervised methods~\cite{donati2020deep,attaiki2021dpfm} and unsupervised methods~\cite{li2022learning,cao2022unsupervised} is still large, while supervised methods at the same time suffer from overfitting.

In this work we propose an unsupervised learning
framework to address these shortcomings. Our universal  framework can be applied to a broad range of shape matching settings, including near-isometric shapes and more challenging non-isometric shapes, as well as partial shapes. 
Our method builds upon the powerful functional map framework~\cite{ovsjanikov2012functional}  integrated into a deep neural network~\cite{litany2017deep}.
While previous unsupervised deep functional map approaches (e.g.~\cite{halimi2019unsupervised,roufosse2019unsupervised})
only regularise the functional map alone and ignore relations between the functional map and the underlying point-wise maps, we propose to explicitly exploit this relation during neural network training. Specifically, inspired by the recent axiomatic  (i.e.~non-learning-based) method by~\citet{ren2021discrete}, we introduce a novel differentiable unsupervised coupling loss, which accounts for the functional map being associated to a point-wise map. 
Furthermore, we propose a test-time adaptation strategy to simultaneously optimise the functional map and the corresponding point-wise map to further improve the matching performance. Overall, we demonstrate that this substantially improves upon previous shape matching approaches on a diverse and extensive selection of datasets. We summarise our main contributions as follows:
\begin{itemize}
\item For the first time we propose a \textit{universal unsupervised learning} framework that allows to obtain accurate correspondences not only for near-isometric shapes but also for more challenging non-isometric and partial shapes, see Tab.~\ref{tab:comparison}.
\item We introduce a \emph{novel unsupervised loss} to enforce the functional map to be associated with a point-wise map. Together with our \textit{test-time adaptation}, our method is much more robust to the choice of the spectral resolution compared to existing deep functional map methods.
\item We demonstrate \textit{state-of-the-art} performance on numerous challenging benchmarks in diverse settings, including near-isometric, non-isometric and partial shape matching, even in comparison to recent supervised methods.
\end{itemize}

\section{Related Work}
\label{sec:related_work}
In the following we will focus on reviewing those methods that are most relevant to our work. A more comprehensive overview can be found in~\cite{van2011survey,tam2012registration,sahilliouglu2020recent}.

\subsection{Axiomatic Functional Map Methods}
Several shape matching approaches~\cite{huang2008non,ovsjanikov2010one,windheuser2011geometrically,holzschuh2020simulated,roetzer2022scalable} directly establish correspondences between a given pair of shapes, while some other methods~\cite{ezuz2019elastic,eisenberger2019divergence,bernard2020mina} attempt to solve the problem by finding a non-rigid deformation to align them. Nevertheless, they solve the correspondence problem by minimising an energy function defined on the shape surface and thus often result in complex optimisation problems. In contrast, the functional map framework encodes the correspondence relationship into a small matrix (i.e. the functional map) that can be efficiently computed~\cite{ovsjanikov2012functional}. Due to its simplicity and efficiency, the functional map framework has been extended in numerous  works, e.g., in terms of improving the matching accuracy and robustness~\cite{eynard2016coupled,ren2019structured}, or extending it to non-isometric shape matching~\cite{kovnatsky2013coupled,ren2018continuous,eisenberger2020smooth,ren2021discrete,magnet2022smooth}, partial shape matching~\cite{rodola2017partial,litany2017fully}, multi-shape matching~\cite{huang2014functional,cohen2020robust,huang2020consistent,gao2021isometric}, {and matching with multiple solutions~\cite{ren2020maptree}}. Nevertheless, axiomatic functional map methods heavily rely on handcrafted features~\cite{bronstein2010scale,aubry2011wave,salti2014shot}, which limits their  matching performance, especially in the presence of non-isometries and partiality. In contrast, our method directly learns robust features from  training data and achieves more accurate and robust matching performance on challenging non-isometric and partial settings.

\subsection{Deep Functional Map Methods}
Unlike axiomatic approaches that use handcrafted features, deep functional map methods attempt to learn features directly from  training data. The supervised FMNet~\cite{litany2017deep} was first proposed to learn a non-linear transformation of SHOT descriptors~\cite{salti2014shot}. Later works~\cite{halimi2019unsupervised,roufosse2019unsupervised} enable unsupervised training of FMNet by introducing isometry losses in the spatial and spectral domain, respectively. To improve the matching performance, several works~\cite{donati2020deep,sharma2020weakly} replace FMNet by point-based networks~\cite{qi2017pointnet++,thomas2019kpconv}. {In order to extend the learning framework for point cloud matching, DiffFMaps~\cite{marin2020correspondence} learns both the basis functions and features together. A similar idea was used in~\citet{azencot2021data}.} Recently, DiffusionNet~\cite{sharp2020diffusionnet} introduces an implicit diffusion process and has led to state-of-the-art matching performance~\cite{attaiki2021dpfm,cao2022unsupervised,donati2022deep,li2022learning,attaiki2023ncp}. DPFM~\cite{attaiki2021dpfm} enables supervised partial shape matching building upon a cross-attention mechanism and outlier detection. ConsistFMaps~\cite{cao2022unsupervised} introduces a virtual universe shape to enable cycle-consistent multi-shape matching. DUO-FMNet~\cite{donati2022deep} learns orientation-aware features by considering complex functional maps~\cite{donati2022complex}. AttentiveFMaps~\cite{li2022learning} introduces a spectral attention mechanism to combine functional maps of different spectral resolutions. Despite the rapid progress of deep functional map methods, existing unsupervised approaches mostly focus on near-isometric shape matching, or they suffer from large performance gaps compared to supervised methods in the context of non-isometric or partial shape matching. In this work, we close this gap by introducing a universal unsupervised learning framework that achieves state-of-the-art performance in diverse settings, including near-isometric, non-isometric and partial shape matching.

\subsection{Functional Map Refinement}
A common strategy to improve the final matching performance is to iteratively refine functional maps as a post-processing step. As the simplest refinement technique, ICP iteratively converts the functional map to a point-wise map via nearest neighbour search~\cite{ovsjanikov2012functional}. Follow-up works~\cite{rodola2015point,rodola2017regularized,ezuz2017deblurring} iteratively optimise the functional map and point-wise map by minimising an energy function. PMF~\cite{vestner2017product} introduces bijectivity as a hard constraint for point-wise map conversion. ZoomOut~\cite{melzi2019zoomout} increases the spectral resolution of the functional map during the map conversion. RHM~\cite{ezuz2019reversible} prompts reversible harmonic maps, thereby resulting in maps with lower conformal distortion. {FSF~\cite{pai2021fast} theoretically analyses the relationship between functional maps and point-wise maps and proposes a point-wise map recovery method based on spectral alignment.} Another line of works encourages cycle-consistency~\cite{wang2013image,huang2014functional,bernard2019hippi,huang2020consistent} in the case of processing a collection of shapes. Nevertheless,  two-stage `match+refine' approaches  often yield sub-optimal performance, especially given erroneous initial functional maps. In contrast, our approach directly enforces the functional map to be associated with a point-wise map and optimises both of them simultaneously.  

\section{Background}
\label{sec:background}
In this section we explain the background and introduce the notation 
used throughout the rest of the paper, see Tab.~\ref{table:notation}.

\begin{table}[h!]
\small\centering
    \caption{Summary of the \textbf{notation} used in this paper.}%
    \label{table:notation}%
    \begin{tabularx}{\columnwidth}{lp{5.6cm}}
        \toprule
        \textbf{Symbol} &\textbf{Description} \\
        \toprule
        $\mathcal{M}, \mathcal{N}$ &3D shapes (triangle mesh) with $n_{\mathcal{M}}$ ($n_{\mathcal{N}}$) vertices\\
        $X_{\mathcal{M}}$ & vertex positions of shape $\mathcal{M}$\\
        $F_{\mathcal{M}}$ & vertex-wise features of shape $\mathcal{M}$\\
        $L_{\mathcal{M}}$ &Laplacian matrix associated with shape $\mathcal{M}$\\
        $\Phi_{\mathcal{M}}$ &eigenfunctions of Laplacian matrix $L_{\mathcal{M}}$\\
        $C_{\mathcal{MN}}$ &functional map between shapes $\mathcal{M}$ and $\mathcal{N}$\\
        $\Pi_{\mathcal{MN}}$ & point-wise map between shapes $\mathcal{M}$ and $\mathcal{N}$\\
         $E(\cdot)$ & energy of the functional map solver\\
         $L(\cdot)$ & loss function used during neural network training\\
        \bottomrule
    \end{tabularx}
\end{table}

\subsection{Functional Map Pipeline}
\label{subsec:fmap_pipeline}
Consider a pair of 3D shapes $\mathcal{M}$ and $\mathcal{N}$ represented as triangle meshes, with $n_{\mathcal{M}}$ and $n_{\mathcal{N}}$ vertices, respectively. The associated positive semi-definite Laplacian matrices $L_{M} \in \mathbb{R}^{n_{\mathcal{M}} \times n_{\mathcal{M}}}, L_{\mathcal{N}} \in \mathbb{R}^{n_{\mathcal{N}} \times n_{\mathcal{N}}}$~\cite{pinkall1993computing} are computed as $L_{\mathcal{M}}= A_{\mathcal{M}}^{-1}W_{\mathcal{M}}$, where $A_{\mathcal{M}}$ is the diagonal matrix of lumped area elements and $W_{\mathcal{M}}$ is the cotangent weight matrix. The first $k$ eigenfunctions $\Phi_{\mathcal{M}} \in \mathbb{R}^{n_{\mathcal{M}} \times k}, \Phi_{\mathcal{N}} \in \mathbb{R}^{n_{\mathcal{N}} \times k}$ of the respective Laplacian matrices are used as the spectral embedding of each shape. Given $c$-dimensional features defined on each shape $F_{\mathcal{M}} \in \mathbb{R}^{n_{\mathcal{M}} \times c}, F_{\mathcal{N}} \in \mathbb{R}^{n_{\mathcal{N}} \times c}$, {either computed from handcrafted feature descriptors or extracted from learnable feature extractor}, the functional map $C_{\mathcal{MN}} \in \mathbb{R}^{k \times k}$ associated with the spectral embedding can be computed by solving the  continuous optimisation problem
        \begin{equation}
            \label{eq:fmap}  C_{\mathcal{MN}}=\mathrm{argmin}_{C}~ E_{\mathrm{data}}\left(C\right)+\lambda E_{\mathrm{reg}}\left(C\right).
        \end{equation}
Here, minimising $E_{\mathrm{data}}=\left\|C\Phi_{\mathcal{M}}^{\dagger}F_{\mathcal{M}}-\Phi_{\mathcal{N}}^{\dagger}F_{\mathcal{N}}\right\|^{2}_{F}$ enforces descriptor preservation, while minimising the regularisation term $E_{\mathrm{reg}}$ imposes some form of structural properties (see e.g.~\cite{ovsjanikov2012functional}). The operator $^{\dagger}$ denotes the Moore-Penrose inverse. From the optimal $C_{\mathcal{MN}}$, the point-wise map $\Pi_{\mathcal{NM}} \in \{0,1\}^{n_{\mathcal{N}} \times n_{\mathcal{M}}}$ can be recovered based on the relationship $C_{\mathcal{MN}} = \Phi_{\mathcal{N}}^{\dagger}\Pi_{\mathcal{NM}}\Phi_{\mathcal{M}}$, e.g.~either by nearest neighbour search or by other post-processing techniques~\cite{vestner2017product,melzi2019zoomout,ezuz2019reversible}. 

\subsection{Deep Functional Maps}
\label{subsec:deep_fmap}
\begin{figure}[!h]
    \begin{center}  \includegraphics[width=\columnwidth]{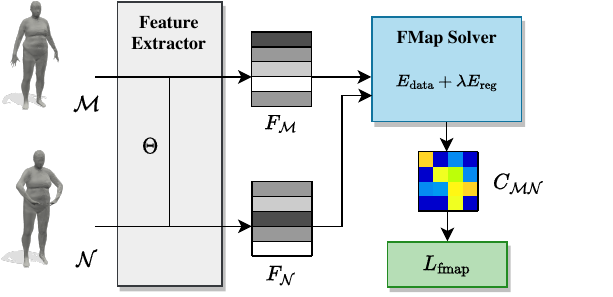}
    \caption{\textbf{Common pipeline of deep functional map methods}. First, the feature extractor computes per-vertex features for each of the two input shapes. Then the functional map solver is used to compute the {(bidirectional)} functional map based on the extracted features. To train the feature extractor, structural regularisation is imposed on the computed functional maps.
    }
    \label{fig:deep_fmap_pipeline}
    \end{center}
\end{figure}

With the recent progress in deep learning, many deep functional map methods~\cite{roufosse2019unsupervised,sharma2020weakly} have been proposed and lead to state-of-the-art matching performance. The common pipeline of those methods is shown in Fig.~\ref{fig:deep_fmap_pipeline}.  

\begin{figure*}[h!t!]
\begin{center}
\centerline{\includegraphics[width=\textwidth]{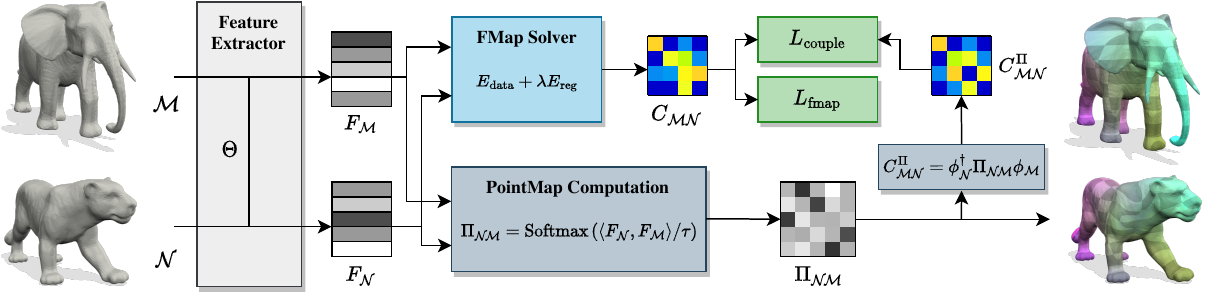}}
\caption{\textbf{Overview of our unsupervised robust spectral shape matching method}. First, the feature extractor with shared weights $\Theta$ takes a pair of shapes $\mathcal{M}$ and $\mathcal{N}$ and extracts vertex-wise features $F_{\mathcal{M}}$ and $F_{\mathcal{N}}$, respectively. Afterwards, the (non-trainable but differentiable) functional map solver is used to compute the functional map $C_{\mathcal{MN}}$ given extracted features. At the same time, the point-wise map $\Pi_{\mathcal{NM}}$ is obtained based on the similarity of the features $F_{\mathcal{M}}$ and $F_{\mathcal{N}}$. Eventually, a coupling loss term is used to regularise the functional map $C_{\mathcal{MN}}$ to be associated to the point-wise map $\Pi_{\mathcal{NM}}$.%
}
\label{fig:framework}
\end{center}
\end{figure*}

\subsubsection{Feature Extraction}
\label{subsubsec:feature_extractor}
\begin{figure}[h]
    \centering
    \def\hspaceColsF{-0.35cm}
\def\heightF{2.3cm}
\def\widthF{2.0cm}
\begin{tabular}{ccccc}%
        \setlength{\tabcolsep}{0pt} 
        $F_\bullet^0$ & $F_\bullet^1$ & $F_\bullet^2$ & $F_\bullet^3$ & $F_\bullet^4$ \\
        \hspace{\hspaceColsF}
        \includegraphics[height=\heightF, width=\widthF]{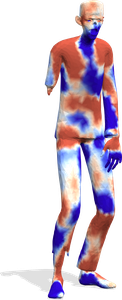}&
        \hspace{\hspaceColsF}
        \includegraphics[height=\heightF, width=\widthF]{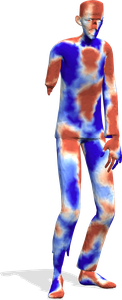}&
        \hspace{\hspaceColsF}
        \includegraphics[height=\heightF, width=\widthF]{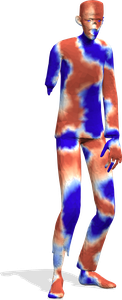}&
        \hspace{\hspaceColsF}
        \includegraphics[height=\heightF, width=\widthF]{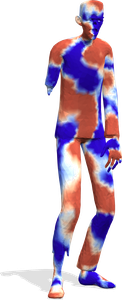}&
        \hspace{\hspaceColsF}
        \includegraphics[height=\heightF, width=\widthF]{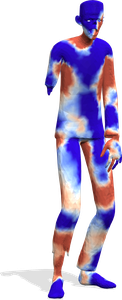}\\
        \hspace{\hspaceColsF}
        \includegraphics[height=\heightF, width=\widthF]{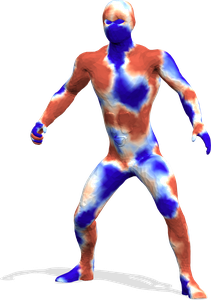}&
        \hspace{\hspaceColsF}
        \includegraphics[height=\heightF, width=\widthF]{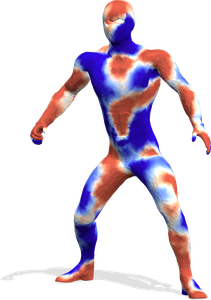}&
        \hspace{\hspaceColsF}
        \includegraphics[height=\heightF, width=\widthF]{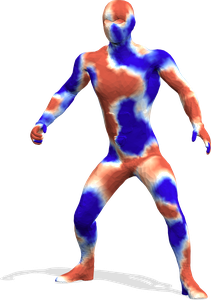}&
        \hspace{\hspaceColsF}
        \includegraphics[height=\heightF, width=\widthF]{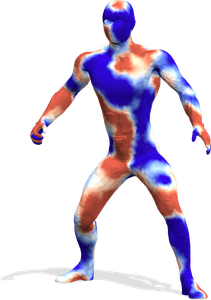}&
        \hspace{\hspaceColsF}
        \includegraphics[height=\heightF, width=\widthF]{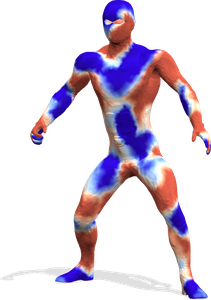}%
\end{tabular}
    \caption{{Visualisation of the first five channels of the extracted features from the \textbf{feature extractor trained 
    by our method} for a shape pair from DT4D-H test dataset. Smaller values are in blue, white represents values in-between and larger values are in red. Values outside of $25\% - 75\%$ quantiles are clipped for improved visuals. Despite  non-isometry and  partiality (upper shape misses one arm) we can see that our network is able to predict consistent features across the shape.}}
    \label{fig:feature-vis}
\end{figure}

{Deep functional map methods use a neural network to learn features directly from training data, rather than using handcrafted features.
Commonly, the feature extractor is used in a Siamese way, i.e. the same network with shared weights processes both shapes $\mathcal{M}$ and $\mathcal{N}$. The extracted features $F_{\mathcal{M}}$ and $F_{\mathcal{N}}$ are then used in a differentiable functional map pipeline, cf. Sec.~\ref{subsec:fmap_pipeline} {and Fig.~\ref{fig:feature-vis}}.
}

\subsubsection{Structural Regularisation}
\label{subsubsec:structural_reg}
{To train the feature extractor in an unsupervised way, a common strategy is to impose structural regularisation on the computed functional maps. Following~\citet{roufosse2019unsupervised},  structural functional map regularisation can be expressed in the form
    \begin{equation}
        \label{eq:fmaps}
        L_{\mathrm{fmap}} = \lambda_{\mathrm{bij}}L_{\mathrm{bij}} + \lambda_{\mathrm{orth}}L_{\mathrm{orth}}.
    \end{equation}
The bijectivity regularisation $L_{\mathrm{bij}}$ enforces the map from $\mathcal{M}$ through $\mathcal{N}$ back to $\mathcal{M}$ to be the identity map (and vice versa), i.e.
    \begin{equation}
        \label{eq:bij}
        L_{\mathrm{bij}}=\left\|C_{\mathcal{MN}}C_{\mathcal{NM}}-I\right\|^{2}_{F}+\left\|C_{\mathcal{NM}}C_{\mathcal{MN}}-I\right\|^{2}_{F}.
    \end{equation}
The orthogonality regularisation $L_{\mathrm{orth}}$ prompts locally area-preserving matchings for both matching directions, i.e.
    \begin{equation}
        \label{eq:orth}
        L_{\mathrm{orth}}=\left\|C_{\mathcal{MN}}^{\top}C_{\mathcal{MN}}-I\right\|^{2}_{F}+\left\|C_{\mathcal{NM}}^{\top}C_{\mathcal{NM}}-I\right\|^{2}_{F}.
    \end{equation} 
In the context of partial shape matching, the functional map $C_{\mathcal{MN}}$ from complete shape $\mathcal{M}$ to partial shape $\mathcal{N}$ becomes a slanted diagonal matrix~\cite{rodola2017partial}. As a result, the terms $L_{\mathrm{bij}}$ and $L_{\mathrm{orth}}$ can be modified based on this property~\cite{attaiki2021dpfm}.}

Most existing deep functional map  methods follow the above formulation and only consider functional maps while \emph{ignoring the relation between the functional map and the point-wise map}. In contrast, our approach explicitly considers this relation, which makes it possible to abandon the use of additional post-processing techniques to obtain point-wise maps during inference.

\section{Unsupervised Robust Spectral Shape Matching}
\label{sec:method}

Our work aims to robustly estimate functional maps and point-wise maps for shape pairs in diverse settings, including near-isometric and non-isometric deformations, as well as shapes with partiality. {The key insight of our method is based on the observation that existing learning-based methods solely focus on optimising functional maps and rely on some non-learnable post-processing techniques to obtain the point maps. However, the two-stage procedure often yields sub-optimal matching results, especially under challenging scenarios. In order to directly obtain point maps without post-processing, we  explicitly consider the map relationship during training. To this end, we introduce a novel coupling loss to enable the optimisation of the functional map and the point-wise map simultaneously (unlike existing axiomatic methods, which often update them alternatingly).} The whole framework of our approach is depicted in Fig.~\ref{fig:framework}. Our framework has four main components: feature extraction (Sec.~\ref{subsubsec:feature_extractor}), functional map solver (Sec.~\ref{subsec:fmap_pipeline}), differentiable point-wise map computation (Sec.~\ref{subsec:pmap_solver}) and unsupervised loss (Sec.~\ref{subsec:unsup_loss}).

\subsection{Differentiable Point-Wise Map Computation}
\label{subsec:pmap_solver}
{As illustrated in Sec.~\ref{subsec:deep_fmap}, most existing deep functional map methods use extracted features to compute functional maps and ignore the associated point-wise map. As a consequence, the existing functional map methods rely on off-the-shelf post-processing techniques during inference to obtain the final point-wise maps. When doing so, the relation between the functional and point-wise maps is underexplored, which in turn hampers the final matching quality as we demonstrate in Sec.~\ref{sec:ablation}. While recent work by \citet{cao2022unsupervised} introduced a universe classifier to compute \emph{shape-to-universe} point-wise maps, it requires the knowledge of the number of universe vertices, which is typically unknown in practice. In contrast, our method directly computes the \emph{pairwise} point-wise map based on  the similarity measurement between features $F_{\mathcal{N}}$ and $F_{\mathcal{M}}$ defined on each shape.} In theory, the point-wise map $\Pi_{\mathcal{NM}}$ should be a (partial) permutation matrix, i.e.
    \begin{equation}
        \label{eq:permutation_mat}
        \left\{\Pi \in\{0,1\}^{n_{\mathcal{N}} \times n_{\mathcal{M}}}: \Pi \mathbf{1}_{n_{\mathcal{M}}} = \mathbf{1}_{n_{\mathcal{N}}}, \mathbf{1}_{n_{\mathcal{N}}}^{\top} \Pi \leq \mathbf{1}_{n_{\mathcal{M}}}^{\top}\right\},
    \end{equation}
where the element at position $(i,j)$ indicates whether the $i$-th point in $\mathcal{N}$ corresponds to the $j$-th point in $\mathcal{M}$. In practice, we use the softmax operator to produce a soft correspondence matrix to make the computation differentiable, i.e.
    \begin{equation}
        \label{eq:soft_corr}
        \Pi_{\mathcal{NM}} = \mathrm{Softmax}\left( {F_{\mathcal{N}}F_{\mathcal{M}}^{T}} / \tau\right),
    \end{equation}
where $\tau$ is the scaling factor to determine the softness of the correspondence matrix. {The softmax operator is applied in each row to ensure non-negativity and $\Pi_{\mathcal{NM}} \mathbf{1}_{n_{\mathcal{M}}} = \mathbf{1}_{n_{\mathcal{N}}}$.} Moreover, we convert the point-wise map to a functional map (via $\phi_{\mathcal{N}}^{\dagger}\Pi_{\mathcal{NM}}\phi_{\mathcal{M}}$) for our unsupervised loss computation and test-time adaptation, which we introduce in the following.

\subsection{Unsupervised Loss}
\label{subsec:unsup_loss}
Our unsupervised loss can be divided into two parts, structural regularisation (see Sec.~\ref{subsubsec:structural_reg}) and a coupling between functional and point-wise maps.

\subsubsection{Coupling Functional and Point-Wise Maps}
\label{subsubsec:coupling}
\begin{figure}[!h]
\centering 
\begin{tabular}{cc}
    \includegraphics[trim={0 0 4.4cm 0}, clip, height=2cm]{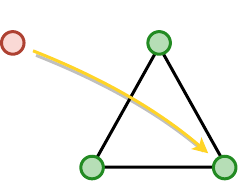}&  
    \includegraphics[trim={4.5cm 0 0 0}, clip, height=2cm]{figures/ours_vs_discrt_opt.pdf}\\
    \hspace{0.7cm}\citet{ren2021discrete} &
    \hspace{0.8cm}Ours%
\end{tabular}
\caption{\textbf{Hard vs.~soft correspondences {in a conceptual 2D case}.} The method proposed by \citet{ren2021discrete} only allows for discrete matchings, i.e.~the red vertex can only be matched to one of the green vertices of the triangle (left). In contrast, our method allows for a {smooth matching}, i.e. the red vertex can also be matched to points in the interior {as a convex combination of the three vertices} (right).
}
\label{fig:discOpt-vs-ours}
\end{figure}

The structural regularisation terms are not sufficient to obtain accurate matching results for shapes undergoing non-isometric deformations. The recent axiomatic approach by \citet{ren2021discrete} optimises for functional maps in the (discrete) constraint set
    \begin{equation}
        \left\{C_{\mathcal{MN}} \;|\; \exists \; \Pi_{\mathcal{NM}}, \;\mathrm{s.t.}\; C_{\mathcal{MN}} = \phi_{\mathcal{N}}^{\dagger}\Pi_{\mathcal{NM}}\phi_{\mathcal{M}} \right\},
    \end{equation}
which leads to more robust and accurate matchings. 
Inspired by this insight, we introduce our coupling loss term to ensure that the functional map is associated with a valid point-wise map, i.e. 
    \begin{equation}
        \label{eq:couple}
        L_{\mathrm{couple}} = \left\|C_{\mathcal{MN}} - \phi_{\mathcal{N}}^{\dagger}\Pi_{\mathcal{NM}}\phi_{\mathcal{M}}\right\|^{2}_{F}.
    \end{equation}
{Unlike the axiomatic approach by \citet{ren2021discrete}, which formulates the relation between functional and point-wise maps as a discrete constraint, our approach allows to incorporate a soft correspondence matrix as the point-wise map. To this end, the correspondence can be interpreted as a {probabilistic  matching over all vertices} rather than the severely restricted case of vertex-to-vertex matching, see Fig.~\ref{fig:discOpt-vs-ours} {for a conceptual 2D illustration}. In practice, this is important for shape pairs with inconsistent discretisation, as we experimentally demonstrate in Sec.~\ref{subsec:ablation}. In contrast to existing deep functional map methods, our approach  uses the extracted features to compute both, functional maps and point-wise maps,  in a differentiable way during training, and thus explicitly models their relationship and allows for their simultaneous optimisation.}

\subsubsection{Total Unsupervised Loss.}
The overall unsupervised loss for training is a weighted sum of the individual losses, i.e. 
    \begin{equation}
        \label{eq:total_loss}
        L_{\mathrm{total}} = L_{\mathrm{fmap}} + \lambda_{\mathrm{couple}}L_{\mathrm{couple}}.
    \end{equation}

\subsection{Inference and Test-Time Adaptation}
\label{subsec:test-time-adaptation}
In the following we explain how we obtain our final point-wise maps during inference. 

\subsubsection{Test-Time Adaptation.}
To improve the final matching performance, we propose a test-time adaptation strategy for each test shape pair individually. {In contrast to post-processing techniques required by existing works, our test-time adaptation can be considered as a differentiable refinement process that directly optimises the feature extractor and seamlessly integrates into the learning paradigm.} Specifically, we compute $L_{\mathrm{total}}$ defined in Eq.~\ref{eq:total_loss} and update the feature extractor in an iterative way during inference. {The effect of this is that  both the functional map and the point-wise map are updated simultaneously, since both of them are computed based on the extracted features. In contrast, most existing post-processing techniques optimise them either separately~\cite{melzi2019zoomout}, or require an additional coupling term~\cite{ren2018continuous}.} 

To address \emph{non-isometric} matching, we introduce an additional smoothness penalty term for the point-wise map, which is  based on the Dirichlet energy~\cite{ezuz2019reversible,magnet2022smooth}, i.e.
    \begin{equation}
        \label{eq:dirichlet}
        L_{\mathrm{dirichlet}} = \left\|\Pi_{\mathcal{NM}}X_{\mathcal{M}}\right\|^{2}_{L_{\mathcal{N}}},
    \end{equation}
where $\left\|X\right\|^{2}_{L} := \mathrm{Trace}\left(X^{\top}LX\right)$. The smoothness penalty term encourages neighbouring vertices on shape $\mathcal{M}$ to be matched to neighbouring vertices on shape $\mathcal{N}$. {Imposing this smoothness penalty term is crucial for non-isometric shape matching, since a smooth point-wise map is a desirable property for non-isometric matching. To avoid degenerate matchings (such as all vertices from shape $\mathcal{M}$ being matched to the same vertex of shape $\mathcal{N}$, for which the Dirichlet energy becomes zero), we only apply this smoothness term during test-time adaptation.

\subsubsection{Point-Wise Maps at Inference Time}
Existing deep functional map methods mainly focus on computing accurate functional maps and either use nearest neighbour search or other post-processing techniques~\cite{vestner2017product,melzi2019zoomout,eisenberger2020smooth} to obtain the final point-wise maps. In contrast, our method explicitly models the relation between the functional map and the point-wise map, and is thereby capable of directly obtaining the point-wise map without a functional map solver or other post-processing. During inference, our method directly obtains the point-wise map based on feature similarities using
    \begin{equation}
        \Pi_{\mathcal{NM}} = \mathrm{NN}\left(F_{\mathcal{N}}, F_{\mathcal{M}}\right),
    \end{equation}
where $\mathrm{NN}$ denotes nearest neighbour search in $F_{\mathcal{N}}$ for each entry in $F_{\mathcal{M}}$. 

Furthermore, in the near-isometric setting, our method obtains more robust and accurate correspondences by first converting the point-wise map to a functional map and then recovering the final point-wise map via nearest neighbour search in the spectral domain, i.e.
    \begin{equation}
    \label{eq:iso_pmap}
        \Pi_{\mathcal{NM}}^{\text{iso}} = \mathrm{NN}\left(\phi_{\mathcal{N}}\phi_{\mathcal{N}}^{\dagger}\Pi_{\mathcal{NM}}\phi_{\mathcal{M}}, \phi_{\mathcal{M}}\right),
    \end{equation}
where $\Pi_{\mathcal{NM}}$ is the soft correspondence matrix defined in Eq.~\ref{eq:soft_corr}. On the one hand, this inference strategy explicitly enforces the functional map $C_{\mathcal{MN}}$ to be associated with a valid point-wise map (by replacing it with $\phi_{\mathcal{N}}^{\dagger}\Pi_{\mathcal{NM}}\phi_{\mathcal{M}}$). On the other hand, this strategy can be understood as a low-pass filter on the computed point-wise map $\Pi_{\mathcal{NM}}$, thereby making it more robust against mismatches. Considering the complete set of eigenfunctions (i.e.~$\phi_{\mathcal{N}}$ is of size $n_{\mathcal{N}} \times n_{\mathcal{N}}$, and $\phi_{\mathcal{N}} \phi_{\mathcal{N}}^{\dagger} = I$),
 Eq.~\ref{eq:iso_pmap} turns to be $\Pi_{\mathcal{NM}}^{\text{iso}} = \mathrm{NN}(\Pi_{\mathcal{NM}}\phi_{\mathcal{M}}, \phi_{\mathcal{M}})=\Pi_{\mathcal{NM}}$. By using the first $k$ eigenfunctions, it is equivalent to applying a low-pass filter on the Fourier domain in signal processing~\cite{ezuz2017deblurring}.

\begin{table*}[h!t!]
\small
\setlength{\tabcolsep}{6pt}
    \centering
    \small
    \caption{\textbf{Near-isometric shape matching and cross-dataset generalisation on FAUST, SCAPE and SHREC'19.} The numbers in parentheses show refined results using the indicated post-processing technique. The \textbf{best} results in each column are highlighted. Our method outperforms previous axiomatic, supervised and unsupervised methods in most settings without any post-processing techniques and demonstrates better cross-dataset generalisation ability (see columns in which \emph{Train} and \emph{Test} are different).}
    \label{tab:complete_shape}
        \begin{tabular}{@{}lccccccccc@{}}
        \toprule
        \multicolumn{1}{l}{Train}  & \multicolumn{3}{c}{\textbf{FAUST}}   & \multicolumn{3}{c}{\textbf{SCAPE}}  & \multicolumn{3}{c}{\textbf{FAUST + SCAPE}} \\ \cmidrule(lr){2-4} \cmidrule(lr){5-7} \cmidrule(lr){8-10}
        \multicolumn{1}{l}{Test} & \multicolumn{1}{c}{\textbf{FAUST}} & \multicolumn{1}{c}{\textbf{SCAPE}} & \multicolumn{1}{c}{\textbf{SHREC'19}} & \multicolumn{1}{c}{\textbf{FAUST}} & \multicolumn{1}{c}{\textbf{SCAPE}} & \multicolumn{1}{c}{\textbf{SHREC'19}} & \multicolumn{1}{c}{\textbf{FAUST}} & \multicolumn{1}{c}{\textbf{SCAPE}} & \multicolumn{1}{c}{\textbf{SHREC'19}}
        \\ \midrule
        \multicolumn{10}{c}{Axiomatic Methods} \\
        \multicolumn{1}{l}{BCICP}  & \multicolumn{1}{c}{6.1}  & \multicolumn{1}{c}{11.0} & \multicolumn{1}{c}{-}  & \multicolumn{1}{c}{6.1} & \multicolumn{1}{c}{11.0} & \multicolumn{1}{c}{-} & \multicolumn{1}{c}{6.1}    & \multicolumn{1}{c}{11.0} & \multicolumn{1}{c}{-}\\
        \multicolumn{1}{l}{ZoomOut} & \multicolumn{1}{c}{6.1}  & \multicolumn{1}{c}{7.5} & \multicolumn{1}{c}{-}  & \multicolumn{1}{c}{6.1} & \multicolumn{1}{c}{7.5} & \multicolumn{1}{c}{-} & \multicolumn{1}{c}{6.1}    & \multicolumn{1}{c}{7.5} & \multicolumn{1}{c}{-}\\
        \multicolumn{1}{l}{Smooth Shells} & \multicolumn{1}{c}{2.5}  & \multicolumn{1}{c}{4.7} & \multicolumn{1}{c}{-}  & \multicolumn{1}{c}{2.5} & \multicolumn{1}{c}{4.7} & \multicolumn{1}{c}{-} & \multicolumn{1}{c}{2.5}    & \multicolumn{1}{c}{4.7} & \multicolumn{1}{c}{-}\\ 
        \multicolumn{1}{l}{DiscreteOp} & \multicolumn{1}{c}{5.6}  & \multicolumn{1}{c}{13.1} & \multicolumn{1}{c}{-}  & \multicolumn{1}{c}{5.6} & \multicolumn{1}{c}{13.1} & \multicolumn{1}{c}{-} & \multicolumn{1}{c}{5.6}    & \multicolumn{1}{c}{13.1} & \multicolumn{1}{c}{-}\\ 
        \midrule
        \multicolumn{10}{c}{Supervised Methods} \\ 
        \multicolumn{1}{l}{FMNet (+ \textit{pmf})} & \multicolumn{1}{c}{11.0 (5.9)}  & \multicolumn{1}{c}{30.0 (11.0)} & \multicolumn{1}{c}{-}  & \multicolumn{1}{c}{33.0 (14.0)} & \multicolumn{1}{c}{17.0 (6.3)} & \multicolumn{1}{c}{-} & \multicolumn{1}{c}{-}    & \multicolumn{1}{c}{-} & \multicolumn{1}{c}{-} \\
        
        \multicolumn{1}{l}{3D-CODED} & \multicolumn{1}{c}{2.5}  & \multicolumn{1}{c}{31.0} & \multicolumn{1}{c}{-}  & \multicolumn{1}{c}{33.0} & \multicolumn{1}{c}{31.0} & \multicolumn{1}{c}{-} & \multicolumn{1}{c}{-}    & \multicolumn{1}{c}{-} & \multicolumn{1}{c}{-} \\
        \multicolumn{1}{l}{HSN} & \multicolumn{1}{c}{3.3} & \multicolumn{1}{c}{25.4} & \multicolumn{1}{c}{-} & \multicolumn{1}{c}{16.7} & \multicolumn{1}{c}{3.5} & \multicolumn{1}{c}{-} & \multicolumn{1}{c}{-} & \multicolumn{1}{c}{-} & \multicolumn{1}{c}{-} \\
        \multicolumn{1}{l}{ACSCNN} & \multicolumn{1}{c}{2.7}  & \multicolumn{1}{c}{8.4} & \multicolumn{1}{c}{-}  & \multicolumn{1}{c}{6.0} & \multicolumn{1}{c}{3.2} & \multicolumn{1}{c}{-} & \multicolumn{1}{c}{-}    & \multicolumn{1}{c}{-} & \multicolumn{1}{c}{-}\\
        \multicolumn{1}{l}{GeomFMaps (+ \textit{zoomout})}& \multicolumn{1}{c}{2.6 (1.9)}  & \multicolumn{1}{c}{3.4 (2.4)} & \multicolumn{1}{c}{9.9 (7.9)}  & \multicolumn{1}{c}{3.0 (1.9)} & \multicolumn{1}{c}{3.0 (2.4)} & \multicolumn{1}{c}{12.2 (9.8)} & \multicolumn{1}{c}{2.6 (1.9)}    & \multicolumn{1}{c}{2.9 (2.4)} & \multicolumn{1}{c}{7.9 (7.5)}\\
        
        \multicolumn{1}{l}{TransMatch}& \multicolumn{1}{c}{1.7}  & \multicolumn{1}{c}{30.4} & \multicolumn{1}{c}{14.5}  & \multicolumn{1}{c}{15.5} & \multicolumn{1}{c}{12.0} & \multicolumn{1}{c}{37.5} & \multicolumn{1}{c}{\textbf{1.6}} & \multicolumn{1}{c}{11.7} & \multicolumn{1}{c}{10.9}\\
        \midrule
        \multicolumn{10}{c}{Unsupervised Methods} \\ 
        \multicolumn{1}{l}{SURFMNet (+ \textit{icp})} & \multicolumn{1}{c}{15.0 (7.4)}  & \multicolumn{1}{c}{32.0 (19.0)} & \multicolumn{1}{c}{-}  & \multicolumn{1}{c}{32.0 (23.0)} & \multicolumn{1}{c}{12.0 (6.1)} & \multicolumn{1}{c}{-} & \multicolumn{1}{c}{33.0 (23.0)}    & \multicolumn{1}{c}{29.0 (17.0)} & \multicolumn{1}{c}{-} \\

        \multicolumn{1}{l}{UnsupFMNet (+ \textit{pmf})}  & \multicolumn{1}{c}{10.0 (5.7)}  & \multicolumn{1}{c}{29.0 (12.0)} & \multicolumn{1}{c}{-}  & \multicolumn{1}{c}{22.0 (9.3)} & \multicolumn{1}{c}{16.0 (10.0)} & \multicolumn{1}{c}{-} & \multicolumn{1}{c}{11.0 (6.7)}    & \multicolumn{1}{c}{13.0 (8.3)} & \multicolumn{1}{c}{-} \\

        \multicolumn{1}{l}{WSupFMNet (+ \textit{zoomout})} & \multicolumn{1}{c}{3.8 (1.9)}  & \multicolumn{1}{c}{4.8 (2.7)} & \multicolumn{1}{c}{-}  & \multicolumn{1}{c}{3.6 (1.9)} & \multicolumn{1}{c}{4.4 (2.6)} & \multicolumn{1}{c}{-} & \multicolumn{1}{c}{3.6 (1.9)}    & \multicolumn{1}{c}{4.5 (2.6)} & \multicolumn{1}{c}{-} \\

        \multicolumn{1}{l}{Deep Shells} & \multicolumn{1}{c}{1.7}  & \multicolumn{1}{c}{5.4} & \multicolumn{1}{c}{27.4}  & \multicolumn{1}{c}{2.7} & \multicolumn{1}{c}{2.5} & \multicolumn{1}{c}{23.4} & \multicolumn{1}{c}{\textbf{1.6}}    & \multicolumn{1}{c}{2.4} & \multicolumn{1}{c}{21.1} \\
        \multicolumn{1}{l}{NeuroMorph} & \multicolumn{1}{c}{8.5}  & \multicolumn{1}{c}{28.5} & \multicolumn{1}{c}{26.3}  & \multicolumn{1}{c}{18.2} & \multicolumn{1}{c}{29.9} & \multicolumn{1}{c}{27.6} & \multicolumn{1}{c}{9.1}    & \multicolumn{1}{c}{27.3} & \multicolumn{1}{c}{25.3} \\
        \multicolumn{1}{l}{ConsistFMaps} & \multicolumn{1}{c}{\textbf{1.5}}  & \multicolumn{1}{c}{3.2} & \multicolumn{1}{c}{19.7}  & \multicolumn{1}{c}{3.2} & \multicolumn{1}{c}{2.0} & \multicolumn{1}{c}{28.3} & \multicolumn{1}{c}{1.7}    & \multicolumn{1}{c}{3.2} & \multicolumn{1}{c}{17.8} \\
        \multicolumn{1}{l}{DUO-FMNet}  & \multicolumn{1}{c}{2.5}  & \multicolumn{1}{c}{4.2} & \multicolumn{1}{c}{6.4}  & \multicolumn{1}{c}{2.7} & \multicolumn{1}{c}{2.6} & \multicolumn{1}{c}{8.4} & \multicolumn{1}{c}{2.5}    & \multicolumn{1}{c}{4.3} & \multicolumn{1}{c}{6.4} \\
        \multicolumn{1}{l}{AttentiveFMaps}  & \multicolumn{1}{c}{1.9}  & \multicolumn{1}{c}{2.6} & \multicolumn{1}{c}{6.4}  & \multicolumn{1}{c}{2.2} & \multicolumn{1}{c}{2.2} & \multicolumn{1}{c}{9.9} & \multicolumn{1}{c}{1.9}    & \multicolumn{1}{c}{2.3} & \multicolumn{1}{c}{5.8}\\
        \multicolumn{1}{l}{AttentiveFMaps-Fast}  & \multicolumn{1}{c}{1.9}  & \multicolumn{1}{c}{2.6} & \multicolumn{1}{c}{5.8}  & \multicolumn{1}{c}{1.9} & \multicolumn{1}{c}{2.1} & \multicolumn{1}{c}{8.1} & \multicolumn{1}{c}{1.9}    & \multicolumn{1}{c}{2.3} & \multicolumn{1}{c}{6.3}\\
        \multicolumn{1}{l}{Ours}  & \multicolumn{1}{c}{1.6}  & \multicolumn{1}{c}{\textbf{2.2}} & \multicolumn{1}{c}{\textbf{5.7}}  & \multicolumn{1}{c}{\textbf{1.6}} & \multicolumn{1}{c}{\textbf{1.9}} & \multicolumn{1}{c}{\textbf{6.7}} & \multicolumn{1}{c}{\textbf{1.6}}    & \multicolumn{1}{c}{\textbf{2.1}} & \multicolumn{1}{c}{\textbf{4.6}} \\\hline
        \end{tabular} 
\end{table*}

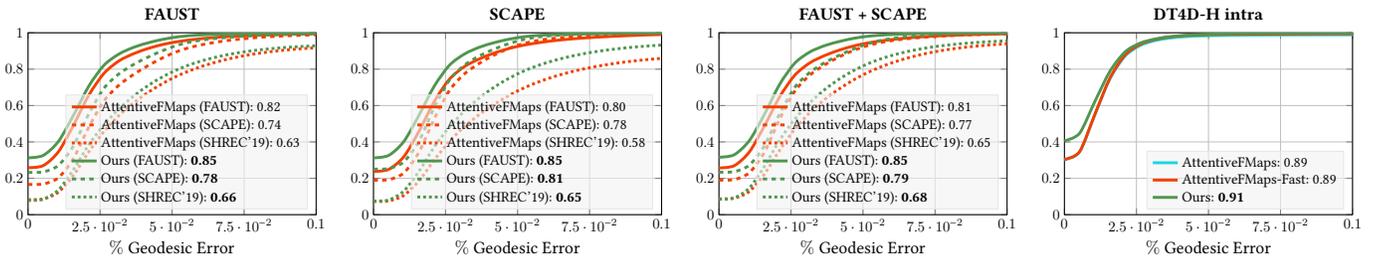
\begin{figure*}[ht]
    \centering
    \begin{tabular}{cccc}
     \hspace{-1.2cm}
     \newcommand{\pckLineWidth}{2pt}
\newcommand{\plotWidth}{\columnwidth}
\newcommand{\plotHeight}{0.7\columnwidth}
\newcommand{\pckTitle}{\textbf{FAUST}}
\definecolor{cPLOT0}{RGB}{28,213,227}
\definecolor{cPLOT1}{RGB}{80,150,80}
\definecolor{cPLOT2}{RGB}{90,130,213}
\definecolor{cPLOT3}{RGB}{247,179,43}
\definecolor{cPLOT4}{RGB}{124,42,43}
\definecolor{cPLOT5}{RGB}{242,64,0}

\pgfplotsset{%
    label style = {font=\large},
    tick label style = {font=\large},
    title style =  {font=\LARGE},
    legend style={  fill= gray!10,
                    fill opacity=0.6, 
                    font=\large,
                    draw=gray!20, %
                    text opacity=1}
}
\begin{tikzpicture}[scale=0.55, transform shape]
	\begin{axis}[
		width=\plotWidth,
		height=\plotHeight,
		grid=major,
		title=\pckTitle,
		legend style={
			at={(0.97,0.03)},
			anchor=south east,
			legend columns=1},
		legend cell align={left},
        xlabel={\LARGE$\%$ Geodesic Error},
		xmin=0,
        xmax=0.1,
        ylabel near ticks,
        xtick={0, 0.025, 0.05, 0.075, 0.1},
	ymin=0,
        ymax=1,
        ytick={0, 0.20, 0.40, 0.60, 0.80, 1.0}
	]

\addplot [color=cPLOT5, smooth, line width=\pckLineWidth]
table[row sep=crcr]{%
0.0 0.25808947368421054 \\
0.005263157894736842 0.26970736842105264 \\
0.010526315789473684 0.34548736842105265 \\
0.015789473684210527 0.4920557894736842 \\
0.021052631578947368 0.6559094736842105 \\
0.02631578947368421 0.7761673684210526 \\
0.031578947368421054 0.8456231578947369 \\
0.03684210526315789 0.8862315789473684 \\
0.042105263157894736 0.9158031578947369 \\
0.04736842105263158 0.9377410526315789 \\
0.05263157894736842 0.9530284210526315 \\
0.05789473684210526 0.9637242105263157 \\
0.06315789473684211 0.9721652631578948 \\
0.06842105263157895 0.9783347368421053 \\
0.07368421052631578 0.98296 \\
0.07894736842105263 0.9864778947368421 \\
0.08421052631578947 0.9891210526315789 \\
0.08947368421052632 0.9914873684210527 \\
0.09473684210526316 0.9931968421052632 \\
0.1 0.9947884210526315 \\
    };
\addlegendentry{\textcolor{black}{AttentiveFMaps (FAUST): {0.82}}}    

\addplot [color=cPLOT5, smooth, dashed, line width=\pckLineWidth]
table[row sep=crcr]{%
0.0 0.1664 \\
0.005263157894736842 0.16884947368421052 \\
0.010526315789473684 0.20285263157894737 \\
0.015789473684210527 0.3098336842105263 \\
0.021052631578947368 0.48037473684210524 \\
0.02631578947368421 0.6245536842105263 \\
0.031578947368421054 0.7123347368421052 \\
0.03684210526315789 0.77312 \\
0.042105263157894736 0.8233884210526315 \\
0.04736842105263158 0.8671936842105263 \\
0.05263157894736842 0.9019084210526316 \\
0.05789473684210526 0.9259357894736842 \\
0.06315789473684211 0.9426231578947368 \\
0.06842105263157895 0.95578 \\
0.07368421052631578 0.9654631578947368 \\
0.07894736842105263 0.9733084210526316 \\
0.08421052631578947 0.9789821052631579 \\
0.08947368421052632 0.9830642105263158 \\
0.09473684210526316 0.9864473684210526 \\
0.1 0.9890568421052631 \\
    };
\addlegendentry{\textcolor{black}{AttentiveFMaps (SCAPE): {0.74}}}  

\addplot [color=cPLOT5, smooth, dotted, line width=\pckLineWidth]
table[row sep=crcr]{%
0.0 0.08273666372582543 \\
0.005263157894736842 0.0867563906621049 \\
0.010526315789473684 0.12378170771564648 \\
0.015789473684210527 0.21680501844795372 \\
0.021052631578947368 0.3444028927298248 \\
0.02631578947368421 0.4566912324517737 \\
0.031578947368421054 0.545454416646043 \\
0.03684210526315789 0.6172611637040998 \\
0.042105263157894736 0.676746132352969 \\
0.04736842105263158 0.7270293221392448 \\
0.05263157894736842 0.7686236485197241 \\
0.05789473684210526 0.8019310369582507 \\
0.06315789473684211 0.8293578449616211 \\
0.06842105263157895 0.8517268153712166 \\
0.07368421052631578 0.8700538608459422 \\
0.07894736842105263 0.884449026783066 \\
0.08421052631578947 0.8961166726805925 \\
0.08947368421052632 0.9051814851689032 \\
0.09473684210526316 0.9122352534444682 \\
0.1 0.9177559429237504 \\
    };
\addlegendentry{\textcolor{black}{AttentiveFMaps (SHREC'19): {0.63}}}  

\addplot [color=cPLOT1, smooth, line width=\pckLineWidth]
table[row sep=crcr]{%
0.0 0.312412 \\
0.005263157894736842 0.3258105 \\
0.010526315789473684 0.4023205 \\
0.015789473684210527 0.5468495 \\
0.021052631578947368 0.705658 \\
0.02631578947368421 0.8208325 \\
0.031578947368421054 0.8832595 \\
0.03684210526315789 0.921061 \\
0.042105263157894736 0.9465335 \\
0.04736842105263158 0.9657115 \\
0.05263157894736842 0.978877 \\
0.05789473684210526 0.9863015 \\
0.06315789473684211 0.9904765 \\
0.06842105263157895 0.993006 \\
0.07368421052631578 0.994677 \\
0.07894736842105263 0.9958235 \\
0.08421052631578947 0.9966505 \\
0.08947368421052632 0.997221 \\
0.09473684210526316 0.997665 \\
0.1 0.9979685 \\
    };
\addlegendentry{\textcolor{black}{Ours (FAUST): \textbf{0.85}}}    

\addplot [color=cPLOT1, smooth, dashed, line width=\pckLineWidth]
table[row sep=crcr]{%
0.0 0.2326715 \\
0.005263157894736842 0.235328 \\
0.010526315789473684 0.2706745 \\
0.015789473684210527 0.3793195 \\
0.021052631578947368 0.553259 \\
0.02631578947368421 0.694306 \\
0.031578947368421054 0.7743975 \\
0.03684210526315789 0.827544 \\
0.042105263157894736 0.8695885 \\
0.04736842105263158 0.9060175 \\
0.05263157894736842 0.9350335 \\
0.05789473684210526 0.9540125 \\
0.06315789473684211 0.965561 \\
0.06842105263157895 0.9736805 \\
0.07368421052631578 0.9801065 \\
0.07894736842105263 0.9849945 \\
0.08421052631578947 0.988798 \\
0.08947368421052632 0.991417 \\
0.09473684210526316 0.9934385 \\
0.1 0.994945 \\
    };
\addlegendentry{\textcolor{black}{Ours (SCAPE): \textbf{0.78}}}  

\addplot [color=cPLOT1, smooth, dotted, line width=\pckLineWidth]
table[row sep=crcr]{%
0.0 0.07800518394111769 \\
0.005263157894736842 0.0846060185858647 \\
0.010526315789473684 0.13572328636173858 \\
0.015789473684210527 0.24823738445233287 \\
0.021052631578947368 0.38378450371887324 \\
0.02631578947368421 0.4935942243641454 \\
0.031578947368421054 0.5815431670781955 \\
0.03684210526315789 0.654493819510435 \\
0.042105263157894736 0.714406217706635 \\
0.04736842105263158 0.7636560198138391 \\
0.05263157894736842 0.8031216998188265 \\
0.05789473684210526 0.8339462185000954 \\
0.06315789473684211 0.8579776773147789 \\
0.06842105263157895 0.877022143212041 \\
0.07368421052631578 0.8919802881772515 \\
0.07894736842105263 0.903522775146554 \\
0.08421052631578947 0.9124473151756853 \\
0.08947368421052632 0.9193282224409486 \\
0.09473684210526316 0.9245499473860204 \\
0.1 0.9286561048274508 \\
    };
\addlegendentry{\textcolor{black}{Ours (SHREC'19): \textbf{0.66}}}  
\end{axis}
\end{tikzpicture}&
     \hspace{-1cm}
     \newcommand{\pckLineWidth}{2pt}
\newcommand{\plotWidth}{\columnwidth}
\newcommand{\plotHeight}{0.7\columnwidth}
\newcommand{\pckTitle}{\textbf{SCAPE}}
\definecolor{cPLOT0}{RGB}{28,213,227}
\definecolor{cPLOT1}{RGB}{80,150,80}
\definecolor{cPLOT2}{RGB}{90,130,213}
\definecolor{cPLOT3}{RGB}{247,179,43}
\definecolor{cPLOT5}{RGB}{242,64,0}

\pgfplotsset{%
    label style = {font=\large},
    tick label style = {font=\large},
    title style =  {font=\LARGE},
    legend style={  fill= gray!10,
                    fill opacity=0.6, 
                    font=\large,
                    draw=gray!20, %
                    text opacity=1}
}
\begin{tikzpicture}[scale=0.55, transform shape]
	\begin{axis}[
		width=\plotWidth,
		height=\plotHeight,
		grid=major,
		title=\pckTitle,
		legend style={
			at={(0.97,0.03)},
			anchor=south east,
			legend columns=1},
		legend cell align={left},
        xlabel={\LARGE$\%$ Geodesic Error},
		xmin=0,
        xmax=0.1,
        ylabel near ticks,
        xtick={0, 0.025, 0.05, 0.075, 0.1},
	ymin=0,
        ymax=1,
        ytick={0, 0.20, 0.40, 0.60, 0.80, 1.0}
	]

\addplot [color=cPLOT5, smooth, line width=\pckLineWidth]
table[row sep=crcr]{%
0.0 0.23745894736842105 \\
0.005263157894736842 0.24896736842105263 \\
0.010526315789473684 0.3188778947368421 \\
0.015789473684210527 0.45857473684210526 \\
0.021052631578947368 0.6213136842105264 \\
0.02631578947368421 0.7439684210526316 \\
0.031578947368421054 0.8143252631578948 \\
0.03684210526315789 0.8569231578947368 \\
0.042105263157894736 0.8897778947368421 \\
0.04736842105263158 0.9150684210526315 \\
0.05263157894736842 0.9333831578947368 \\
0.05789473684210526 0.9468421052631579 \\
0.06315789473684211 0.95752 \\
0.06842105263157895 0.9657094736842106 \\
0.07368421052631578 0.9720873684210526 \\
0.07894736842105263 0.9770336842105263 \\
0.08421052631578947 0.9812115789473684 \\
0.08947368421052632 0.9849136842105263 \\
0.09473684210526316 0.9879536842105263 \\
0.1 0.9904115789473684 \\
    };
\addlegendentry{\textcolor{black}{AttentiveFMaps (FAUST): {0.80}}}    

\addplot [color=cPLOT5, smooth, dashed, line width=\pckLineWidth]
table[row sep=crcr]{%
0.0 0.18994526315789473 \\
0.005263157894736842 0.19249263157894736 \\
0.010526315789473684 0.22945157894736842 \\
0.015789473684210527 0.34391894736842105 \\
0.021052631578947368 0.5259463157894737 \\
0.02631578947368421 0.6822547368421052 \\
0.031578947368421054 0.77422 \\
0.03684210526315789 0.8329821052631579 \\
0.042105263157894736 0.8792747368421052 \\
0.04736842105263158 0.9162526315789473 \\
0.05263157894736842 0.9429273684210526 \\
0.05789473684210526 0.9598747368421052 \\
0.06315789473684211 0.9704663157894737 \\
0.06842105263157895 0.9777010526315789 \\
0.07368421052631578 0.98298 \\
0.07894736842105263 0.9869557894736842 \\
0.08421052631578947 0.9898021052631579 \\
0.08947368421052632 0.9920073684210526 \\
0.09473684210526316 0.9938705263157894 \\
0.1 0.995238947368421 \\
    };
\addlegendentry{\textcolor{black}{AttentiveFMaps (SCAPE): {0.78}}}  

\addplot [color=cPLOT5, smooth, dotted, line width=\pckLineWidth]
table[row sep=crcr]{%
0.0 0.07293506660344173 \\
0.005263157894736842 0.07576035229640657 \\
0.010526315789473684 0.10597655135672278 \\
0.015789473684210527 0.18601261601996497 \\
0.021052631578947368 0.29988787649222504 \\
0.02631578947368421 0.4019759997128429 \\
0.031578947368421054 0.4846319005076257 \\
0.03684210526315789 0.5520675310346912 \\
0.042105263157894736 0.608849066833923 \\
0.04736842105263158 0.6583746908810622 \\
0.05263157894736842 0.699793794762028 \\
0.05789473684210526 0.7333231631864613 \\
0.06315789473684211 0.761420634125975 \\
0.06842105263157895 0.7845901493783616 \\
0.07368421052631578 0.8038169222427725 \\
0.07894736842105263 0.8194698173340865 \\
0.08421052631578947 0.8326204784566063 \\
0.08947368421052632 0.8434790781501794 \\
0.09473684210526316 0.852238786232329 \\
0.1 0.8595891575528927 \\
    };
\addlegendentry{\textcolor{black}{AttentiveFMaps (SHREC'19): {0.58}}}

\addplot [color=cPLOT1, smooth, line width=\pckLineWidth]
table[row sep=crcr]{%
0.0 0.312412 \\
0.005263157894736842 0.3258105 \\
0.010526315789473684 0.4023205 \\
0.015789473684210527 0.5468495 \\
0.021052631578947368 0.705658 \\
0.02631578947368421 0.8208325 \\
0.031578947368421054 0.8832595 \\
0.03684210526315789 0.921061 \\
0.042105263157894736 0.9465335 \\
0.04736842105263158 0.9657115 \\
0.05263157894736842 0.978877 \\
0.05789473684210526 0.9863015 \\
0.06315789473684211 0.9904765 \\
0.06842105263157895 0.993006 \\
0.07368421052631578 0.994677 \\
0.07894736842105263 0.9958235 \\
0.08421052631578947 0.9966505 \\
0.08947368421052632 0.997221 \\
0.09473684210526316 0.997665 \\
0.1 0.9979685 \\
    };
\addlegendentry{\textcolor{black}{Ours (FAUST): \textbf{0.85}}}    

\addplot [color=cPLOT1, smooth, dashed, line width=\pckLineWidth]
table[row sep=crcr]{%
0.0 0.2509255 \\
0.005263157894736842 0.2536675 \\
0.010526315789473684 0.290562 \\
0.015789473684210527 0.404467 \\
0.021052631578947368 0.588076 \\
0.02631578947368421 0.738729 \\
0.031578947368421054 0.8219865 \\
0.03684210526315789 0.87358 \\
0.042105263157894736 0.911547 \\
0.04736842105263158 0.942448 \\
0.05263157894736842 0.964943 \\
0.05789473684210526 0.978441 \\
0.06315789473684211 0.985795 \\
0.06842105263157895 0.9904385 \\
0.07368421052631578 0.993671 \\
0.07894736842105263 0.995846 \\
0.08421052631578947 0.997343 \\
0.08947368421052632 0.9983675 \\
0.09473684210526316 0.999008 \\
0.1 0.9994695 \\
    };
\addlegendentry{\textcolor{black}{Ours (SCAPE): \textbf{0.81}}}  

\addplot [color=cPLOT1, smooth, dotted, line width=\pckLineWidth]
table[row sep=crcr]{%
0.0 0.0747335767871278 \\
0.005263157894736842 0.08089611903416981 \\
0.010526315789473684 0.13000564868219455 \\
0.015789473684210527 0.23882543304989165 \\
0.021052631578947368 0.3715482112191519 \\
0.02631578947368421 0.48138626973496534 \\
0.031578947368421054 0.569854267777763 \\
0.03684210526315789 0.6431208874665377 \\
0.042105263157894736 0.7032637670098068 \\
0.04736842105263158 0.7527034328496374 \\
0.05263157894736842 0.7927429547330856 \\
0.05789473684210526 0.8244095804672726 \\
0.06315789473684211 0.8496019473784636 \\
0.06842105263157895 0.8703660497197384 \\
0.07368421052631578 0.8871364015060633 \\
0.07894736842105263 0.9005057365295932 \\
0.08421052631578947 0.9112651536762719 \\
0.08947368421052632 0.9199275684028965 \\
0.09473684210526316 0.9267739979256678 \\
0.1 0.9322739062998865 \\
    };
\addlegendentry{\textcolor{black}{Ours (SHREC'19): \textbf{0.65}}}  
\end{axis}
\end{tikzpicture}&
     \hspace{-1cm}
     \newcommand{\pckLineWidth}{2pt}
\newcommand{\plotWidth}{\columnwidth}
\newcommand{\plotHeight}{0.7\columnwidth}
\newcommand{\pckTitle}{\textbf{FAUST + SCAPE}}
\definecolor{cPLOT0}{RGB}{28,213,227}
\definecolor{cPLOT1}{RGB}{80,150,80}
\definecolor{cPLOT2}{RGB}{90,130,213}
\definecolor{cPLOT3}{RGB}{247,179,43}
\definecolor{cPLOT5}{RGB}{242,64,0}

\pgfplotsset{%
    label style = {font=\large},
    tick label style = {font=\large},
    title style =  {font=\LARGE},
    legend style={  fill= gray!10,
                    fill opacity=0.6, 
                    font=\large,
                    draw=gray!20, %
                    text opacity=1}
}
\begin{tikzpicture}[scale=0.55, transform shape]
	\begin{axis}[
		width=\plotWidth,
		height=\plotHeight,
		grid=major,
		title=\pckTitle,
		legend style={
			at={(0.97,0.03)},
			anchor=south east,
			legend columns=1},
		legend cell align={left},
        xlabel={\LARGE$\%$ Geodesic Error},
		xmin=0,
        xmax=0.1,
        ylabel near ticks,
        xtick={0, 0.025, 0.05, 0.075, 0.1},
	ymin=0,
        ymax=1,
        ytick={0, 0.20, 0.40, 0.60, 0.80, 1.0}
	]
\addplot [color=cPLOT5, smooth, line width=\pckLineWidth]
table[row sep=crcr]{%
0.0 0.25719473684210525 \\
0.005263157894736842 0.2690484210526316 \\
0.010526315789473684 0.3446252631578947 \\
0.015789473684210527 0.48797052631578947 \\
0.021052631578947368 0.6474957894736842 \\
0.02631578947368421 0.7666421052631579 \\
0.031578947368421054 0.834958947368421 \\
0.03684210526315789 0.8757147368421052 \\
0.042105263157894736 0.9063884210526316 \\
0.04736842105263158 0.9298431578947368 \\
0.05263157894736842 0.9465821052631579 \\
0.05789473684210526 0.9586452631578948 \\
0.06315789473684211 0.9683884210526316 \\
0.06842105263157895 0.9754884210526316 \\
0.07368421052631578 0.9807589473684211 \\
0.07894736842105263 0.9848326315789474 \\
0.08421052631578947 0.98796 \\
0.08947368421052632 0.9906273684210526 \\
0.09473684210526316 0.9926715789473685 \\
0.1 0.9944042105263158 \\
    };
\addlegendentry{\textcolor{black}{AttentiveFMaps (FAUST): {0.81}}}    

\addplot [color=cPLOT5, smooth, dashed, line width=\pckLineWidth]
table[row sep=crcr]{%
0.0 0.19041052631578947 \\
0.005263157894736842 0.1930578947368421 \\
0.010526315789473684 0.23041894736842106 \\
0.015789473684210527 0.3434494736842105 \\
0.021052631578947368 0.5250231578947369 \\
0.02631578947368421 0.6784694736842105 \\
0.031578947368421054 0.7675842105263158 \\
0.03684210526315789 0.8249831578947369 \\
0.042105263157894736 0.8703442105263158 \\
0.04736842105263158 0.9070926315789474 \\
0.05263157894736842 0.9340031578947369 \\
0.05789473684210526 0.9517 \\
0.06315789473684211 0.96336 \\
0.06842105263157895 0.9719115789473685 \\
0.07368421052631578 0.9783652631578947 \\
0.07894736842105263 0.9835273684210526 \\
0.08421052631578947 0.9874147368421052 \\
0.08947368421052632 0.990301052631579 \\
0.09473684210526316 0.9926526315789473 \\
0.1 0.99436 \\
    };
\addlegendentry{\textcolor{black}{AttentiveFMaps (SCAPE): {0.77}}}  

\addplot [color=cPLOT5, smooth, dotted, line width=\pckLineWidth]
table[row sep=crcr]{%
0.0 0.08734770756073278 \\
0.005263157894736842 0.09112325651251495 \\
0.010526315789473684 0.12774948188926694 \\
0.015789473684210527 0.22314514468372101 \\
0.021052631578947368 0.35432728729122076 \\
0.02631578947368421 0.4700142822867528 \\
0.031578947368421054 0.5620849682710309 \\
0.03684210526315789 0.6353572555339139 \\
0.042105263157894736 0.6957003893623412 \\
0.04736842105263158 0.7467024164646806 \\
0.05263157894736842 0.7888724739149902 \\
0.05789473684210526 0.8227990448248437 \\
0.06315789473684211 0.850717514882105 \\
0.06842105263157895 0.8734737695224313 \\
0.07368421052631578 0.8919850111556751 \\
0.07894736842105263 0.9064722751720581 \\
0.08421052631578947 0.9179580864002781 \\
0.08947368421052632 0.9269761414021956 \\
0.09473684210526316 0.9340096008705394 \\
0.1 0.9395052585641768 \\
    };
\addlegendentry{\textcolor{black}{AttentiveFMaps (SHREC'19): {0.65}}}

\addplot [color=cPLOT1, smooth, line width=\pckLineWidth]
table[row sep=crcr]{%
0.0 0.3156175 \\
0.005263157894736842 0.3291785 \\
0.010526315789473684 0.4068525 \\
0.015789473684210527 0.552004 \\
0.021052631578947368 0.716368 \\
0.02631578947368421 0.8328685 \\
0.031578947368421054 0.894642 \\
0.03684210526315789 0.930021 \\
0.042105263157894736 0.953291 \\
0.04736842105263158 0.971183 \\
0.05263157894736842 0.982451 \\
0.05789473684210526 0.988614 \\
0.06315789473684211 0.9923915 \\
0.06842105263157895 0.994758 \\
0.07368421052631578 0.9962195 \\
0.07894736842105263 0.9971035 \\
0.08421052631578947 0.99766 \\
0.08947368421052632 0.9980335 \\
0.09473684210526316 0.998292 \\
0.1 0.9984605 \\
    };
\addlegendentry{\textcolor{black}{Ours (FAUST): \textbf{0.85}}}    

\addplot [color=cPLOT1, smooth, dashed, line width=\pckLineWidth]
table[row sep=crcr]{%
0.0 0.233816 \\
0.005263157894736842 0.2365165 \\
0.010526315789473684 0.272099 \\
0.015789473684210527 0.3797695 \\
0.021052631578947368 0.554364 \\
0.02631578947368421 0.6999725 \\
0.031578947368421054 0.7824525 \\
0.03684210526315789 0.8370525 \\
0.042105263157894736 0.878866 \\
0.04736842105263158 0.915387 \\
0.05263157894736842 0.944451 \\
0.05789473684210526 0.9633105 \\
0.06315789473684211 0.974568 \\
0.06842105263157895 0.982541 \\
0.07368421052631578 0.988246 \\
0.07894736842105263 0.992311 \\
0.08421052631578947 0.995135 \\
0.08947368421052632 0.9968635 \\
0.09473684210526316 0.997932 \\
0.1 0.998679 \\
    };
\addlegendentry{\textcolor{black}{Ours (SCAPE): \textbf{0.79}}}  

\addplot [color=cPLOT1, smooth, dotted, line width=\pckLineWidth]
table[row sep=crcr]{%
0.0 0.08493615477767051 \\
0.005263157894736842 0.09203290215688979 \\
0.010526315789473684 0.14640146827953232 \\
0.015789473684210527 0.2659891711550705 \\
0.021052631578947368 0.4096697315647983 \\
0.02631578947368421 0.5248116948502533 \\
0.031578947368421054 0.6158872492806904 \\
0.03684210526315789 0.6895222612865015 \\
0.042105263157894736 0.7490931881426793 \\
0.04736842105263158 0.7974102020112331 \\
0.05263157894736842 0.835978516115747 \\
0.05789473684210526 0.8656676307839956 \\
0.06315789473684211 0.888657200558445 \\
0.06842105263157895 0.906745263324939 \\
0.07368421052631578 0.9209401749769046 \\
0.07894736842105263 0.9317746874805177 \\
0.08421052631578947 0.9401239687376612 \\
0.08947368421052632 0.9467333047435706 \\
0.09473684210526316 0.9518978816497174 \\
0.1 0.9558481808031708 \\
    };
\addlegendentry{\textcolor{black}{Ours (SHREC'19): \textbf{0.68}}}  
\end{axis}
\end{tikzpicture}&
     \hspace{-1cm}
     \newcommand{\pckLineWidth}{2pt}
\newcommand{\plotWidth}{\columnwidth}
\newcommand{\plotHeight}{0.7\columnwidth}
\newcommand{\pckTitle}{\textbf{DT4D-H intra}}
\definecolor{cPLOT0}{RGB}{28,213,227}
\definecolor{cPLOT1}{RGB}{80,150,80}
\definecolor{cPLOT2}{RGB}{90,130,213}
\definecolor{cPLOT3}{RGB}{247,179,43}
\definecolor{cPLOT5}{RGB}{242,64,0}

\pgfplotsset{%
    label style = {font=\large},
    tick label style = {font=\large},
    title style =  {font=\LARGE},
    legend style={  fill= gray!10,
                    fill opacity=0.6, 
                    font=\large,
                    draw=gray!20, %
                    text opacity=1}
}
\begin{tikzpicture}[scale=0.55, transform shape]
	\begin{axis}[
		width=\plotWidth,
		height=\plotHeight,
		grid=major,
		title=\pckTitle,
		legend style={
			at={(0.97,0.03)},
			anchor=south east,
			legend columns=1},
		legend cell align={left},
        xlabel={\LARGE$\%$ Geodesic Error},
		xmin=0,
        xmax=0.1,
        ylabel near ticks,
        xtick={0, 0.025, 0.05, 0.075, 0.1},
	ymin=0,
        ymax=1,
        ytick={0, 0.20, 0.40, 0.60, 0.80, 1.0}
	]   

\addplot [color=cPLOT0, smooth, line width=\pckLineWidth]
table[row sep=crcr]{%
0.0 0.30256685034312564 \\
0.005263157894736842 0.3451083465738143 \\
0.010526315789473684 0.5475264527906426 \\
0.015789473684210527 0.746411547953078 \\
0.021052631578947368 0.8718221155471417 \\
0.02631578947368421 0.9303296034616815 \\
0.031578947368421054 0.9575403130387748 \\
0.03684210526315789 0.9714952165241202 \\
0.042105263157894736 0.9788796862851155 \\
0.04736842105263158 0.982714918359758 \\
0.05263157894736842 0.9847043710489841 \\
0.05789473684210526 0.9859220445556269 \\
0.06315789473684211 0.9867536594435617 \\
0.06842105263157895 0.9872958993948818 \\
0.07368421052631578 0.9876893952199046 \\
0.07894736842105263 0.9880142659139313 \\
0.08421052631578947 0.9882877522734187 \\
0.08947368421052632 0.9885179676143471 \\
0.09473684210526316 0.9887066022108786 \\
0.1 0.9888725871336331 \\
    };
\addlegendentry{\textcolor{black}{AttentiveFMaps: 0.89}}  

\addplot [color=cPLOT5, smooth, line width=\pckLineWidth]
table[row sep=crcr]{%
0.0 0.3037696494371387 \\
0.005263157894736842 0.3467665055271965 \\
0.010526315789473684 0.5503891011122004 \\
0.015789473684210527 0.750904972786586 \\
0.021052631578947368 0.8777123829485143 \\
0.02631578947368421 0.9365383185152631 \\
0.031578947368421054 0.9637243500895846 \\
0.03684210526315789 0.9774612082079713 \\
0.042105263157894736 0.9847361482032385 \\
0.04736842105263158 0.9884784151989453 \\
0.05263157894736842 0.9904793617524763 \\
0.05789473684210526 0.9916557925695548 \\
0.06315789473684211 0.9924157398330009 \\
0.06842105263157895 0.9929613603326459 \\
0.07368421052631578 0.9933616172543186 \\
0.07894736842105263 0.9937074473479598 \\
0.08421052631578947 0.99397180622697 \\
0.08947368421052632 0.9942344748318177 \\
0.09473684210526316 0.9944657043372435 \\
0.1 0.9946654947432474 \\
    };
\addlegendentry{\textcolor{black}{AttentiveFMaps-Fast: 0.89}} 

\addplot [color=cPLOT1, smooth, line width=\pckLineWidth]
table[row sep=crcr]{%
0.0 0.4061668264912008 \\
0.005263157894736842 0.4477582998207186 \\
0.010526315789473684 0.6207939173182833 \\
0.015789473684210527 0.7882817409629649 \\
0.021052631578947368 0.8949142443458122 \\
0.02631578947368421 0.9434205987924088 \\
0.031578947368421054 0.9660722204744704 \\
0.03684210526315789 0.978677410918353 \\
0.042105263157894736 0.9862126891836062 \\
0.04736842105263158 0.9905672949729183 \\
0.05263157894736842 0.9933010146646907 \\
0.05789473684210526 0.9951420416857761 \\
0.06315789473684211 0.9964023939567375 \\
0.06842105263157895 0.9972267095223837 \\
0.07368421052631578 0.9977543200003032 \\
0.07894736842105263 0.9980962055255068 \\
0.08421052631578947 0.9983224109373046 \\
0.08947368421052632 0.9984873650177576 \\
0.09473684210526316 0.99861547733208 \\
0.1 0.9987161478370623 \\
    };
\addlegendentry{\textcolor{black}{Ours: \textbf{0.91}}}  
\end{axis}
\end{tikzpicture}
    \end{tabular}
    \caption{{\textbf{Near-isometric shape matching on FAUST, SCAPE, SHREC'19 and DT4D-H intra  and cross-dataset generalisation on FAUST, SCAPE and SHREC'19.} Proportion of correct keypoints (PCK) curves and corresponding area under curve (scores in the legend) of our method in comparison to the existing state-of-the-art method. The title of each figure indicates the used training dataset and the names in parentheses shown in the legend correspond to the test dataset.}
    }
    \label{fig:iso_pck}
\end{figure*}

\subsection{Implementation Details}
We use DiffusionNet as our feature extractor and wave kernel signatures (WKS)~\cite{aubry2011wave} as input features. {Following~\citet{li2022learning}, the dimension of the WKS is 128}. For the spectral resolution, we choose the first 200 eigenfunctions of the Laplacian matrices as the spectral embedding. {The dimension of the output features from DiffusionNet is 256 rather than 128 in the original implementation. The total number of learnable parameters is 510,336.} In the context of the functional map computation, we use the regularised functional map solver proposed by \citet{ren2019structured} and set $\lambda=100$ in Eq.~\ref{eq:fmap}. For point-wise map computation, we set $\tau = 0.07$ in Eq.~\ref{eq:soft_corr}. In terms of our unsupervised loss, we empirically set $\lambda_{\mathrm{bij}} = 1.0, \lambda_{\mathrm{orth}} = 1.0$ in Eq.~\ref{eq:fmaps}, and $\lambda_{\mathrm{couple}} = 1.0$ in Eq.~\ref{eq:total_loss}. For training and test-time adaptation, we use the Adam optimiser~\cite{kingma2015adam} with learning rate equal to $10^{-3}$. For test-time adaptation, we choose the number of iterations equal to 15. In the context of non-isometric matching, the loss weight for the smoothness penalty term $L_{\mathrm{dirichlet}}$ is empirically set to be $5.0$. We emphasise that throughout all experiments we use the same set of parameters.

\section{Experimental Results}
\label{sec:experiment}
In this section we evaluate  our method and compare it to previous approaches in a broad range of different challenging scenarios.
We refer the reader to our supplementary material, where we provide our complete shape matching results on all datasets, as well as our video, which shows additional animated results.

\subsection{Near-Isometric Shape Matching}
\label{subsec:near-isometric}

\textbf{Datasets.} We evaluate our method on three common near-isometric benchmark datasets: FAUST~\cite{bogo2014faust}, SCAPE \cite{anguelov2005scape} and SHREC'19~\cite{melzi2019shrec}. Instead of the original datasets, we choose the more challenging remeshed versions from~\cite{ren2018continuous,donati2020deep}. The FAUST dataset consists of 100 human shapes (10 persons in 10 poses), {where the training and testing split is 80/20}. The SCAPE dataset contains 71 different poses of the same person, {which is split into 51 and 20 shapes for training and testing}. The SHREC'19 dataset is an even more challenging dataset with 44 human shapes and is used only as a test set.

\textbf{Baselines.} We perform an extensive comparison with existing non-rigid shape matching methods that are categorised as follows: 
\begin{enumerate}
    \item \textbf{Axiomatic approaches}, including BCICP~\cite{ren2018continuous}, ZoomOut~\cite{melzi2019zoomout}, Smooth Shells~\cite{eisenberger2020smooth}, and DiscreteOp~\cite{ren2021discrete};
    \item \textbf{Supervised approaches}, including FMNet~\cite{litany2017deep}, 3D-CODED~\cite{groueix20183d}, HSN~\cite{wiersma2020cnns}, ACSCNN~\cite{li2020shape}, GeomFMaps~\cite{donati2020deep}, and TransMatch~\cite{trappolini2021shape}; and
    \item \textbf{Unsupervised approaches}, including UnsupFMNet~\cite{halimi2019unsupervised}, SURFMNet~\cite{roufosse2019unsupervised}, WSupFMNet~\cite{sharma2020weakly}, Deep Shells~\cite{eisenberger2020deep}, NeuroMorph~\cite{eisenberger2021neuromorph}, ConsistFMaps~\cite{cao2022unsupervised}, DUO-FMNet~\cite{donati2022deep}, and AttentiveFMaps~\cite{li2022learning}.
\end{enumerate}
3D-CODED and TransMatch are template-based methods that require a large amount of training data. Therefore, the networks pre-trained on the large SURREAL dataset~\cite{varol2017learning} are used. GeomFMaps and WSupFMNet are strong baselines related to our work. For a fair comparison, we use them with DiffusionNet as feature extractor, as it  significantly improves shape matching accuracy~\cite{sharp2020diffusionnet}. {As for axiomatic approaches, they are initialised as originally proposed by respective authors.}

\textbf{Results.} The mean geodesic error~\cite{kim2011blended} is used for method evaluation. The results are summarised in Tab.~\ref{tab:complete_shape}. Our method outperforms the previous state-of-the-art in most settings, even in comparison to the supervised methods without relying on any additional post-processing techniques. Moreover, our method demonstrates substantially better cross-dataset generalisation ability compared to existing learning-based methods. {Fig.~\ref{fig:iso_pck} shows the proportion of correct keypoints (PCK) curves of our method in comparison to the existing state-of-the-art method~\cite{li2022learning} for near-isometric shape matching.} Fig.~\ref{fig:near-isometric-qualitative} shows some qualitative results on the SHREC'19 dataset trained on FAUST and SCAPE datasets.

\begin{figure}[ht]
    \centering
    \def\heightNI{2.6cm}
\def\widthNI{1.8cm}
\def\hspaceColsNI{-0.4cm}
\def\pathNI{figures/shapePlots/}
\def\firstColNI{15-3}
\def\secondColNI{2-36}
\def\thirdColNI{21-30}
\def\fourthColNI{37-14}
\def\fifthColNI{23-10}%
\begin{tabular}{lccccc}%
    \setlength{\tabcolsep}{0pt} 
    \rotatedCentering{90}{\heightNI}{Source}&
    \hspace{\hspaceColsNI}
    \includegraphics[height=\heightNI, width=\widthNI]{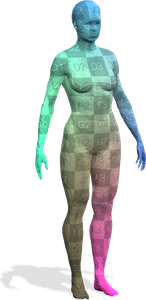}&
    \hspace{\hspaceColsNI}
    \includegraphics[height=\heightNI, width=\widthNI]{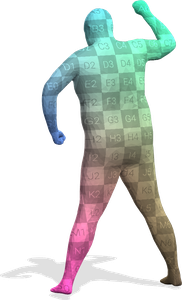}&
    \hspace{\hspaceColsNI}
    \includegraphics[height=\heightNI, width=\widthNI]{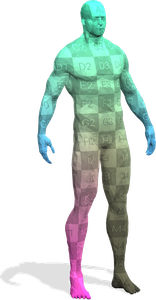}&
    \hspace{-0.8cm}
    \includegraphics[height=\heightNI, width=\widthNI]{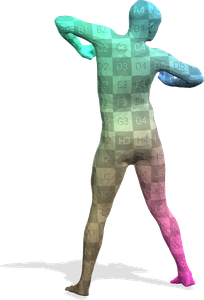}&
    \hspace{\hspaceColsNI}
    \includegraphics[height=\heightNI, width=\widthNI]{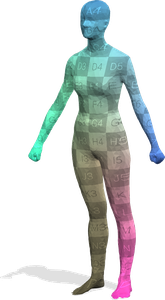}\\
    \rotatedCentering{90}{\heightNI}{Ours}&
    \hspace{\hspaceColsNI}
    \includegraphics[height=\heightNI, width=\widthNI]{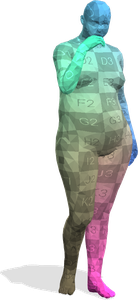}&
    \hspace{\hspaceColsNI}
    \includegraphics[height=\heightNI, width=\widthNI]{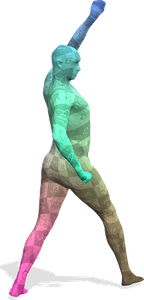}&
    \hspace{\hspaceColsNI}
    \includegraphics[height=\heightNI, width=\widthNI]{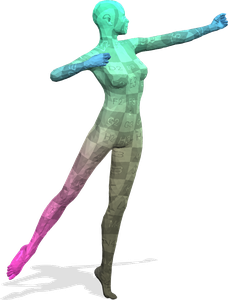}&
    \hspace{-0.8cm}
    \includegraphics[height=\heightNI, width=\widthNI]{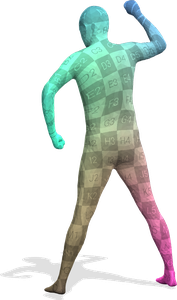}&
    \hspace{\hspaceColsNI}
    \includegraphics[height=\heightNI, width=\widthNI]{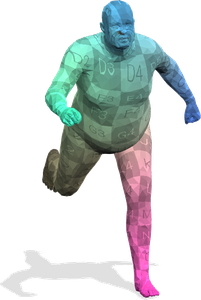}\\
\end{tabular}
    \caption{\textbf{Cross-dataset generalisation on SHREC'19 (trained on FAUST and SCAPE).} Our method demonstrates previously unseen cross-dataset generalisation ability.}
    \label{fig:near-isometric-qualitative}
\end{figure}

\subsection{Matching with Anisotropic Meshing}
\label{subsec:anisotropic}

\textbf{Datasets.} To evaluate the robustness to different meshing, we follow \citet{donati2022deep} and use an anisotropic remeshed version of the FAUST and SCAPE datasets  (denoted F\_a and S\_a, respectively). In this anisotropic remeshing the triangle scale is coordinate-dependent, so that methods that overfit  to mesh connectivity are  likely to fail to predict the correct maps. For a fair comparison, {all methods are trained on the original datasets and evaluated on the anisotropic remeshed ones}.

\textbf{Results.} Tab.~\ref{tab:anisotropic} summarises the matching performance. We observe that existing state-of-the-art methods~\cite{eisenberger2020deep,cao2022unsupervised} overfit to mesh connectivity and thus suffer from huge performance drops when testing on the anisotropic remeshed datasets. In contrast, our method is much more robust to varying mesh connectivity and outperforms the state of the art in most settings.

\begin{table}[h!]
\setlength{\tabcolsep}{4pt}
    \small
    \centering
    \caption{\textbf{Anisotropic meshing on FAUST, SCAPE and their remeshed versions.} Our method outperforms previous state-of-the-art methods in most settings and is much more robust to varying mesh connectivity.}
    \label{tab:anisotropic}
        \begin{tabular}{@{}lcccccccc@{}}
        \toprule
        \multicolumn{1}{l}{Train}  & \multicolumn{4}{c}{\textbf{FAUST}}   & \multicolumn{4}{c}{\textbf{SCAPE}}\\ \cmidrule(lr){2-5} \cmidrule(lr){6-9}
        \multicolumn{1}{l}{Test} & \multicolumn{1}{c}{\textbf{F}} & \multicolumn{1}{c}{\textbf{F\_a}} & \multicolumn{1}{c}{\textbf{S}} & \multicolumn{1}{c}{\textbf{S\_a}} & \multicolumn{1}{c}{\textbf{F}} & \multicolumn{1}{c}{\textbf{F\_a}} & \multicolumn{1}{c}{\textbf{S}} & \multicolumn{1}{c}{\textbf{S\_a}}
        \\ \midrule
        \multicolumn{9}{c}{Axiomatic Methods} \\
        \multicolumn{1}{l}{BCICP}  & \multicolumn{1}{c}{6.1}  & \multicolumn{1}{c}{8.5} & \multicolumn{1}{c}{-}  & \multicolumn{1}{c}{-} & \multicolumn{1}{c}{-}  & \multicolumn{1}{c}{-} & \multicolumn{1}{c}{11.0}  & \multicolumn{1}{c}{14.0}\\
        \multicolumn{1}{l}{ZoomOut} & \multicolumn{1}{c}{6.1}  & \multicolumn{1}{c}{8.7} & \multicolumn{1}{c}{-}  & \multicolumn{1}{c}{-} & \multicolumn{1}{c}{-}  & \multicolumn{1}{c}{-} & \multicolumn{1}{c}{7.5}  & \multicolumn{1}{c}{15.0}\\
        \multicolumn{1}{l}{Smooth Shells} & \multicolumn{1}{c}{2.5}  & \multicolumn{1}{c}{5.4} & \multicolumn{1}{c}{-}  & \multicolumn{1}{c}{-} & \multicolumn{1}{c}{-}  & \multicolumn{1}{c}{-} & \multicolumn{1}{c}{4.7}  & \multicolumn{1}{c}{5.0}\\ 
        \multicolumn{1}{l}{DiscreteOp} & \multicolumn{1}{c}{5.6}  & \multicolumn{1}{c}{6.2} & \multicolumn{1}{c}{-}  & \multicolumn{1}{c}{-} & \multicolumn{1}{c}{-}  & \multicolumn{1}{c}{-} & \multicolumn{1}{c}{13.1}  & \multicolumn{1}{c}{14.6}\\
        \midrule
        \multicolumn{9}{c}{Supervised Methods} \\ 
        \multicolumn{1}{l}{FMNet} & \multicolumn{1}{c}{11.0}  & \multicolumn{1}{c}{43.0} & \multicolumn{1}{c}{30.0}  & \multicolumn{1}{c}{44.0} & \multicolumn{1}{c}{33.0}  & \multicolumn{1}{c}{43.0} & \multicolumn{1}{c}{17.0}  & \multicolumn{1}{c}{41.0}\\
        \multicolumn{1}{l}{GeomFMaps} & \multicolumn{1}{c}{2.6}  & \multicolumn{1}{c}{3.2} & \multicolumn{1}{c}{3.4}  & \multicolumn{1}{c}{3.8} & \multicolumn{1}{c}{3.0}  & \multicolumn{1}{c}{8.4} & \multicolumn{1}{c}{3.0}  & \multicolumn{1}{c}{3.1}\\
        \midrule
        \multicolumn{9}{c}{Unsupervised Methods} \\
        \multicolumn{1}{l}{UnsupFMNet} & \multicolumn{1}{c}{10.0}  & \multicolumn{1}{c}{42.0} & \multicolumn{1}{c}{29.0}  & \multicolumn{1}{c}{43.0} & \multicolumn{1}{c}{22.0}  & \multicolumn{1}{c}{44.0} & \multicolumn{1}{c}{16.0}  & \multicolumn{1}{c}{44.0}\\
        \multicolumn{1}{l}{Deep Shells} & \multicolumn{1}{c}{1.7}  & \multicolumn{1}{c}{12.0} & \multicolumn{1}{c}{5.4}  & \multicolumn{1}{c}{16.0} & \multicolumn{1}{c}{2.7}  & \multicolumn{1}{c}{15.0} & \multicolumn{1}{c}{2.5}  & \multicolumn{1}{c}{10.0}\\
        \multicolumn{1}{l}{ConsistFMaps} & \multicolumn{1}{c}{\textbf{1.5}}  & \multicolumn{1}{c}{15.3} & \multicolumn{1}{c}{3.2}  & \multicolumn{1}{c}{14.1} & \multicolumn{1}{c}{3.2}  & \multicolumn{1}{c}{23.3} & \multicolumn{1}{c}{2.0}  & \multicolumn{1}{c}{4.9}\\
        \multicolumn{1}{l}{DUO-FMNet} & \multicolumn{1}{c}{2.5}  & \multicolumn{1}{c}{3.0} & \multicolumn{1}{c}{4.2}  & \multicolumn{1}{c}{4.4} & \multicolumn{1}{c}{2.7}  & \multicolumn{1}{c}{3.1} & \multicolumn{1}{c}{2.6}  & \multicolumn{1}{c}{2.7}\\
        \multicolumn{1}{l}{AttentiveFMaps} & \multicolumn{1}{c}{1.9}  & \multicolumn{1}{c}{2.4} & \multicolumn{1}{c}{2.6}  & \multicolumn{1}{c}{2.8} & \multicolumn{1}{c}{2.2}  & \multicolumn{1}{c}{2.5} & \multicolumn{1}{c}{2.2}  & \multicolumn{1}{c}{2.3}\\
        \multicolumn{1}{l}{AttentiveFMaps-Fast} & \multicolumn{1}{c}{1.9}  & \multicolumn{1}{c}{2.3} & \multicolumn{1}{c}{2.6}  & \multicolumn{1}{c}{2.8} & \multicolumn{1}{c}{1.9}  & \multicolumn{1}{c}{2.4} & \multicolumn{1}{c}{2.1}  & \multicolumn{1}{c}{2.2}\\
        \multicolumn{1}{l}{Ours} & \multicolumn{1}{c}{1.6}  & \multicolumn{1}{c}{\textbf{1.9}} & \multicolumn{1}{c}{\textbf{2.2}}  & \multicolumn{1}{c}{\textbf{2.4}} & \multicolumn{1}{c}{\textbf{1.6}}  & \multicolumn{1}{c}{\textbf{2.1}} & \multicolumn{1}{c}{\textbf{1.9}}  & \multicolumn{1}{c}{\textbf{1.9}}\\
        \hline
    \end{tabular}
\end{table}

\subsection{Matching with Topological Noise}
\textbf{Datasets.} When working with data acquired from real scans, the mesh topology is often corrupted due to self-intersections of separate parts of the scanned objects. Such topological noise presents an additional challenge to shape matching methods as it distorts the intrinsic shape geometry non-isometrically. To evaluate our method's robustness to topological noise, the TOPKIDS dataset~\cite{lahner2016shrec} is used, which contains synthetic shapes of children with topological merging based on the outer hull of intersecting shape parts. Due to the small amount of training data (26 shapes), in our comparison we consider only axiomatic and unsupervised methods. {The ground truth correspondences are provided by matching the other shapes to one selected reference shape.}

\textbf{Results.} The matching results are summarised in Tab.~\ref{tab:topkids}. Our method achieves the best matching performance and is much more robust against topological noise, in particular compared to existing methods based on functional maps. As shown in {both Fig.~\ref{fig:pck_nonisoright_topkidsmiddle_partialright}~(middle) and} Fig.~\ref{fig:topo-noise}, due to the topological degradation, none of the existing functional map methods provide satisfactory matching results (see the column \emph{Fully intrinsic} in Tab.~\ref{tab:topkids}). In comparison to Deep Shells and NeuroMorph, our method does not rely on shape alignment in the spatial domain and thus does not require shape rigid alignment as pre-processing.

\begin{table}[h!t!]
\setlength{\tabcolsep}{14pt}
    \centering
    \small
    \caption{\textbf{Topological noise on TOPKIDS.} Our method is more robust to topological noise compared to existing methods.}
    \label{tab:topkids}
    \begin{tabular}{@{}lcc@{}}
    \toprule
    \multicolumn{1}{c}{\textbf{Geo. error ($\times$100)}}      & \multicolumn{1}{c}{\textbf{TOPKIDS}} & \multicolumn{1}{c}{\textbf{Fully intrinsic}}
    \\ \midrule
    \multicolumn{3}{c}{Axiomatic Methods} \\
    \multicolumn{1}{l}{ZoomOut}  & \multicolumn{1}{c}{33.7} & \multicolumn{1}{c}{\cmark}  \\ 
    \multicolumn{1}{l}{Smooth Shells}  & \multicolumn{1}{c}{11.8} & \multicolumn{1}{c}{\xmark}\\
    \multicolumn{1}{l}{DiscreteOp}  & \multicolumn{1}{c}{35.5} & \multicolumn{1}{c}{\cmark}
    \\\midrule
    \multicolumn{3}{c}{Unsupervised Methods} \\
    \multicolumn{1}{l}{UnsupFMNet}  & \multicolumn{1}{c}{38.5} & \multicolumn{1}{c}{\cmark}\\
    \multicolumn{1}{l}{SURFMNet}  & \multicolumn{1}{c}{48.6} & \multicolumn{1}{c}{\cmark}\\
    \multicolumn{1}{l}{WSupFMNet}  & \multicolumn{1}{c}{47.9} & \multicolumn{1}{c}{\cmark}\\
    \multicolumn{1}{l}{Deep Shells}  & \multicolumn{1}{c}{13.7} & \multicolumn{1}{c}{\xmark} \\
    \multicolumn{1}{l}{NeuroMorph}  & \multicolumn{1}{c}{13.8} & \multicolumn{1}{c}{\xmark} \\
    \multicolumn{1}{l}{ConsistFMaps}  & \multicolumn{1}{c}{39.3} & \multicolumn{1}{c}{\cmark} \\
    \multicolumn{1}{l}{AttentiveFMaps}  & \multicolumn{1}{c}{23.4} & \multicolumn{1}{c}{\cmark} \\
    \multicolumn{1}{l}{AttentiveFMaps-Fast}  & \multicolumn{1}{c}{28.5} & \multicolumn{1}{c}{\cmark}\\
    \multicolumn{1}{l}{Ours}  & \multicolumn{1}{c}{\textbf{9.2}} & \multicolumn{1}{c}{\cmark} \\ \hline
    \end{tabular} 
\end{table}

\begin{figure*}[ht]
    \centering
    \begin{tabular}{cc|c|c}
    \hspace{-1cm}
     \newcommand{\pckLineWidth}{2pt}
\newcommand{\plotWidth}{\columnwidth}
\newcommand{\plotHeight}{0.7\columnwidth}
\newcommand{\pckTitle}{\textbf{SMAL}}
\definecolor{cPLOT0}{RGB}{28,213,227}
\definecolor{cPLOT1}{RGB}{80,150,80}
\definecolor{cPLOT2}{RGB}{90,130,213}
\definecolor{cPLOT3}{RGB}{247,179,43}
\definecolor{cPLOT5}{RGB}{242,64,0}

\pgfplotsset{%
    label style = {font=\large},
    tick label style = {font=\large},
    title style =  {font=\LARGE},
    legend style={  fill= gray!10,
                    fill opacity=0.6, 
                    font=\large,
                    draw=gray!20, %
                    text opacity=1}
}
\begin{tikzpicture}[scale=0.55, transform shape]
	\begin{axis}[
		width=\plotWidth,
		height=\plotHeight,
		grid=major,
		title=\pckTitle,
		legend style={
			at={(0.97,0.03)},
			anchor=south east,
			legend columns=1},
		legend cell align={left},
        xlabel={\LARGE$\%$ Geodesic Error},
		xmin=0,
        xmax=0.2,
        ylabel near ticks,
        xtick={0, 0.05, 0.1, 0.15, 0.2},
	ymin=0,
        ymax=1,
        ytick={0, 0.20, 0.40, 0.60, 0.80, 1.0}
	]   

\addplot [color=cPLOT0, smooth, line width=\pckLineWidth]
table[row sep=crcr]{%
0.0 0.06488340934619913 \\
0.010526315789473684 0.13704104694753083 \\
0.021052631578947368 0.3482007280994979 \\
0.031578947368421054 0.5091134238269884 \\
0.042105263157894736 0.6313163984788405 \\
0.05263157894736842 0.7207291821737424 \\
0.06315789473684211 0.788068912316791 \\
0.07368421052631578 0.8386001001475146 \\
0.08421052631578947 0.877629210593983 \\
0.09473684210526316 0.9058721630509805 \\
0.10526315789473684 0.9262048152007687 \\
0.11578947368421053 0.9395975152589625 \\
0.12631578947368421 0.9487271792234507 \\
0.1368421052631579 0.9551447402254672 \\
0.14736842105263157 0.9603266974327049 \\
0.15789473684210525 0.9645085328389114 \\
0.16842105263157894 0.9678350543367934 \\
0.17894736842105263 0.9708245929815539 \\
0.18947368421052632 0.9734906822211095 \\
0.2 0.9757561814023359 \\
    };
\addlegendentry{\textcolor{black}{AttentiveFMaps: 0.78}}  

\addplot [color=cPLOT5, smooth, line width=\pckLineWidth]
table[row sep=crcr]{%
0.0 0.06688229960346997 \\
0.010526315789473684 0.13851348608084882 \\
0.021052631578947368 0.34888687390886575 \\
0.031578947368421054 0.509796862946773 \\
0.042105263157894736 0.6397829234954189 \\
0.05263157894736842 0.7352979388558823 \\
0.06315789473684211 0.8039084597582926 \\
0.07368421052631578 0.8531255497963216 \\
0.08421052631578947 0.8885195761324113 \\
0.09473684210526316 0.9149734067748441 \\
0.10526315789473684 0.9334655100079847 \\
0.11578947368421053 0.9458540282307724 \\
0.12631578947368421 0.9541554451827692 \\
0.1368421052631579 0.9602630902274972 \\
0.14736842105263157 0.964895589449324 \\
0.15789473684210525 0.9682559445669974 \\
0.16842105263157894 0.9704605432325993 \\
0.17894736842105263 0.9721305707054987 \\
0.18947368421052632 0.9734162482575686 \\
0.2 0.9743432894398506 \\
    };
\addlegendentry{\textcolor{black}{AttentiveFMaps-Fast: 0.79}} 

\addplot [color=cPLOT1, smooth, line width=\pckLineWidth]
table[row sep=crcr]{%
0.0 0.11681666238107483 \\
0.010526315789473684 0.2009649010028285 \\
0.021052631578947368 0.4125514271020828 \\
0.031578947368421054 0.5778947030084854 \\
0.042105263157894736 0.6992298791463101 \\
0.05263157894736842 0.7843635896117254 \\
0.06315789473684211 0.8472923630753407 \\
0.07368421052631578 0.893088840318848 \\
0.08421052631578947 0.9245872975057855 \\
0.09473684210526316 0.944853432759064 \\
0.10526315789473684 0.9571213679609154 \\
0.11578947368421053 0.9645358704037027 \\
0.12631578947368421 0.9692600925687838 \\
0.1368421052631579 0.9726851375674981 \\
0.14736842105263157 0.9756203394188737 \\
0.15789473684210525 0.9781679094883003 \\
0.16842105263157894 0.9803272049370018 \\
0.17894736842105263 0.9822512213936745 \\
0.18947368421052632 0.9838518899460016 \\
0.2 0.98525777834919 \\
    };
\addlegendentry{\textcolor{black}{Ours: \textbf{0.82}}}  
\end{axis}
\end{tikzpicture}&
     \hspace{-1cm}
     \newcommand{\pckLineWidth}{2pt}
\newcommand{\plotWidth}{\columnwidth}
\newcommand{\plotHeight}{0.7\columnwidth}
\newcommand{\pckTitle}{\textbf{DT4D-H inter}}
\definecolor{cPLOT0}{RGB}{28,213,227}
\definecolor{cPLOT1}{RGB}{80,150,80}
\definecolor{cPLOT2}{RGB}{90,130,213}
\definecolor{cPLOT3}{RGB}{247,179,43}
\definecolor{cPLOT5}{RGB}{242,64,0}

\pgfplotsset{%
    label style = {font=\large},
    tick label style = {font=\large},
    title style =  {font=\LARGE},
    legend style={  fill= gray!10,
                    fill opacity=0.6, 
                    font=\large,
                    draw=gray!20, %
                    text opacity=1}
}
\begin{tikzpicture}[scale=0.55, transform shape]
	\begin{axis}[
		width=\plotWidth,
		height=\plotHeight,
		grid=major,
		title=\pckTitle,
		legend style={
			at={(0.97,0.03)},
			anchor=south east,
			legend columns=1},
		legend cell align={left},
        xlabel={\LARGE$\%$ Geodesic Error},
		xmin=0,
        xmax=0.2,
        ylabel near ticks,
        xtick={0, 0.05, 0.1, 0.15, 0.2},
	ymin=0,
        ymax=1,
        ytick={0, 0.20, 0.40, 0.60, 0.80, 1.0}
	]   

\addplot [color=cPLOT0, smooth, line width=\pckLineWidth]
table[row sep=crcr]{%
0.0 0.03456135231465077 \\
0.010526315789473684 0.08081747504012339 \\
0.021052631578947368 0.26366498530545884 \\
0.031578947368421054 0.4222855743377035 \\
0.042105263157894736 0.5487909831794402 \\
0.05263157894736842 0.6365265648123059 \\
0.06315789473684211 0.6959819288409029 \\
0.07368421052631578 0.7381741251016112 \\
0.08421052631578947 0.7694500489818038 \\
0.09473684210526316 0.7939294870458762 \\
0.10526315789473684 0.8120624465889905 \\
0.11578947368421053 0.8260296600454384 \\
0.12631578947368421 0.8372674406486441 \\
0.1368421052631579 0.8469622944327491 \\
0.14736842105263157 0.8553124413781603 \\
0.15789473684210525 0.8624852533505638 \\
0.16842105263157894 0.8689187527356859 \\
0.17894736842105263 0.8752861787940055 \\
0.18947368421052632 0.8819866811180357 \\
0.2 0.8887820205515142 \\
    };
\addlegendentry{\textcolor{black}{AttentiveFMaps: 0.68}}  

\addplot [color=cPLOT5, smooth, line width=\pckLineWidth]
table[row sep=crcr]{%
0.0 0.033933134627008775 \\
0.010526315789473684 0.0792694416074369 \\
0.021052631578947368 0.25661285616024343 \\
0.031578947368421054 0.4117703274485691 \\
0.042105263157894736 0.5368621422765075 \\
0.05263157894736842 0.6242908060112137 \\
0.06315789473684211 0.6831338974925485 \\
0.07368421052631578 0.7232717760593618 \\
0.08421052631578947 0.7514663276153156 \\
0.09473684210526316 0.7721072597286199 \\
0.10526315789473684 0.7866159826583571 \\
0.11578947368421053 0.7968690205723575 \\
0.12631578947368421 0.8043884569689643 \\
0.1368421052631579 0.8105727744544261 \\
0.14736842105263157 0.8161636825979115 \\
0.15789473684210525 0.8214432957458783 \\
0.16842105263157894 0.8263249056839735 \\
0.17894736842105263 0.8313235508681243 \\
0.18947368421052632 0.8369647956312399 \\
0.2 0.8432126852450132 \\
    };
\addlegendentry{\textcolor{black}{AttentiveFMaps-Fast: 0.66}} 

\addplot [color=cPLOT1, smooth, line width=\pckLineWidth]
table[row sep=crcr]{%
0.0 0.050371427975905124 \\
0.010526315789473684 0.12944817308293557 \\
0.021052631578947368 0.3864891510515455 \\
0.031578947368421054 0.5803202576234445 \\
0.042105263157894736 0.7183634866706964 \\
0.05263157894736842 0.8076292806970007 \\
0.06315789473684211 0.8629943514600746 \\
0.07368421052631578 0.897993205077433 \\
0.08421052631578947 0.9198287721199742 \\
0.09473684210526316 0.9337159472247119 \\
0.10526315789473684 0.942982262334035 \\
0.11578947368421053 0.9494709965191654 \\
0.12631578947368421 0.9543339516851825 \\
0.1368421052631579 0.95824395022615 \\
0.14736842105263157 0.9615212497655127 \\
0.15789473684210525 0.9646511870271172 \\
0.16842105263157894 0.9679835129332806 \\
0.17894736842105263 0.9716687162598745 \\
0.18947368421052632 0.9755318173291369 \\
0.2 0.9789688809221085 \\
    };
\addlegendentry{\textcolor{black}{Ours: \textbf{0.81}}}  
\end{axis}
\end{tikzpicture}&
     \hspace{-1cm}
     \newcommand{\pckLineWidth}{2pt}
\newcommand{\plotWidth}{\columnwidth}
\newcommand{\plotHeight}{0.7\columnwidth}
\newcommand{\pckTitle}{\textbf{TOPKIDS}}
\definecolor{cPLOT0}{RGB}{28,213,227}
\definecolor{cPLOT1}{RGB}{80,150,80}
\definecolor{cPLOT2}{RGB}{90,130,213}
\definecolor{cPLOT3}{RGB}{247,179,43}
\definecolor{cPLOT5}{RGB}{242,64,0}

\pgfplotsset{%
    label style = {font=\large},
    tick label style = {font=\large},
    title style =  {font=\LARGE},
    legend style={  fill= gray!10,
                    fill opacity=0.6, 
                    font=\large,
                    draw=gray!20, %
                    text opacity=1}
}
\begin{tikzpicture}[scale=0.55, transform shape]
	\begin{axis}[
		width=\plotWidth,
		height=\plotHeight,
		grid=major,
		title=\pckTitle,
		legend style={
			at={(0.97,0.03)},
			anchor=south east,
			legend columns=1},
		legend cell align={left},
        xlabel={\LARGE$\%$ Geodesic Error},
		xmin=0,
        xmax=0.2,
        ylabel near ticks,
        xtick={0, 0.05, 0.1, 0.15, 0.2},
	ymin=0,
        ymax=1,
        ytick={0, 0.20, 0.40, 0.60, 0.80, 1.0}
	]
\addplot [color=cPLOT0, smooth, line width=\pckLineWidth]
table[row sep=crcr]{%
0.0 0.06531700633935275 \\
0.010526315789473684 0.16553915459815935 \\
0.021052631578947368 0.2928293328585914 \\
0.031578947368421054 0.37656451974952204 \\
0.042105263157894736 0.4298994527567283 \\
0.05263157894736842 0.46736665299203517 \\
0.06315789473684211 0.4941521599467463 \\
0.07368421052631578 0.5145363912905498 \\
0.08421052631578947 0.5312052510584939 \\
0.09473684210526316 0.5456061860936738 \\
0.10526315789473684 0.5583274635622673 \\
0.11578947368421053 0.5697406206218604 \\
0.12631578947368421 0.5801591417491659 \\
0.1368421052631579 0.5905737926977468 \\
0.14736842105263157 0.5999860673565905 \\
0.15789473684210525 0.6084191867980463 \\
0.16842105263157894 0.6156177192262738 \\
0.17894736842105263 0.6220847878755041 \\
0.18947368421052632 0.6285015442013112 \\
0.2 0.6345622440844318 \\
    };
\addlegendentry{\textcolor{black}{AttentiveFMaps: 0.50}} 

\addplot [color=cPLOT5, smooth, line width=\pckLineWidth]
table[row sep=crcr]{%
0.0 0.058250059987770234 \\
0.010526315789473684 0.15188516405687616 \\
0.021052631578947368 0.25884142329692783 \\
0.031578947368421054 0.3198199592857198 \\
0.042105263157894736 0.3573645630955237 \\
0.05263157894736842 0.38468802489298953 \\
0.06315789473684211 0.40502194391336993 \\
0.07368421052631578 0.4204329955957366 \\
0.08421052631578947 0.4346404216946739 \\
0.09473684210526316 0.4478919136485723 \\
0.10526315789473684 0.4598972080530679 \\
0.11578947368421053 0.4700680377419829 \\
0.12631578947368421 0.4795538457965989 \\
0.1368421052631579 0.48872229919577687 \\
0.14736842105263157 0.497252173105354 \\
0.15789473684210525 0.5059407243426501 \\
0.16842105263157894 0.5141106716308159 \\
0.17894736842105263 0.5223348014211296 \\
0.18947368421052632 0.5300519377984876 \\
0.2 0.5377148916736975 \\
    };
\addlegendentry{\textcolor{black}{AttentiveFMaps-Fast: 0.41}}

\addplot [color=cPLOT1, smooth, line width=\pckLineWidth]
table[row sep=crcr]{%
0.0 0.1457277097056342 \\
0.010526315789473684 0.30846098472827477 \\
0.021052631578947368 0.48728259271013136 \\
0.031578947368421054 0.609193222543017 \\
0.042105263157894736 0.6856911752184716 \\
0.05263157894736842 0.737915366931645 \\
0.06315789473684211 0.7737454815663387 \\
0.07368421052631578 0.7998188756356769 \\
0.08421052631578947 0.8187479197789354 \\
0.09473684210526316 0.832680563188408 \\
0.10526315789473684 0.843435789864776 \\
0.11578947368421053 0.8532583034684542 \\
0.12631578947368421 0.8613972893268211 \\
0.1368421052631579 0.8683016881719597 \\
0.14736842105263157 0.8742308019784354 \\
0.15789473684210525 0.8792426834271206 \\
0.16842105263157894 0.8829967567902286 \\
0.17894736842105263 0.8861354717360848 \\
0.18947368421052632 0.8891309900691214 \\
0.2 0.8914879289125572 \\
    };
\addlegendentry{\textcolor{black}{Ours: \textbf{0.76}}}

\end{axis}
\end{tikzpicture}&
     \hspace{-1cm}
     \newcommand{\pckLineWidth}{2pt}
\newcommand{\plotWidth}{\columnwidth}
\newcommand{\plotHeight}{0.7\columnwidth}
\newcommand{\pckTitle}{\textbf{SHREC'16}}
\definecolor{cPLOT0}{RGB}{28,213,227}
\definecolor{cPLOT1}{RGB}{80,150,80}
\definecolor{cPLOT2}{RGB}{90,130,213}
\definecolor{cPLOT3}{RGB}{247,179,43}
\definecolor{cPLOT5}{RGB}{242,64,0}

\pgfplotsset{%
    label style = {font=\large},
    tick label style = {font=\large},
    title style =  {font=\LARGE},
    legend style={  fill= gray!10,
                    fill opacity=0.6, 
                    font=\large,
                    draw=gray!20, %
                    text opacity=1}
}
\begin{tikzpicture}[scale=0.55, transform shape]
	\begin{axis}[
		width=\plotWidth,
		height=\plotHeight,
		grid=major,
		title=\pckTitle,
		legend style={
			at={(0.97,0.03)},
			anchor=south east,
			legend columns=1},
		legend cell align={left},
        xlabel={\LARGE$\%$ Geodesic Error},
		xmin=0,
        xmax=0.2,
        ylabel near ticks,
        xtick={0, 0.05, 0.1, 0.15, 0.2},
	ymin=0,
        ymax=1,
        ytick={0, 0.20, 0.40, 0.60, 0.80, 1.0}
	]   

\addplot [color=cPLOT0, smooth, line width=\pckLineWidth]
table[row sep=crcr]{%
0.0 0.09820953899901268 \\
0.010526315789473684 0.2295535239614187 \\
0.021052631578947368 0.4490357047922837 \\
0.031578947368421054 0.6150508847877268 \\
0.042105263157894736 0.7275171831092884 \\
0.05263157894736842 0.8033685064935064 \\
0.06315789473684211 0.8551836504139135 \\
0.07368421052631578 0.8906309333940913 \\
0.08421052631578947 0.9150563435102909 \\
0.09473684210526316 0.931423017771702 \\
0.10526315789473684 0.9422823156375788 \\
0.11578947368421053 0.949004851143009 \\
0.12631578947368421 0.9532175609478241 \\
0.1368421052631579 0.9562068048910154 \\
0.14736842105263157 0.9580342902711324 \\
0.15789473684210525 0.9593598105111263 \\
0.16842105263157894 0.9602723665223665 \\
0.17894736842105263 0.9610579479000532 \\
0.18947368421052632 0.9617877553732816 \\
0.2 0.9624594155844156 \\
    };
\addlegendentry{\textcolor{black}{DPFM (CUTS): 0.82}}

\addplot [color=cPLOT0, smooth, dashed, line width=\pckLineWidth]
table[row sep=crcr]{%
0.0 0.025942907493282673 \\
0.010526315789473684 0.07528458314250837 \\
0.021052631578947368 0.19621478735278905 \\
0.031578947368421054 0.32684963805656314 \\
0.042105263157894736 0.4429562515998823 \\
0.05263157894736842 0.5427140272454511 \\
0.06315789473684211 0.6240629260910979 \\
0.07368421052631578 0.6905988740311462 \\
0.08421052631578947 0.7444385044431017 \\
0.09473684210526316 0.7867607237823916 \\
0.10526315789473684 0.8196455966886137 \\
0.11578947368421053 0.845340359174661 \\
0.12631578947368421 0.8644849641539301 \\
0.1368421052631579 0.879182984740388 \\
0.14736842105263157 0.8888336488112548 \\
0.15789473684210525 0.8954491076357266 \\
0.16842105263157894 0.8999575106092874 \\
0.17894736842105263 0.9036222205582479 \\
0.18947368421052632 0.9066243398036781 \\
0.2 0.9091440999824122 \\
    };
\addlegendentry{\textcolor{black}{DPFM (HOLES): 0.67}}

\addplot [color=cPLOT5, smooth, line width=\pckLineWidth]
table[row sep=crcr]{%
0.0 0.06289279068884332 \\
0.010526315789473684 0.16662750626566417 \\
0.021052631578947368 0.3553011316169211 \\
0.031578947368421054 0.5125396350725299 \\
0.042105263157894736 0.6261522651325283 \\
0.05263157894736842 0.7080046233006759 \\
0.06315789473684211 0.76742993848257 \\
0.07368421052631578 0.8098086124401914 \\
0.08421052631578947 0.8398517600820232 \\
0.09473684210526316 0.8611206045416572 \\
0.10526315789473684 0.8762032923217133 \\
0.11578947368421053 0.8865950387331967 \\
0.12631578947368421 0.8941529961266803 \\
0.1368421052631579 0.899795653907496 \\
0.14736842105263157 0.9039063093339409 \\
0.15789473684210525 0.9070533815599605 \\
0.16842105263157894 0.9095145534290271 \\
0.17894736842105263 0.9117609364319891 \\
0.18947368421052632 0.91373556998557 \\
0.2 0.9156437495253285 \\
    };
\addlegendentry{\textcolor{black}{DPFM-unsup (CUTS): 0.75}} 

\addplot [color=cPLOT5, smooth, dashed, line width=\pckLineWidth]
table[row sep=crcr]{%
0.0 0.016573474430616044 \\
0.010526315789473684 0.05070864989961011 \\
0.021052631578947368 0.14013575708606366 \\
0.031578947368421054 0.2422644059060253 \\
0.042105263157894736 0.3385394446079396 \\
0.05263157894736842 0.42517862087099767 \\
0.06315789473684211 0.498719223488971 \\
0.07368421052631578 0.5608059749835005 \\
0.08421052631578947 0.6114232030383397 \\
0.09473684210526316 0.6514807726800184 \\
0.10526315789473684 0.6827609745394516 \\
0.11578947368421053 0.7071897621464804 \\
0.12631578947368421 0.7265468140793123 \\
0.1368421052631579 0.7420989759012017 \\
0.14736842105263157 0.7540839444085529 \\
0.15789473684210525 0.7635430579475186 \\
0.16842105263157894 0.7714993704952975 \\
0.17894736842105263 0.7782332423758535 \\
0.18947368421052632 0.784388109239527 \\
0.2 0.7896531368140968 \\
    };
\addlegendentry{\textcolor{black}{DPFM-unsup (HOLES): 0.56}}

\addplot [color=cPLOT1, smooth, line width=\pckLineWidth]
table[row sep=crcr]{%
0.0 0.30081240031897927 \\
0.010526315789473684 0.523318238778765 \\
0.021052631578947368 0.7392498765854029 \\
0.031578947368421054 0.8435755582137161 \\
0.042105263157894736 0.8925260594668489 \\
0.05263157894736842 0.919430441634389 \\
0.06315789473684211 0.9356464076858814 \\
0.07368421052631578 0.9457165641376167 \\
0.08421052631578947 0.9525720076706918 \\
0.09473684210526316 0.9573353364471785 \\
0.10526315789473684 0.9603566207184628 \\
0.11578947368421053 0.9624736557302347 \\
0.12631578947368421 0.9640732987772461 \\
0.1368421052631579 0.9652884578871421 \\
0.14736842105263157 0.9662745879851143 \\
0.15789473684210525 0.9669972753854332 \\
0.16842105263157894 0.9678516841345789 \\
0.17894736842105263 0.9686693058403585 \\
0.18947368421052632 0.9694786207944103 \\
0.2 0.9702392819169134 \\
    };
\addlegendentry{\textcolor{black}{Ours (CUTS): \textbf{0.90}}}  

\addplot [color=cPLOT1, smooth, dashed, line width=\pckLineWidth]
table[row sep=crcr]{%
0.0 0.13387466674560872 \\
0.010526315789473684 0.30691445179108456 \\
0.021052631578947368 0.53647557469513 \\
0.031578947368421054 0.6711887103599234 \\
0.042105263157894736 0.7441372476974756 \\
0.05263157894736842 0.7889208913716934 \\
0.06315789473684211 0.8188245414541472 \\
0.07368421052631578 0.8400239612301723 \\
0.08421052631578947 0.8551817030931929 \\
0.09473684210526316 0.8658946019318742 \\
0.10526315789473684 0.874153912593751 \\
0.11578947368421053 0.8804071667760827 \\
0.12631578947368421 0.8854623594497972 \\
0.1368421052631579 0.8897705050490979 \\
0.14736842105263157 0.8934534993670126 \\
0.15789473684210525 0.8964469117700836 \\
0.16842105263157894 0.8988787328409904 \\
0.17894736842105263 0.9011242274854117 \\
0.18947368421052632 0.9031198357541257 \\
0.2 0.9050370824416075 \\
    };
\addlegendentry{\textcolor{black}{Ours (HOLES): \textbf{0.79}}}  
\end{axis}
\end{tikzpicture}
    \end{tabular}
    \caption{{(Left) \textbf{Non-isometric matching on SMAL and DT4D-H.} (Middle) \textbf{Matching with topological noise on TOPKIDS.} (Right) \textbf{Partial shape matching on SHREC'16.} Our method substantially outperforms existing state-of-the-art methods.}}
    \label{fig:pck_nonisoright_topkidsmiddle_partialright}
\end{figure*}
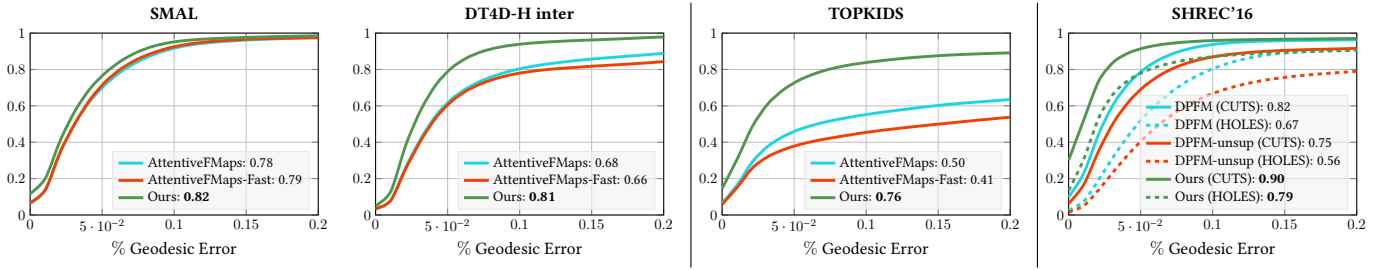

\begin{figure}[h!]
    \centering
    \footnotesize
    \def\columnOneT{kid00-kid03}
\def\columnTwoT{kid00-kid24}
\def\columnThreeT{kid00-kid22}
\def\columnFourT{kid00-kid21}
\def\columnFiveT{kid00-kid25}
\def\hspaceColsT{-0.3cm}
\def\heightT{2.3cm}
\def\widthT{2.0cm}
\begin{tabular}{lccccc}%
        \setlength{\tabcolsep}{0pt} 
        \rotatedCentering{90}{\heightT}{Source}&
        \multicolumn{5}{c}{\includegraphics[height=\heightT, width=\widthT]{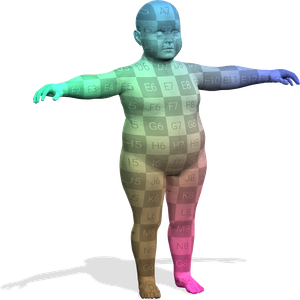}}\\
        \rotatedCentering{90}{\heightT}{DiscreteOp}&
        \hspace{\hspaceColsT}
        \includegraphics[height=\heightT, width=\widthT]{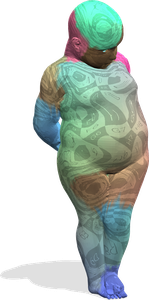}&
        \hspace{\hspaceColsT}
        \includegraphics[height=\heightT, width=\widthT]{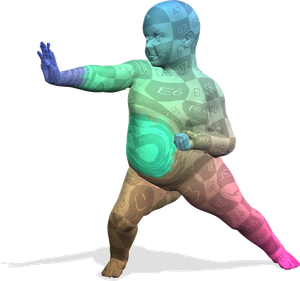}&
        \hspace{\hspaceColsT}
        \includegraphics[height=\heightT, width=\widthT]{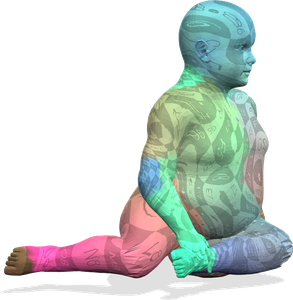}&
        \hspace{\hspaceColsT}
        \includegraphics[height=\heightT, width=\widthT]{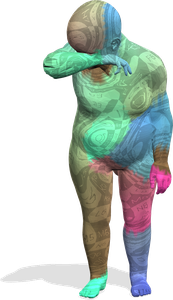}&
        \hspace{\hspaceColsT}
        \includegraphics[height=\heightT, width=\widthT]{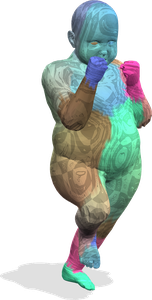}\\
        \rotatedCentering{90}{\heightT}{AttentiveFMaps}&
        \hspace{\hspaceColsT}
        \includegraphics[height=\heightT, width=\widthT]{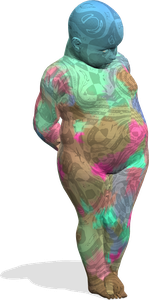}&
        \hspace{\hspaceColsT}
        \includegraphics[height=\heightT, width=\widthT]{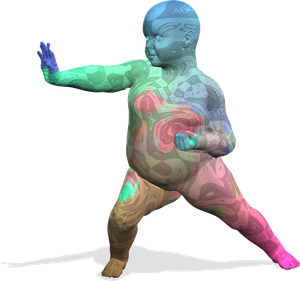}&
        \hspace{\hspaceColsT}
        \includegraphics[height=\heightT, width=\widthT]{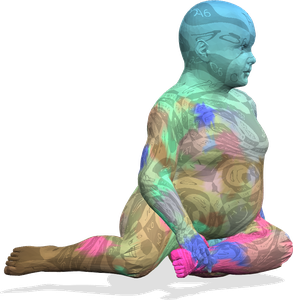}&
        \hspace{\hspaceColsT}
        \includegraphics[height=\heightT, width=\widthT]{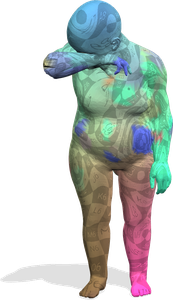}&
        \hspace{\hspaceColsT}
        \includegraphics[height=\heightT, width=\widthT]{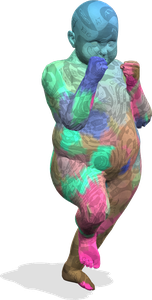}\\
        \rotatedCentering{90}{\heightT}{AttentiveFMaps-Fast}&
        \hspace{\hspaceColsT}
        \includegraphics[height=\heightT, width=\widthT]{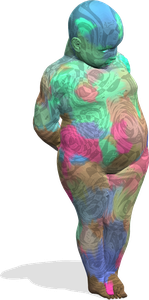}&
        \hspace{\hspaceColsT}
        \includegraphics[height=\heightT, width=\widthT]{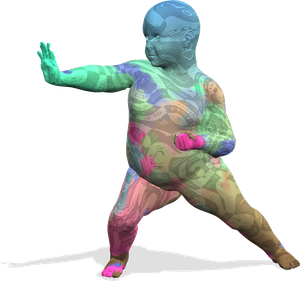}&
        \hspace{\hspaceColsT}
        \includegraphics[height=\heightT, width=\widthT]{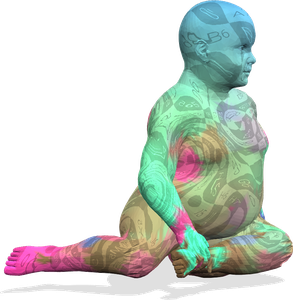}&
        \hspace{\hspaceColsT}
        \includegraphics[height=\heightT, width=\widthT]{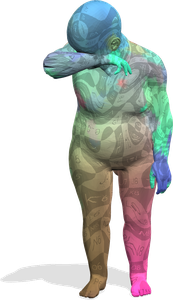}&
        \hspace{\hspaceColsT}
        \includegraphics[height=\heightT, width=\widthT]{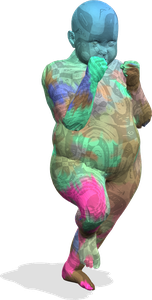}\\
        \rotatedCentering{90}{\heightT}{Ours}&
        \hspace{\hspaceColsT}
        \includegraphics[height=\heightT, width=\widthT]{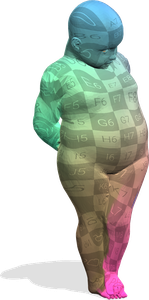}&
        \hspace{\hspaceColsT}
        \includegraphics[height=\heightT, width=\widthT]{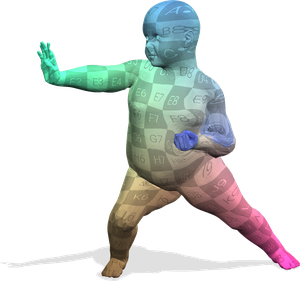}&
        \hspace{\hspaceColsT}
        \includegraphics[height=\heightT, width=\widthT]{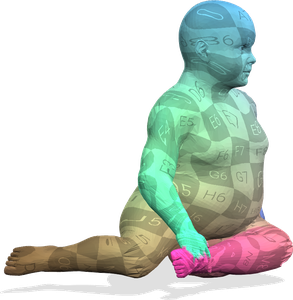}&
        \hspace{\hspaceColsT}
        \includegraphics[height=\heightT, width=\widthT]{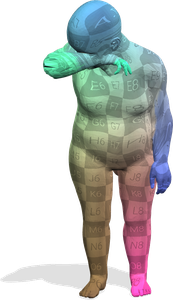}&
        \hspace{\hspaceColsT}
        \includegraphics[height=\heightT, width=\widthT]{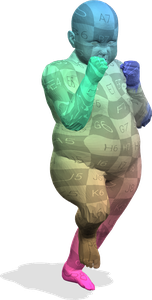}\\
    \end{tabular}
    \caption{\textbf{Topological noise on TOPKIDS.}  Ours is the first deep functional map method that properly handles matching under topological noise.}
    \label{fig:topo-noise}
\end{figure}

\subsection{Non-isometric Shape Matching}
\textbf{Datasets.} Regarding non-isometric shape matching, we first consider the SMAL~\cite{zuffi20173d} dataset for inter-class shape matching. The SMAL dataset contains 49 four-legged animal shapes of eight species. {Following~\citet{donati2022deep}, we use five species for training and three unseen species for testing, resulting in a 29/20 split of the dataset.} To this end, no shapes similar to the test data are seen during training, and the deformation across species is highly non-isometric, presenting great challenges to existing approaches~\cite{donati2022deep}. We further test on a recent non-isometric shape matching benchmark introduced by \citet{magnet2022smooth}, which is based on the large-scale animation dataset DeformingThings4D~\cite{li20214dcomplete}. Following \citet{donati2022deep}, nine classes of humanoid shapes are used for evaluation, {resulting in 198/95 shapes for training/testing.} The dataset is denoted as DT4D-H.

\begin{table}
    \setlength{\tabcolsep}{9pt}
    \small
    \centering
    \caption{\textbf{Non-isometric matching on SMAL and DT4D-H.} Our method sets to new state of the art on the SMAL dataset by a large margin. For DT4D-H inter-class matching, our method is the first unsupervised method that shows comparable performance  to the state-of-the-art supervised method.}
    \label{tab:non-isometry}
    \begin{tabular}{@{}lccc@{}}
    \toprule
    \multicolumn{1}{l}{\multirow{2}{*}{\textbf{Geo. error ($\times$100)}}}  & \multicolumn{1}{c}{\multirow{2}{*}{\textbf{SMAL}}}   & \multicolumn{2}{c}{\textbf{DT4D-H}}\\ \cmidrule(lr){3-4}
    &  & \multicolumn{1}{c}{\textbf{intra-class}} & \multicolumn{1}{c}{\textbf{inter-class}}
    \\ \midrule
    \multicolumn{4}{c}{Axiomatic Methods} \\
    \multicolumn{1}{l}{ZoomOut}  & \multicolumn{1}{c}{38.4} & \multicolumn{1}{c}{4.0} & \multicolumn{1}{c}{29.0} \\
    \multicolumn{1}{l}{Smooth Shells}  & \multicolumn{1}{c}{36.1} & \multicolumn{1}{c}{1.1} & \multicolumn{1}{c}{6.3} \\
    \multicolumn{1}{l}{DiscreteOp}  & \multicolumn{1}{c}{38.1} & \multicolumn{1}{c}{3.6} & \multicolumn{1}{c}{27.6} \\
    \midrule
    \multicolumn{4}{c}{Supervised Methods} \\ 
    \multicolumn{1}{l}{FMNet}  & \multicolumn{1}{c}{42.0} & \multicolumn{1}{c}{9.6} & \multicolumn{1}{c}{38.0} \\
    \multicolumn{1}{l}{GeomFMaps}  & \multicolumn{1}{c}{8.4} & \multicolumn{1}{c}{2.1} & \multicolumn{1}{c}{\textbf{4.1}} \\
    \midrule
    \multicolumn{4}{c}{Unsupervised Methods} \\
    \multicolumn{1}{l}{WSupFMNet}  & \multicolumn{1}{c}{7.6} & \multicolumn{1}{c}{3.3} & \multicolumn{1}{c}{22.6} \\
    \multicolumn{1}{l}{Deep Shells}  & \multicolumn{1}{c}{29.3} & \multicolumn{1}{c}{3.4} & \multicolumn{1}{c}{31.1} \\
    \multicolumn{1}{l}{DUO-FMNet}  & \multicolumn{1}{c}{6.7} & \multicolumn{1}{c}{2.6} & \multicolumn{1}{c}{15.8} \\
    \multicolumn{1}{l}{AttentiveFMaps}  & \multicolumn{1}{c}{5.4} & \multicolumn{1}{c}{1.7} & \multicolumn{1}{c}{11.6} \\
    \multicolumn{1}{l}{AttentiveFMaps-Fast}  & \multicolumn{1}{c}{5.8} & \multicolumn{1}{c}{1.2} & \multicolumn{1}{c}{14.6} \\
    \multicolumn{1}{l}{Ours}  & \multicolumn{1}{c}{\textbf{3.9}} & \multicolumn{1}{c}{\textbf{0.9}} & \multicolumn{1}{c}{\textbf{4.1}} \\
    \hline
    \end{tabular}
\end{table}

\begin{figure}[h!]
    \centering
    \def\heightIQ{2cm}
\def\widthIQ{1.8cm}
\def\hspaceColsIQ{-0.44cm}
\def\pathIQ{figures/shapePlots/}
\def\firstColIQ{bison-pig}
\def\secondColIQ{dog-elephant_a}
\def\thirdColIQ{giraffe_a-camel_a}
\def\fourthColIQ{elephant_a-bison}
\def\fifthColIQ{leopard-dog}
\begin{tabular}{lccccc}%
    \setlength{\tabcolsep}{0pt} 
    \rotatedCentering{90}{\heightIQ}{Source}&
    \hspace{\hspaceColsIQ}    \includegraphics[height=\heightIQ, width=\widthIQ]{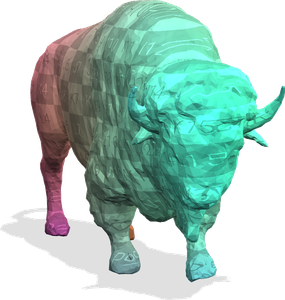}&
    \hspace{\hspaceColsIQ}
    \includegraphics[height=\heightIQ, width=\widthIQ]{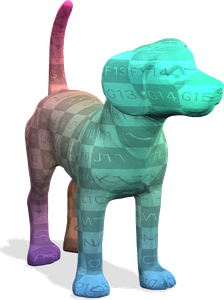}&
    \hspace{\hspaceColsIQ}
    \includegraphics[height=\heightIQ, width=\widthIQ]{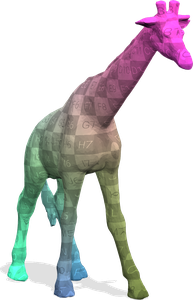}&
    \hspace{\hspaceColsIQ}
    \includegraphics[height=\heightIQ, width=\widthIQ]{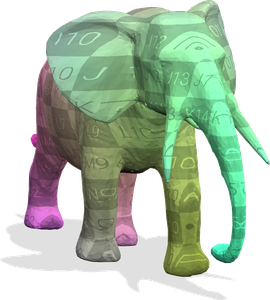}&
    \hspace{\hspaceColsIQ}
    \includegraphics[height=\heightIQ, width=\widthIQ]{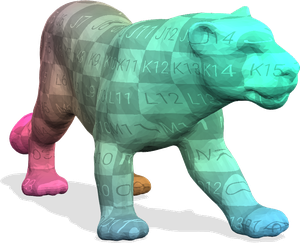}\\
    \rotatedCentering{90}{\heightIQ}{Ours}&
    \hspace{\hspaceColsIQ}
    \includegraphics[height=\heightIQ, width=\widthIQ]{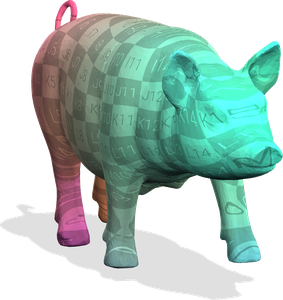}&
    \hspace{\hspaceColsIQ}
    \includegraphics[height=\heightIQ, width=\widthIQ]{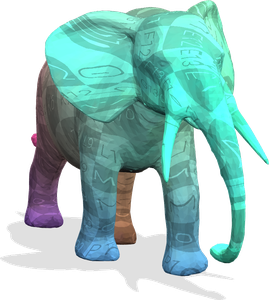}&
    \hspace{\hspaceColsIQ}
    \includegraphics[height=\heightIQ, width=\widthIQ]{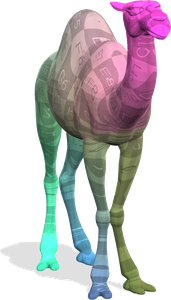}&
    \hspace{\hspaceColsIQ}
    \includegraphics[height=\heightIQ, width=\widthIQ]{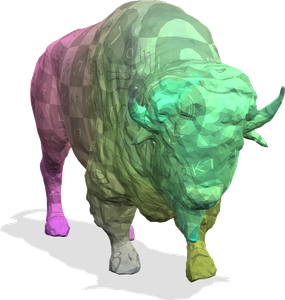}&
    \hspace{\hspaceColsIQ}
    \includegraphics[height=\heightIQ, width=\widthIQ]{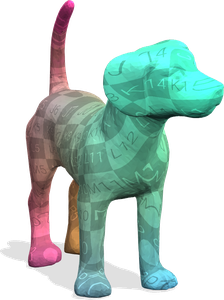}\\
\end{tabular}
    \caption{ 
    \textbf{Non-isometric matching on SHREC’20. } Our method obtains smooth and accurate matching results even in the presence of large non-isometric shape deformations.}
    \label{fig:non-isometric-qualitative}
\end{figure}
\begin{figure*}[h!]
    \centering
    \footnotesize
    \def\columnOne{Floating029-KettlebellSwing057}
\def\columnTwo{Standing2HMagicAttack01034-StandingReactLargeFromLeft021}
\def\columnThree{BrooklynUprock134-StandingReactLargeFromLeft016}
\def\columnFour{Standing2HMagicAttack01043-Falling257}
\def\columnFive{GoalkeeperScoop065-Standing2HMagicAttack01043}
\def\columnSix{StandingReactLargeFromLeft005-GoalkeeperScoop078}
\def\columnSeven{hippo_02-horse_02}
\def\columnEight{cougar_01-hippo_02}
\def\columnNine{hippo_03-horse_01}
\def\columnTen{Shuffling156-Shuffling188}
\def\heightQ{2cm}
\def\widthQ{1.8cm}
\def\hspaceCols{-0.27cm}
\begin{tabular}{lcccccccccc}%
        \setlength{\tabcolsep}{0pt} 
        \rotatedCentering{90}{\heightQ}{Source}&
        \hspace{\hspaceCols}
        \includegraphics[height=\heightQ, width=\widthQ]{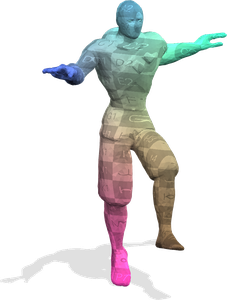}&
        \hspace{\hspaceCols}
        \includegraphics[height=\heightQ, width=\widthQ]{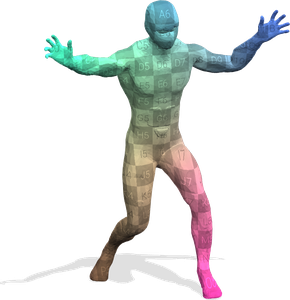}&
        \hspace{\hspaceCols}
        \includegraphics[height=\heightQ, width=\widthQ]{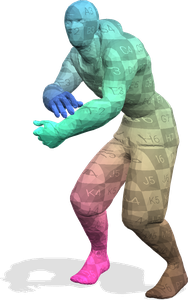}&
        \hspace{\hspaceCols}
        \includegraphics[height=\heightQ, width=\widthQ]{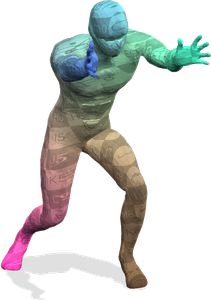}&
        \hspace{\hspaceCols}
        \includegraphics[height=\heightQ, width=\widthQ]{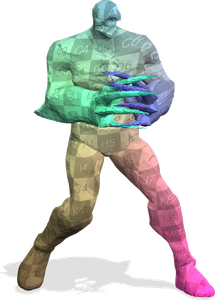}&
        \hspace{\hspaceCols}
        \includegraphics[height=\heightQ, width=\widthQ]{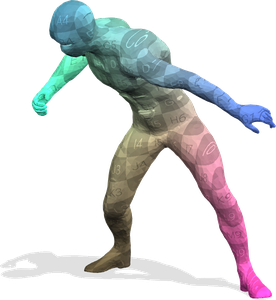}&
        \hspace{\hspaceCols}
        \includegraphics[height=\heightQ, width=\widthQ]{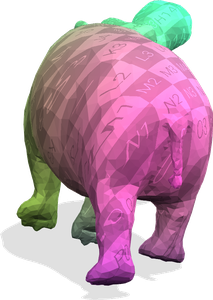}&
        \hspace{\hspaceCols}
        \includegraphics[height=\heightQ, width=\widthQ]{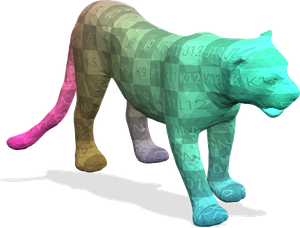}&
        \hspace{\hspaceCols}
        \includegraphics[height=\heightQ, width=\widthQ]{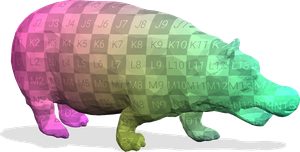}&
        \hspace{\hspaceCols}
        \includegraphics[height=\heightQ, width=\widthQ]{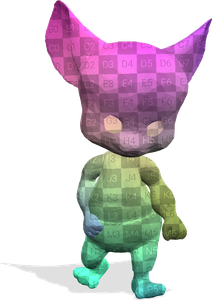}\\
        \rotatedCentering{90}{\heightQ}{DiscreteOp}&
        \hspace{\hspaceCols}
        \includegraphics[height=\heightQ, width=\widthQ]{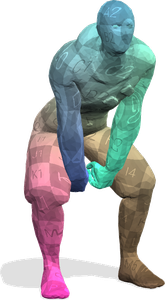}&
        \hspace{\hspaceCols}
        \includegraphics[height=\heightQ, width=\widthQ]{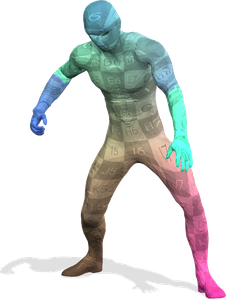}&
        \hspace{\hspaceCols}
        \includegraphics[height=\heightQ, width=\widthQ]{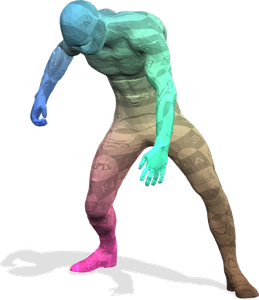}&
        \hspace{\hspaceCols}
        \includegraphics[height=\heightQ, width=\widthQ]{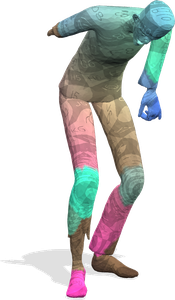}&
        \hspace{\hspaceCols}
        \includegraphics[height=\heightQ, width=\widthQ]{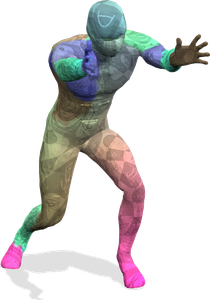}&
        \hspace{\hspaceCols}
        \includegraphics[height=\heightQ, width=\widthQ]{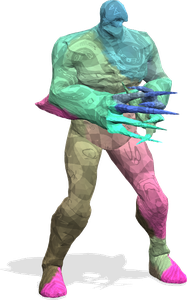}&
        \hspace{\hspaceCols}
        \includegraphics[height=\heightQ, width=\widthQ]{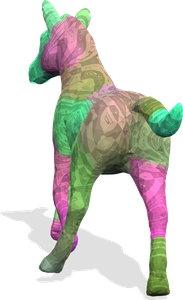}&
        \hspace{\hspaceCols}
        \includegraphics[height=\heightQ, width=\widthQ]{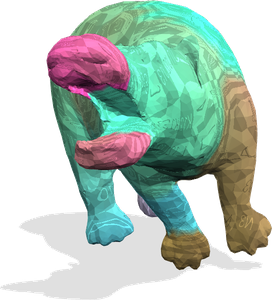}&
        \hspace{\hspaceCols}
        \includegraphics[height=\heightQ, width=\widthQ]{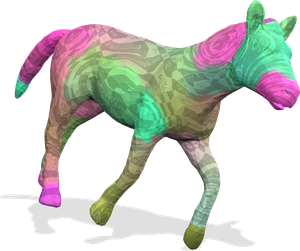}&
        \hspace{\hspaceCols}
        \includegraphics[height=\heightQ, width=\widthQ]{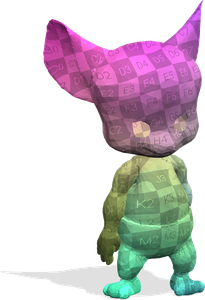}\\
        \rotatedCentering{90}{\heightQ}{AttentiveFMaps}&
        \hspace{\hspaceCols}
        \includegraphics[height=\heightQ, width=\widthQ]{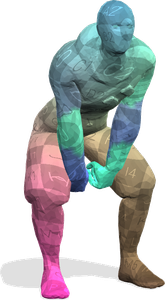}&
        \hspace{\hspaceCols}
        \includegraphics[height=\heightQ, width=\widthQ]{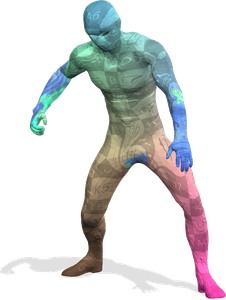}&
        \hspace{\hspaceCols}
        \includegraphics[height=\heightQ, width=\widthQ]{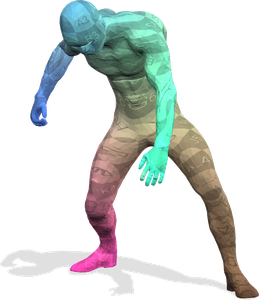}&
        \hspace{\hspaceCols}
        \includegraphics[height=\heightQ, width=\widthQ]{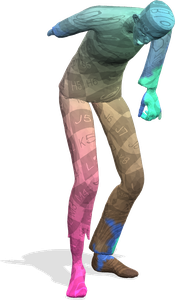}&
        \hspace{\hspaceCols}
        \includegraphics[height=\heightQ, width=\widthQ]{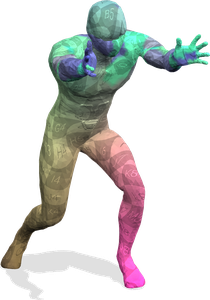}&
        \hspace{\hspaceCols}
        \includegraphics[height=\heightQ, width=\widthQ]{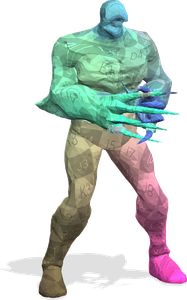}&
        \hspace{\hspaceCols}
        \includegraphics[height=\heightQ, width=\widthQ]{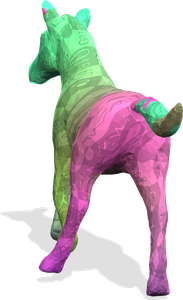}&
        \hspace{\hspaceCols}
        \includegraphics[height=\heightQ, width=\widthQ]{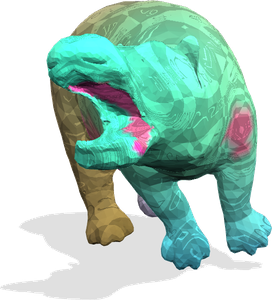}&
        \hspace{\hspaceCols}
        \includegraphics[height=\heightQ, width=\widthQ]{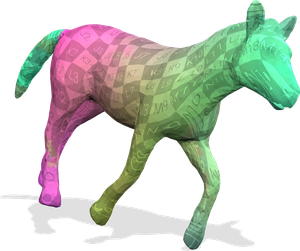}&
        \hspace{\hspaceCols}
        \includegraphics[height=\heightQ, width=\widthQ]{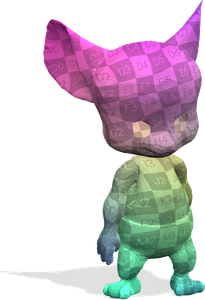}\\
        \rotatedCentering{90}{\heightQ}{AttentiveFMaps-Fast}&
        \hspace{\hspaceCols}
        \includegraphics[height=\heightQ, width=\widthQ]{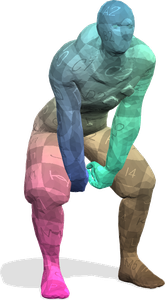}&
        \hspace{\hspaceCols}
        \includegraphics[height=\heightQ, width=\widthQ]{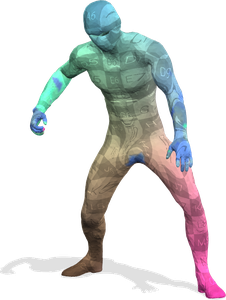}&
        \hspace{\hspaceCols}
        \includegraphics[height=\heightQ, width=\widthQ]{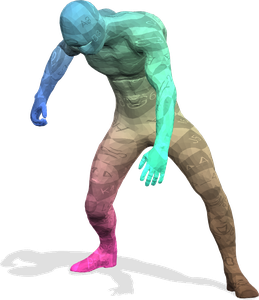}&
        \hspace{\hspaceCols}
        \includegraphics[height=\heightQ, width=\widthQ]{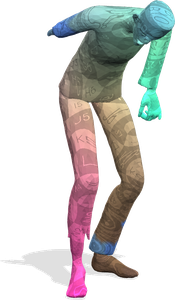}&
        \hspace{\hspaceCols}
        \includegraphics[height=\heightQ, width=\widthQ]{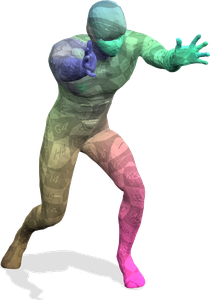}&
        \hspace{\hspaceCols}
        \includegraphics[height=\heightQ, width=\widthQ]{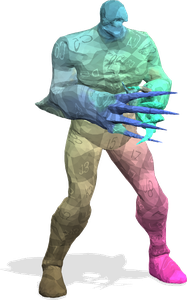}&
        \hspace{\hspaceCols}
        \includegraphics[height=\heightQ, width=\widthQ]{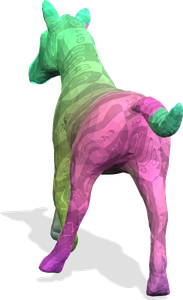}&
        \hspace{\hspaceCols}
        \includegraphics[height=\heightQ, width=\widthQ]{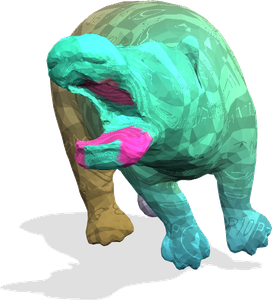}&
        \hspace{\hspaceCols}
        \includegraphics[height=\heightQ, width=\widthQ]{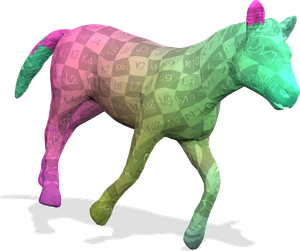}&
        \hspace{\hspaceCols}
        \includegraphics[height=\heightQ, width=\widthQ]{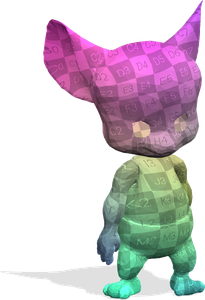}\\
        \rotatedCentering{90}{\heightQ}{GeomFMaps}&
        \hspace{\hspaceCols}
        \includegraphics[height=\heightQ, width=\widthQ]{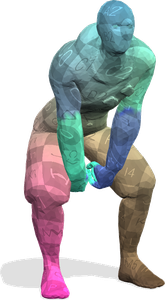}&
        \hspace{\hspaceCols}
        \includegraphics[height=\heightQ, width=\widthQ]{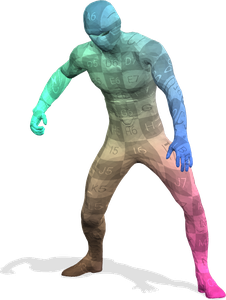}&
        \hspace{\hspaceCols}
        \includegraphics[height=\heightQ, width=\widthQ]{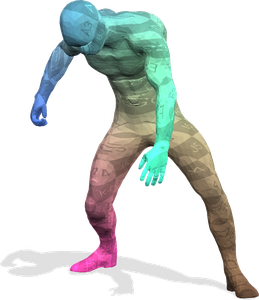}&
        \hspace{\hspaceCols}
        \includegraphics[height=\heightQ, width=\widthQ]{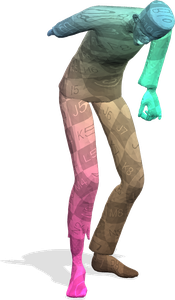}&
        \hspace{\hspaceCols}
        \includegraphics[height=\heightQ, width=\widthQ]{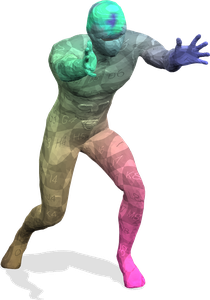}&
        \hspace{\hspaceCols}
        \includegraphics[height=\heightQ, width=\widthQ]{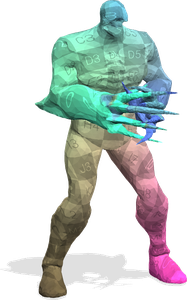}&
        \hspace{\hspaceCols}
        \includegraphics[height=\heightQ, width=\widthQ]{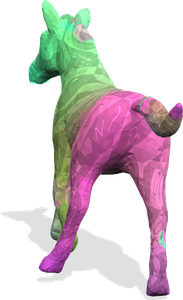}&
        \hspace{\hspaceCols}
        \includegraphics[height=\heightQ, width=\widthQ]{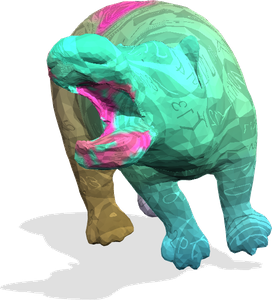}&
        \hspace{\hspaceCols}
        \includegraphics[height=\heightQ, width=\widthQ]{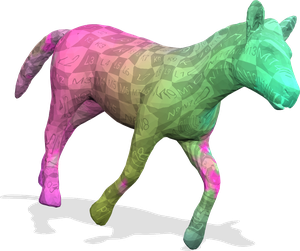}&
        \hspace{\hspaceCols}
        \includegraphics[height=\heightQ, width=\widthQ]{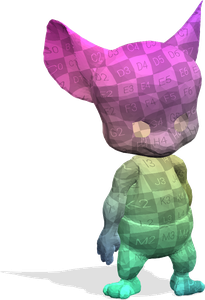}\\
        \rotatedCentering{90}{\heightQ}{Ours}&
        \hspace{\hspaceCols}
        \includegraphics[height=\heightQ, width=\widthQ]{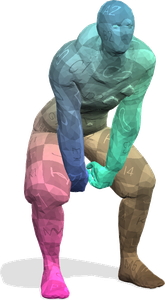}&
        \hspace{\hspaceCols}
        \includegraphics[height=\heightQ, width=\widthQ]{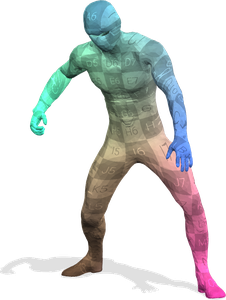}&
        \hspace{\hspaceCols}
        \includegraphics[height=\heightQ, width=\widthQ]{\pathOurs\columnThree\trgtEnd}&
        \hspace{\hspaceCols}
        \includegraphics[height=\heightQ, width=\widthQ]{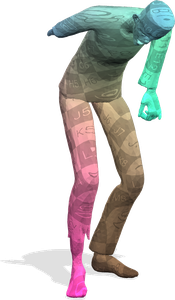}&
        \hspace{\hspaceCols}
        \includegraphics[height=\heightQ, width=\widthQ]{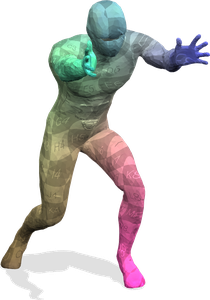}&
        \hspace{\hspaceCols}
        \includegraphics[height=\heightQ, width=\widthQ]{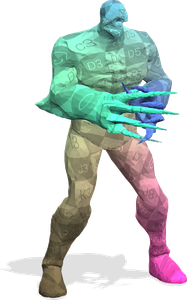}&
        \hspace{\hspaceCols}
        \includegraphics[height=\heightQ, width=\widthQ]{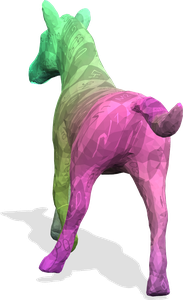}&
        \hspace{\hspaceCols}
        \includegraphics[height=\heightQ, width=\widthQ]{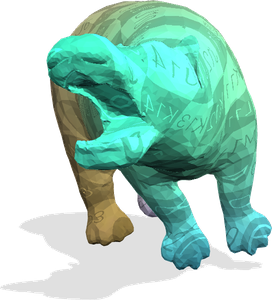}&
        \hspace{\hspaceCols}
        \includegraphics[height=\heightQ, width=\widthQ]{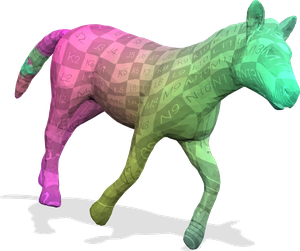}&
        \hspace{\hspaceCols}
        \includegraphics[height=\heightQ, width=\widthQ]{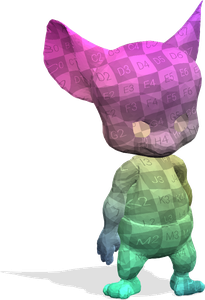}\\
    \end{tabular}
    \caption{\textbf{Non-isometric matching on SMAL and DT4D-H.} Comparing to existing methods, our approach demonstrates superior matching performance for both isometric and non-isometric shapes. For example, for shape matching between the lion and the hippo (the third column from the right), our method is the only one that provides smooth and accurate correspondences, while all competitors match the tail of the lion wrongly to the mouth of the hippo.}
    \label{fig:qualitative}
\end{figure*}

\textbf{Results.} Tab.~\ref{tab:non-isometry} summarises the matching results on the SMAL and DT4D-H datasets. Our approach outperforms the existing state-of-the-art on the challenging SMAL dataset even in comparison to supervised methods. Meanwhile, our method demonstrates near-perfect matching results for intra-class matching on the DT4D-H dataset. In the context of inter-class matching, our method outperforms existing axiomatic and unsupervised methods by a large margin and shows comparable matching performance for inter-class matching compared to the state-of-the-art supervised method.
{In Fig.~\ref{fig:pck_nonisoright_topkidsmiddle_partialright}~(left), we summarise the PCK curves of our method in comparison to AttentiveFMaps~\cite{li2022learning} on the SMAL and DT4D-H datasets.} Fig.~\ref{fig:qualitative} provides some qualitative results of our method in comparison to recent state-of-the-art methods on both SMAL and DT4D-H datasets. We observe that our method consistently outperforms existing approaches even in comparison to supervised methods. To further demonstrate the superior performance of our method, we also consider the highly non-isometric SHREC'20 dataset~\cite{dyke2020track}, which contains animals with extremely different appearance (e.g. giraffe and dog).
{Since SHREC'20 only contains 10 shapes, we directly use the training data for qualitative evaluation.} The qualitative results are shown in Fig.~\ref{fig:non-isometric-qualitative}. We observe that our method is capable of obtaining smooth and accurate matching results even in the presence of large non-isometric deformations.

\subsection{Partial Shape Matching}
\textbf{Datasets.} We evaluate our method in the context of partial shape matching on the challenging SHREC’16 partial dataset~\cite{cosmo2016shrec}. The dataset contains 200 training shapes, categorised into 8 classes (humans and animals). Each class has a complete shape to which the other partial shapes are matched. The dataset is divided into two subsets, namely CUTS (missing a large part) with 120 pairs, and HOLES (missing many small parts) with 80 pairs.
{Following \citet{attaiki2021dpfm}, we train our method for each subset individually and evaluate it on the corresponding test set (200 shapes for each subset).}

\textbf{Baseline.} Only a few existing shape matching approaches can be applied to partial shape matching. We categorise them as follows:
\begin{enumerate}
    \item \textbf{Axiomatic approaches}, including PFM~\cite{rodola2017partial}, FSP~\cite{litany2017fully};
    \item \textbf{Supervised approaches}, including GeomFMaps~\cite{donati2020deep}, DPFM~\cite{attaiki2021dpfm}; and
    \item \textbf{Unsupervised approaches}, including DPFM-unsup~\cite{attaiki2021dpfm}, ConsistFMaps~\cite{cao2022unsupervised}.
\end{enumerate}

\begin{table}[h!]
    \setlength{\tabcolsep}{3pt}
    \small
    \centering
    \caption{\textbf{Partial shape matching on  SHREC'16.} The numbers in parentheses show refined results using the indicated post-processing technique.  Our method is the first unsupervised approach that bridges the huge performance gap between supervised and unsupervised methods.}
    \label{tab:partial}
        \begin{tabular}{@{}lcccc@{}}
        \toprule
        \multicolumn{1}{l}{Train}  & \multicolumn{2}{c}{\textbf{CUTS}} & \multicolumn{2}{c}{\textbf{HOLES}}\\ \cmidrule(lr){2-3} \cmidrule(lr){4-5}
        \multicolumn{1}{l}{Test} & \multicolumn{1}{c}{\textbf{CUTS}} & \multicolumn{1}{c}{\textbf{HOLES}} & \multicolumn{1}{c}{\textbf{CUTS}} & \multicolumn{1}{c}{\textbf{HOLES}}
        \\ \midrule
        \multicolumn{5}{c}{Axiomatic Methods} \\
        \multicolumn{1}{l}{PFM (+\textit{zoomout})}  & \multicolumn{1}{c}{9.7 (9.0)} & \multicolumn{1}{c}{23.2 (22.4)} & \multicolumn{1}{c}{9.7 (9.0)} & \multicolumn{1}{c}{23.2 (22.4)} \\
        \multicolumn{1}{l}{FSP (+\textit{zoomout})}  & \multicolumn{1}{c}{16.1 (15.2)} & \multicolumn{1}{c}{33.7 (32.7)}  & \multicolumn{1}{c}{16.1 (15.2)} & \multicolumn{1}{c}{33.7 (32.7)} \\
        \midrule
        \multicolumn{5}{c}{Supervised Methods} \\ 
        \multicolumn{1}{l}{GeomFMaps (+\textit{zoomout})}  & \multicolumn{1}{c}{12.8 (10.4)} & \multicolumn{1}{c}{20.6 (17.4)} & \multicolumn{1}{c}{19.8 (16.7)} & \multicolumn{1}{c}{15.3 (13.0)}\\
        \multicolumn{1}{l}{DPFM (+\textit{zoomout})}  & \multicolumn{1}{c}{3.2 (\textbf{1.8})} & \multicolumn{1}{c}{15.8 (13.9)} & \multicolumn{1}{c}{8.6 (7.4)} & \multicolumn{1}{c}{13.1 (11.9)}\\
        \midrule
        \multicolumn{5}{c}{Unsupervised Methods} \\
        \multicolumn{1}{l}{DPFM-unsup (+\textit{zoomout})}  & \multicolumn{1}{c}{9.0 (7.8)} & \multicolumn{1}{c}{22.8 (20.0)} & \multicolumn{1}{c}{16.5 (14.6)} & \multicolumn{1}{c}{20.5 (18.4)} \\
        \multicolumn{1}{l}{ConsistFMaps}  & \multicolumn{1}{c}{8.4} & \multicolumn{1}{c}{23.7} & \multicolumn{1}{c}{15.7} & \multicolumn{1}{c}{17.9} \\
        \multicolumn{1}{l}{Ours}  & \multicolumn{1}{c}{3.3} & \multicolumn{1}{c}{\textbf{13.7}}  & \multicolumn{1}{c}{\textbf{5.2}}  & \multicolumn{1}{c}{\textbf{9.1}} \\
        \hline
    \end{tabular}
\end{table}

\textbf{Results.} Tab.~\ref{tab:partial} summarises the partial shape matching results on the SHREC'16 dataset. Our approach outperforms existing axiomatic and unsupervised methods by a large margin without relying on any post-processing techniques. In comparison to the supervised DPFM, our method achieves better performance on the more challenging HOLES subset. {In the context of generalisation ability, our method substantially outperforms existing methods even in comparison to the supervised DPFM.}
{Fig.~\ref{fig:pck_nonisoright_topkidsmiddle_partialright}~(right) summarises the PCK curves on the SHREC'16 partial datasets.}
Fig.~\ref{fig:partial} shows qualitative results on the SHREC'16 partial shape datasets of our method in comparison to other learning-based approaches. Compared to other learning-based approaches, we observe that our approach is capable of obtaining more accurate and smoother correspondences, even for partial shapes with several large missing parts. Overall, our method is the first unsupervised approach that bridges the huge performance gap between supervised and unsupervised methods.

\begin{figure}[h!]
    \centering
    \footnotesize
    \def\columnOneP{cat-cuts_cat_shape_17}
\def\columnTwoP{michael-cuts_michael_shape_10}
\def\columnThreeP{horse-holes_horse_shape_12}
\def\columnFourP{cat-holes_cat_shape_24}
\def\columnFiveP{victoria-cuts_victoria_shape_15}
\def\heightP{2cm}
\def\widthP{1.8cm}
\def\hspaceColsP{-0.44cm}
\begin{tabular}{lccccc}%
    \setlength{\tabcolsep}{0pt} 
    \rotatedCentering{90}{\heightP}{Source}&
    \hspace{\hspaceColsP}
    \includegraphics[height=\heightP, width=\widthP]{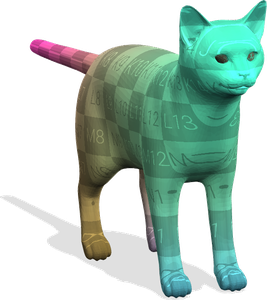}&
    \hspace{\hspaceColsP}
    \includegraphics[height=\heightP, width=\widthP]{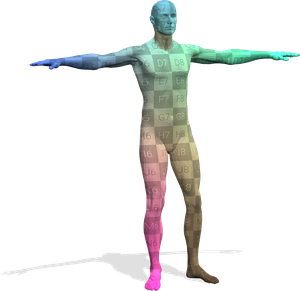}&
    \hspace{\hspaceColsP}
    \includegraphics[height=\heightP, width=\widthP]{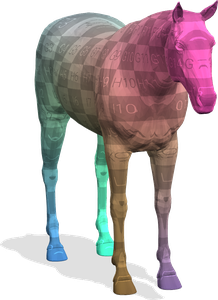}&
    \hspace{\hspaceColsP}
    \includegraphics[height=\heightP, width=\widthP]{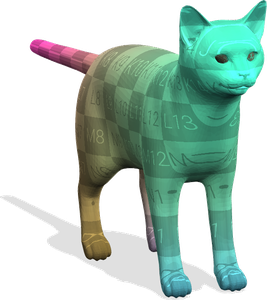}&
    \hspace{\hspaceColsP}
    \includegraphics[height=\heightP, width=\widthP]{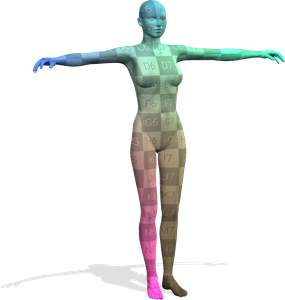}\\
    \rotatedCentering{90}{\heightP}{DPFM}&
    \hspace{\hspaceColsP}
    \includegraphics[height=\heightP, width=\widthP]{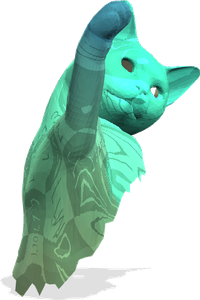}&
    \hspace{\hspaceColsP}
    \includegraphics[height=\heightP, width=\widthP]{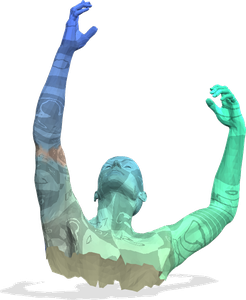}&
    \hspace{\hspaceColsP}
    \includegraphics[height=\heightP, width=\widthP]{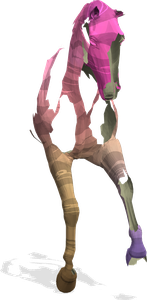}&
    \hspace{\hspaceColsP}
    \includegraphics[height=\heightP, width=\widthP]{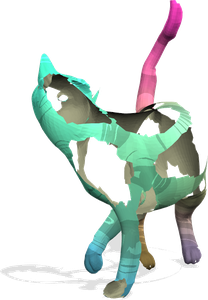}&
    \hspace{\hspaceColsP}
    \includegraphics[height=\heightP, width=\widthP]{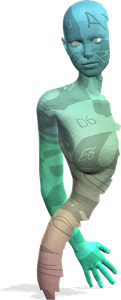}\\
    \rotatedCentering{90}{\heightP}{DPFM-unsup}&
    \hspace{\hspaceColsP}
    \includegraphics[height=\heightP, width=\widthP]{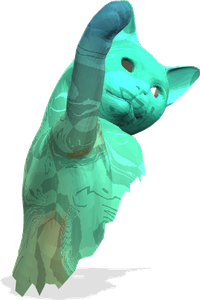}&
    \hspace{\hspaceColsP}
    \includegraphics[height=\heightP, width=\widthP]{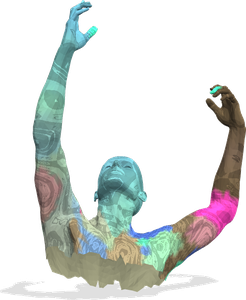}&
    \hspace{\hspaceColsP}
    \includegraphics[height=\heightP, width=\widthP]{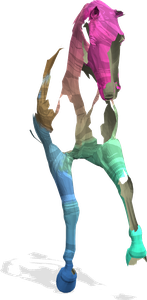}&
    \hspace{\hspaceColsP}
    \includegraphics[height=\heightP, width=\widthP]{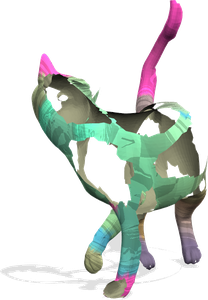}&
    \hspace{\hspaceColsP}
    \includegraphics[height=\heightP, width=\widthP]{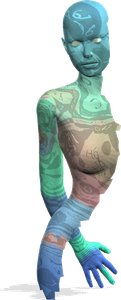}\\
    \rotatedCentering{90}{\heightP}{GeomFMaps}&
    \hspace{\hspaceColsP}
    \includegraphics[height=\heightP, width=\widthP]{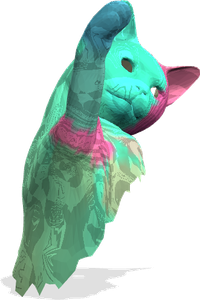}&
    \hspace{\hspaceColsP}
    \includegraphics[height=\heightP, width=\widthP]{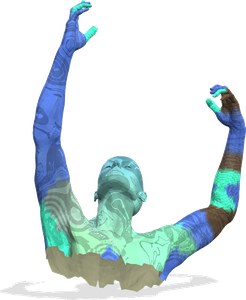}&
    \hspace{\hspaceColsP}
    \includegraphics[height=\heightP, width=\widthP]{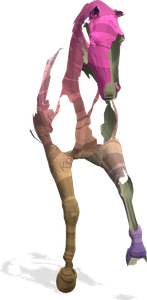}&
    \hspace{\hspaceColsP}
    \includegraphics[height=\heightP, width=\widthP]{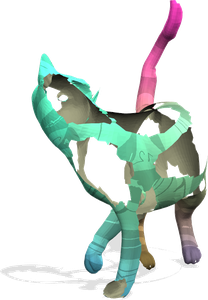}&
    \hspace{\hspaceColsP}
    \includegraphics[height=\heightP, width=\widthP]{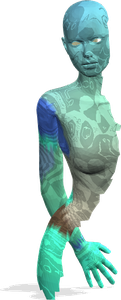}\\
    \rotatedCentering{90}{\heightP}{Ours}&
    \hspace{\hspaceColsP}
    \includegraphics[height=\heightP, width=\widthP]{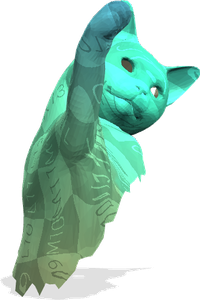}&
    \hspace{\hspaceColsP}
    \includegraphics[height=\heightP, width=\widthP]{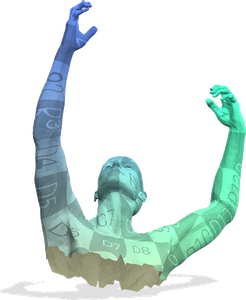}&
    \hspace{\hspaceColsP}
    \includegraphics[height=\heightP, width=\widthP]{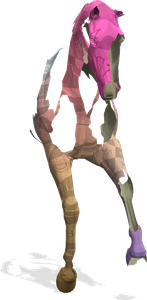}&
    \hspace{\hspaceColsP}
    \includegraphics[height=\heightP, width=\widthP]{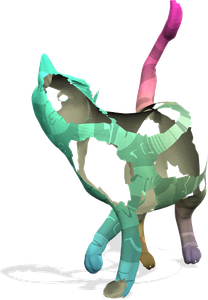}&
    \hspace{\hspaceColsP}
    \includegraphics[height=\heightP, width=\widthP]{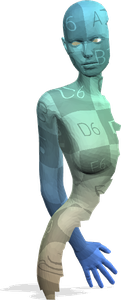}\\
\end{tabular}
    \caption{\textbf{Partial matching on SHREC'16.} Our method achieves near-perfect matching performanc, even for shapes with several large missing parts, and outperforms existing (supervised and unsupervised) methods.}
    \label{fig:partial}
\end{figure}

\subsection{Runtime Comparison}
We evaluate the runtime of our method and compare it to the state-of-the-art supervised method GeomFMaps \cite{donati2020deep} (with and without ZoomOut~\cite{melzi2019zoomout}).
In Fig.~\ref{fig:runtime} (left) we show the runtime for shapes with a different number of vertices ranging from 1k to 20k
when using the default setting of the number of eigenfunctions for our method ($k=200$) and GeomFMaps ($k=30$), respectively.
We observe that our method is slower compared to GeomFMaps for shapes with a relatively small number of vertices.
Nevertheless, when increasing the shape resolution, 
the runtime difference between GeomFmaps and our method is decreasing.
In Fig.~\ref{fig:runtime} (middle) we show the runtime for a different number of eigenfunctions ranging from 30 to 200.
Here, we fix the number of vertices of the input shapes to 10k.
We observe that our approach requires less computational time compared to GeomFMaps when the number of eigenfunctions is the same for both methods.
The main reason is that our method does not compute functional maps (i.e. solving Eq.~\ref{eq:fmap}) or rely on any off-the-shelf post-processing techniques during inference.
Overall, we consider matching a pair of high-resolution shapes within a few seconds as reasonable, especially in light of the remarkable performance of our method.
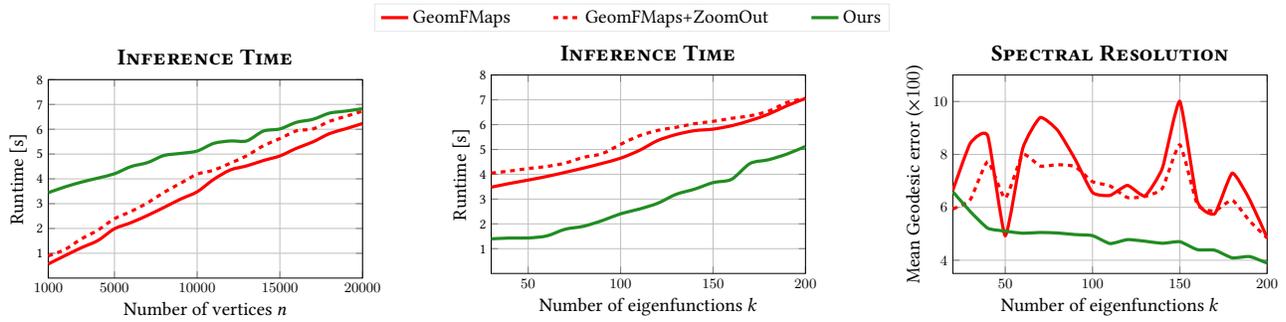
\begin{figure*}[ht]
    \centering
    \begin{tabular}{ccc}
    \multicolumn{3}{c}{\begin{tikzpicture}[scale=0.6, transform shape]
    \begin{axis}[%
    hide axis,
    xmin=10,
    xmax=11,
    ymin=0,
    ymax=0.4,
    legend style={%
        label style = {font=\LARGE},
        title style =  {},
        legend columns = 3,
        legend style={  %
                        fill opacity=0.6, 
                        font=\LARGE,
                        draw=gray!20, %
                        text opacity=1}
    }
    ]
    \addlegendimage{color=red, line width=2}
    \addlegendentry{GeomFMaps$\qquad$};
    \addlegendimage{color=red, dashed, line width=2}
    \addlegendentry{GeomFMaps+ZoomOut$\qquad$};
    \addlegendimage{color=ForestGreen, line width=2}
    \addlegendentry{Ours};
    
    \end{axis}
\end{tikzpicture}}\\
     \newcommand{\runtimeLineWidth}{2pt}
\newcommand{\rtplotWidth}{\columnwidth}
\newcommand{\rtplotHeight}{0.7\columnwidth}
\newcommand{\rtTextSize}{\LARGE}
\pgfplotsset{%
    label style = {font=\rtTextSize},
    title style =  {},
    legend style={  fill= gray!10,
                    fill opacity=0.6, 
                    font=\rtTextSize,
                    draw=gray!20, %
                    text opacity=1}
}
\begin{tikzpicture}[scale=0.6, transform shape]
    \begin{axis}[%
        width=\rtplotWidth,
        height=\rtplotHeight,
        title=\textsc{\textbf{{\huge Inference Time}}},
        grid=major,
        legend style={
                draw=none,
			at={(0.98,0.02)},
			anchor=south east,
			legend columns=3,
                fill= gray!10,
                fill opacity=0.6, 
                font=\rtTextSize,
                draw=gray!20, %
                text opacity=1
        },
        legend to name=void,
        ylabel={{\rtTextSize Runtime [s]}},
        xlabel={{\rtTextSize Number of vertices $n$}},
        xmin=999,
        xmax=20000,
        xtick={1000, 5000, 10000, 15000, 20000},
        xticklabels={{$1000$}, $5000$, {$10000$}, {$15000$}, {$20000$}},
        xtick scale label code/.code={},
        ymin=0,
        ymax=8,
        ytick={ 1, 2, 3, 4, 5, 6, 7, 8},
        yticklabels={ 1, 2, 3, 4, 5, 6, 7, 8},
        ylabel near ticks,
    ]
\addplot [color=red, smooth, line width=\runtimeLineWidth]
  table[row sep=crcr]{%
1000 0.56826\\
2000 0.90114\\
3000 1.2233\\
4000 1.5142\\
5000 1.9882\\
6000 2.2335\\
7000 2.5261\\
8000 2.8497\\
9000 3.1839\\
10000 3.4787\\
11000 3.9789\\
12000 4.3586\\
13000 4.5241\\
14000 4.7391\\
15000 4.9238\\
16000 5.2319\\
17000 5.4910\\
18000 5.8239\\
19000 6.0231\\
20000 6.2319\\
};
\addlegendentry{\textcolor{black}{GeomFMaps}}
\addplot [color=red, smooth, dashed, line width=\runtimeLineWidth]
  table[row sep=crcr]{%
1000 0.89353\\
2000 1.1502\\
3000 1.58\\
4000 1.9326\\
5000 2.3915\\
6000 2.6856\\
7000 3.0325\\
8000 3.459\\
9000 3.8232\\
10000 4.1888\\
11000 4.3489\\
12000 4.6312\\
13000 4.9324\\
14000 5.3341\\
15000 5.6238\\
16000 5.9353\\
17000 6.0132\\
18000 6.3201\\
19000 6.5231\\
20000 6.7319\\
};
\addlegendentry{\textcolor{black}{GeomFMaps+ZoomOut}}
\addplot [color=ForestGreen, line width=\runtimeLineWidth, smooth]
  table[row sep=crcr]{%
1000 3.444\\
2000 3.6738\\
3000 3.8686\\
4000 4.0312\\
5000 4.2055\\
6000 4.496\\
7000 4.6557\\
8000 4.9444\\
9000  5.0252\\
10000 5.1218\\
11000 5.4311\\
12000 5.5275\\
13000 5.5309\\
14000 5.9263\\
15000 6.0128\\
16000 6.2793\\
17000 6.4008\\
18000 6.6497\\
19000 6.7331\\
20000 6.8287\\
};
\addlegendentry{\textcolor{black}{Ours}}
\end{axis}
\end{tikzpicture}%
&
      \newcommand{\runtimeLineWidth}{2pt}
\newcommand{\rtplotWidth}{\columnwidth}
\newcommand{\rtplotHeight}{0.7\columnwidth}
\newcommand{\rtTextSize}{\LARGE}
\newcommand{\specTextsize}{\LARGE}
\pgfplotsset{%
    label style = {font=\rtTextSize},
    title style =  {},
    legend style={  fill= gray!10,
                    fill opacity=0.6, 
                    font=\rtTextSize,
                    draw=gray!20, %
                    text opacity=1}
}
\begin{tikzpicture}[scale=0.6, transform shape]
    \begin{axis}[%
        width=\rtplotWidth,
        height=\rtplotHeight,
        title=\textsc{\textbf{{\huge Inference Time}}},
        grid=major,
        legend style={
                draw=none,
			at={(0.98,0.02)},
			anchor=south east,
			legend columns=3,
                fill= gray!10,
                fill opacity=0.6, 
                font=\rtTextSize,
                draw=gray!20, %
                text opacity=1
        },
        legend to name=void,
        ylabel={{\rtTextSize Runtime [s]}},
        xlabel={{\specTextsize Number of eigenfunctions $k$}},
        xmin=30,
        xmax=200,
        xtick={50, 100, 150, 200},
        xticklabels={$50$, {$100$}, {$150$}, {$200$}},
        xtick scale label code/.code={},
        ymin=0,
        ymax=8,
        ytick={ 1, 2, 3, 4, 5, 6, 7, 8},
        yticklabels={ 1, 2, 3, 4, 5, 6, 7, 8},
        ylabel near ticks,
    ]
\addplot [color=red, smooth, line width=\runtimeLineWidth]
  table[row sep=crcr]{%
30 3.4787\\
40 3.6293\\
50 3.7655\\
60 3.9098\\
70 4.0769\\
80 4.2495\\
90 4.4379\\
100 4.6477\\
110 4.9554\\
120 5.3577\\
130 5.6014\\
140 5.7584\\
150 5.8180\\
160 5.9549\\
170 6.1633\\
180 6.4219\\
190 6.7624\\
200 7.0561\\
};
\addlegendentry{\textcolor{black}{GeomFMaps}}

\addplot [color=red, smooth, dashed, line width=\runtimeLineWidth]
  table[row sep=crcr]{%
30 4.0453\\
40 4.1293\\
50 4.2312\\
60 4.3122\\
70 4.4509\\
80 4.6828\\
90 4.8298\\
100 5.2063\\
110 5.5482\\
120 5.7677\\
130 5.8914\\
140 6.0384\\
150 6.1320\\
160 6.2549\\
170 6.3563\\
180 6.5423\\
190 6.8945\\
200 7.0561\\
};
\addlegendentry{\textcolor{black}{GeomFMaps+ZoomOut}}
\addplot [color=ForestGreen, line width=\runtimeLineWidth, smooth]
  table[row sep=crcr]{%
30 1.3953\\
40 1.4293\\
50 1.4312\\
60 1.5122\\
70 1.7831\\
80 1.9028\\
90 2.1298\\
100 2.4063\\
110 2.5942\\
120 2.8287\\
130 3.1914\\
140 3.4000\\
150 3.6630\\
160 3.7956\\
170 4.4339\\
180 4.5848\\
190 4.8147\\
200 5.1218\\
};
\addlegendentry{\textcolor{black}{Ours}}
\end{axis}
\end{tikzpicture}%
&
     \newcommand{\specResLineWidth}{2pt}
\newcommand{\plotWidth}{\columnwidth}
\newcommand{\plotHeight}{0.7\columnwidth}
\newcommand{\specResTitle}{Spectral Resolution}
\newcommand{\specTension}{0.3}
\newcommand{\specTextsize}{\LARGE}
\definecolor{othercolor}{rgb}{0.00000,0.44706,0.74118}
\definecolor{ourscolor}{rgb}{0.85098,0.32549,0.09804}

\pgfplotsset{%
    label style = {font=\specTextsize},
    title style =  {},
    legend style={  fill= gray!10,
                    fill opacity=0.6, 
                    font=\specTextsize,
                    draw=gray!20, %
                    text opacity=1}
}
\begin{tikzpicture}[scale=0.6, transform shape]
	\begin{axis}[
		width=\plotWidth,
		height=\plotHeight,
		grid=major,
		title=\textsc{\textbf{\huge\specResTitle}},
		legend style={
                draw=none,
			at={(0.97,0.97)},
			anchor=north east,
			legend columns=1},
            legend to name=void2,
		legend cell align={left},
		ylabel={{\specTextsize Mean Geodesic error ($\times 100$)}},
            xlabel={{\specTextsize Number of eigenfunctions $k$}},
	xmin=20,
        xmax=200,
        ylabel near ticks,
        xtick={50, 100, 150, 200},
	ymin=3.5,
        ymax=11,
        ytick={4, 6, 8, 10},
	]
	
	\addplot [color=red, smooth, tension=\specTension, line width=\specResLineWidth]
    table[row sep=crcr]{%
20 6.64\\
30 8.42\\
40 8.7\\
50 4.93\\
60 8.22\\
70 9.39\\
80 8.91\\
90 7.82\\
100 6.56\\
110 6.45\\
120 6.83\\
130 6.43\\
140 7.48\\
150 10\\
160 6.23\\
170 5.76\\
180 7.29\\
190 6.29\\
200 4.84\\
    };
    \addlegendentry{\textcolor{black}{GeomFMaps}}
    
    \addplot [color=red, dashed, smooth, tension=\specTension, line width=\specResLineWidth]
          table[row sep=crcr]{%
20 5.94\\
30 6.29\\
40 7.73\\
50 6.34\\
60 7.99\\
70 7.56\\
80 7.61\\
90 7.54\\
100 6.97\\
110 6.81\\
120 6.37\\
130 6.41\\
140 6.72\\
150 8.36\\
160 6.11\\
170 5.86\\
180 6.27\\
190 5.48\\
200 4.84\\
        };
        \addlegendentry{\textcolor{black}{GeomFMaps+ZoomOut}}

    \addplot [color=ForestGreen, smooth, tension=\specTension, line width=\specResLineWidth]
    table[row sep=crcr]{%
20 6.59\\
30 5.84\\
40 5.21\\
50 5.1\\
60 5.02\\
70 5.05\\
80 5.03\\
90 4.97\\
100 4.93\\
110 4.63\\
120 4.78\\
130 4.72\\
140 4.64\\
150 4.7\\
160 4.4\\
170 4.38\\
180 4.09\\
190 4.14\\
200 3.89\\
    };
    \addlegendentry{\textcolor{black}{Ours}}
        
	\end{axis}
\end{tikzpicture}
    \end{tabular}
    \caption{\textbf{Inference time and robustness to spectral resolution.} We compare our method to the state-of-the-art supervised method GeomFMaps~\cite{donati2020deep} (with and without ZoomOut~\cite{melzi2019zoomout}). \textbf{Left:} Runtime comparison with a different number of vertices. Compared to GeomFMaps, our method requires more computational time due to the choice of a larger number of eigenfunctions (200 versus 30). Nevertheless, for shapes with higher resolution, the runtime between GeomFMaps and our approach become comparable.
    \textbf{Middle:} Runtime comparison with a different number of eigenfunctions (while the number of vertices for shapes is fixed to 10k). Our method is faster than GeomFMaps when the number of eigenfunctions is the same for both methods.
    \textbf{Right:} We observe an unstable matching performance for GeomFMaps w.r.t. the spectral resolution. In contrast, our unsupervised method is more robust to the choice of the spectral resolution and consistently outperforms GeoFMaps.}
    \label{fig:runtime}
\end{figure*}

\begin{figure*}
    \centering
    \def\columnOne{tr_reg_083-tr_reg_086}
\def\columnTwo{tr_reg_082-tr_reg_080}
\def\columnThree{mesh065-mesh058}
\def\columnFour{mesh058-mesh065}
\def\columnFive{44-13}
\def\columnSix{dog-pig}
\def\columnSeven{wolf-cuts_wolf_shape_2}
\def\columnEight{wolf-holes_wolf_shape_2}
\def\columnNine{InvertedDoubleKickToKipUp216-InvertedDoubleKickToKipUp244}
\def\columnTen{DancingRunningMan305-StandingReactLargeFromLeft001}
\def\columnEleven{hippo_01-hippo_06}
\def\columnTwelve{kid00-kid05}
\def\columnOneW{tr_reg_095-tr_reg_087}
\def\columnTwoW{tr_reg_094-tr_reg_096}
\def\columnThreeW{mesh070-mesh059}
\def\columnFourW{mesh059-mesh070}
\def\columnFiveW{10-32}
\def\columnSixW{dog-rhino}
\def\columnSevenW{victoria-cuts_victoria_shape_26}
\def\columnEightW{david-holes_david_shape_18}
\def\columnNineW{Flair027-Flair013}
\def\columnTenW{Standing2HMagicAttack01043-Strafing000}
\def\columnElevenW{hippo_05-horse_05}
\def\columnTwelveW{kid00-kid17}
\def\pathBW{figures/visBestworst/}
\def\heightBW{2cm}
\def\widthBW{1.5cm}
\def\hspaceCols{-0.35cm}
\def\hspaceTextCols{-0.4cm}
\def\datasetTextSize{\scriptsize}
\begin{tabular}{lcccccccccccc}%
        \setlength{\tabcolsep}{0pt} 
        &\hspace{\hspaceTextCols}{\datasetTextSize FAUST}
        &\hspace{\hspaceTextCols}{\datasetTextSize FAUST\_a}
        &\hspace{\hspaceTextCols}{\datasetTextSize SCAPE}
        &\hspace{\hspaceTextCols}{\datasetTextSize SCAPE\_a}
        &\hspace{\hspaceTextCols}{\datasetTextSize SHREC'19}
        &\hspace{\hspaceTextCols}{\datasetTextSize SHREC'20}
        &\hspace{\hspaceTextCols}{\datasetTextSize SHREC'16CUTS}
        &\hspace{\hspaceTextCols}{\datasetTextSize SHREC'16HOLES}
        &\hspace{\hspaceTextCols}{\datasetTextSize DT4D-H intra}
        &\hspace{\hspaceTextCols}{\datasetTextSize DT4D-H inter}
        &\hspace{\hspaceTextCols}{\datasetTextSize SMAL}
        &\hspace{\hspaceTextCols}{\datasetTextSize TOPKIDS}\\
        \rotatebox{90}{\hspace{-0.35cm}{\footnotesize Best}}
        \rotatebox{90}{\hspace{-2cm}\rule{3.8cm}{0.5pt}}&
        \hspace{\hspaceCols}
        \includegraphics[height=\heightBW, width=\widthBW]{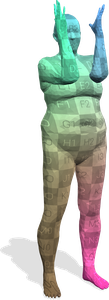}&
        \hspace{\hspaceCols}
        \includegraphics[height=\heightBW, width=\widthBW]{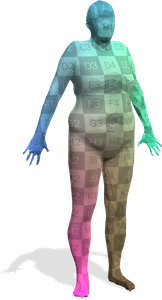}&
        \hspace{\hspaceCols}
        \includegraphics[height=\heightBW, width=\widthBW]{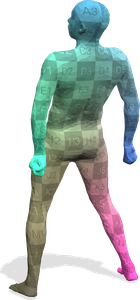}&
        \hspace{\hspaceCols}
        \includegraphics[height=\heightBW, width=\widthBW]{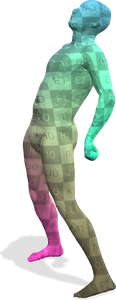}&
        \hspace{\hspaceCols}
        \includegraphics[height=\heightBW, width=\widthBW]{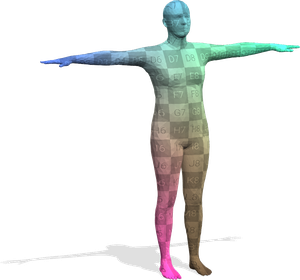}&
        \hspace{\hspaceCols}
        \includegraphics[height=\heightBW, width=\widthBW]{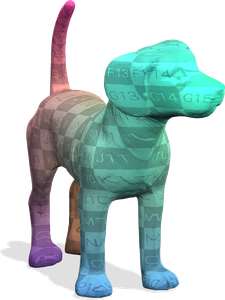}&
        \hspace{\hspaceCols}
        \includegraphics[height=\heightBW, width=\widthBW]{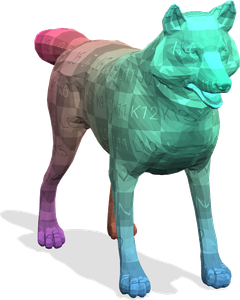}&
        \hspace{\hspaceCols}
        \includegraphics[height=\heightBW, width=\widthBW]{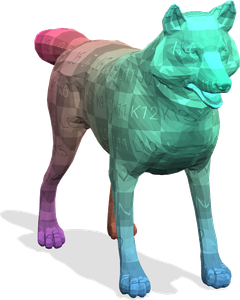}&
        \hspace{\hspaceCols}
        \includegraphics[height=\heightBW, width=\widthBW]{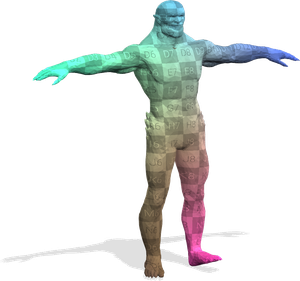}&
        \hspace{\hspaceCols}
        \includegraphics[height=\heightBW, width=\widthBW]{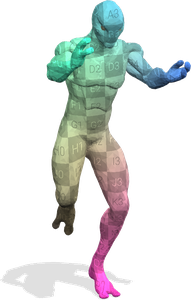}&
        \hspace{\hspaceCols}
        \includegraphics[height=\heightBW, width=\widthBW]{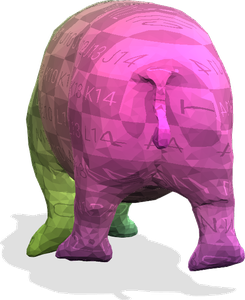}&
        \hspace{\hspaceCols}
        \includegraphics[height=\heightBW, width=\widthBW]{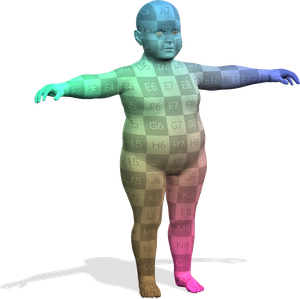}\\
        &
        \hspace{\hspaceCols}
        \includegraphics[height=\heightBW, width=\widthBW]{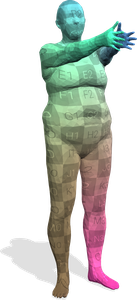}&
        \hspace{\hspaceCols}
        \includegraphics[height=\heightBW, width=\widthBW]{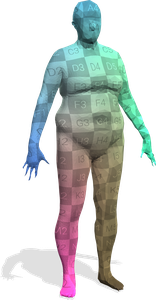}&
        \hspace{\hspaceCols}
        \includegraphics[height=\heightBW, width=\widthBW]{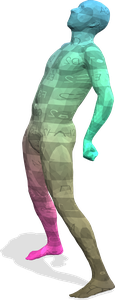}&
        \hspace{\hspaceCols}
        \includegraphics[height=\heightBW, width=\widthBW]{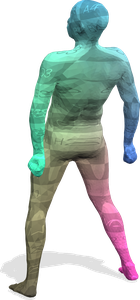}&
        \hspace{\hspaceCols}
        \includegraphics[height=\heightBW, width=\widthBW]{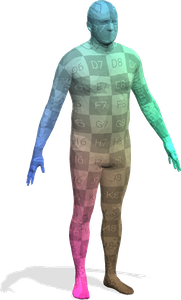}&
        \hspace{\hspaceCols}
        \includegraphics[height=\heightBW, width=\widthBW]{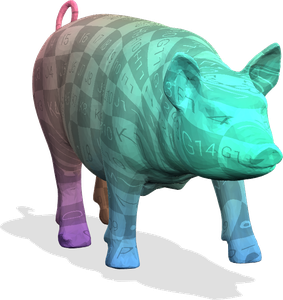}&
        \hspace{\hspaceCols}
        \includegraphics[height=\heightBW, width=\widthBW]{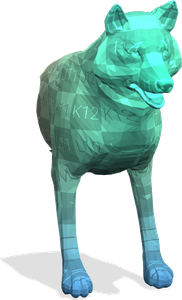}&
        \hspace{\hspaceCols}
        \includegraphics[height=\heightBW, width=\widthBW]{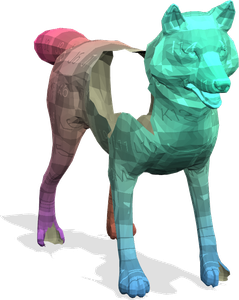}&
        \hspace{\hspaceCols}
        \includegraphics[height=\heightBW, width=\widthBW]{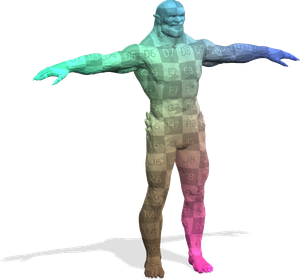}&
        \hspace{\hspaceCols}
        \includegraphics[height=\heightBW, width=\widthBW]{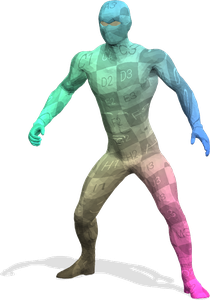}&
        \hspace{\hspaceCols}
        \includegraphics[height=\heightBW, width=\widthBW]{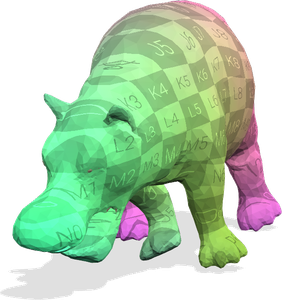}&
        \hspace{\hspaceCols}
        \includegraphics[height=\heightBW, width=\widthBW]{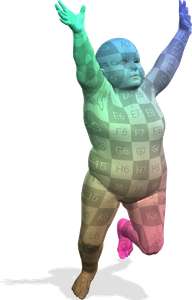}\\
        \rotatebox{90}{\hspace{-0.5cm}{\footnotesize Worst}}
        \rotatebox{90}{\hspace{-2cm}\rule{3.8cm}{0.5pt}}&
        \hspace{\hspaceCols}
        \includegraphics[height=\heightBW, width=\widthBW]{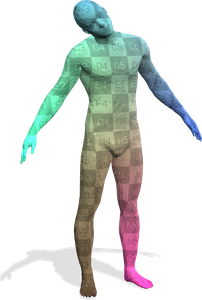}&
        \hspace{\hspaceCols}
        \includegraphics[height=\heightBW, width=\widthBW]{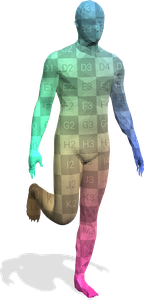}&
        \hspace{\hspaceCols}
        \includegraphics[height=\heightBW, width=\widthBW]{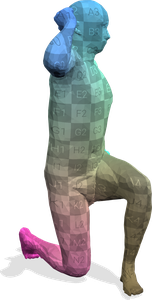}&
        \hspace{\hspaceCols}
        \includegraphics[height=\heightBW, width=\widthBW]{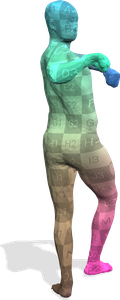}&
        \hspace{\hspaceCols}
        \includegraphics[height=\heightBW, width=\widthBW]{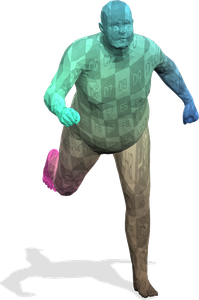}&
        \hspace{\hspaceCols}
        \includegraphics[height=\heightBW, width=\widthBW]{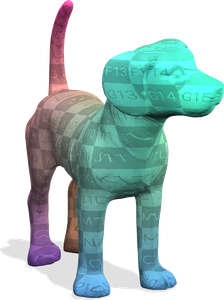}&
        \hspace{\hspaceCols}
        \includegraphics[height=\heightBW, width=\widthBW]{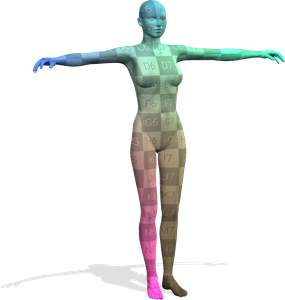}&
        \hspace{\hspaceCols}
        \includegraphics[height=\heightBW, width=\widthBW]{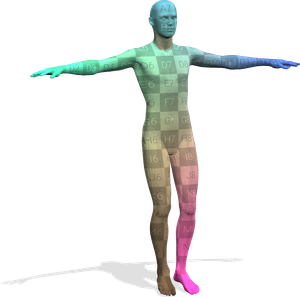}&
        \hspace{\hspaceCols}
        \includegraphics[height=\heightBW, width=\widthBW]{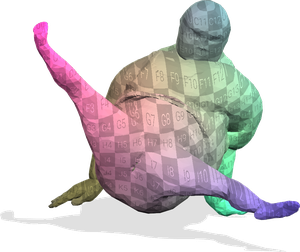}&
        \hspace{\hspaceCols}
        \includegraphics[height=\heightBW, width=\widthBW]{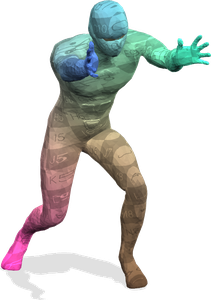}&
        \hspace{\hspaceCols}
        \includegraphics[height=\heightBW, width=\widthBW]{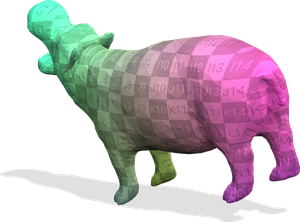}&
        \hspace{\hspaceCols}
        \includegraphics[height=\heightBW, width=\widthBW]{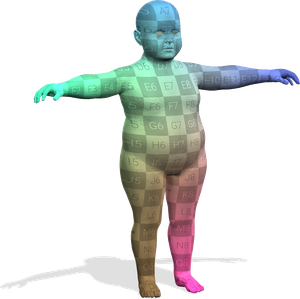}\\
        &
        \hspace{\hspaceCols}
        \includegraphics[height=\heightBW, width=\widthBW]{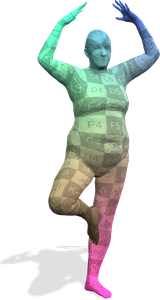}&
        \hspace{\hspaceCols}
        \includegraphics[height=\heightBW, width=\widthBW]{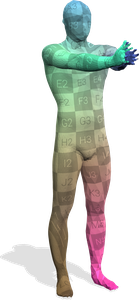}&
        \hspace{\hspaceCols}
        \includegraphics[height=\heightBW, width=\widthBW]{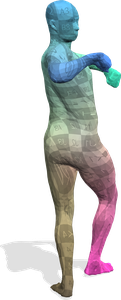}&
        \hspace{\hspaceCols}
        \includegraphics[height=\heightBW, width=\widthBW]{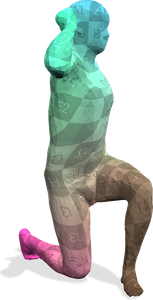}&
        \hspace{\hspaceCols}
        \includegraphics[height=\heightBW, width=\widthBW]{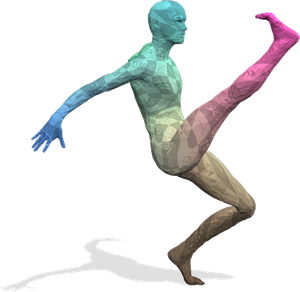}&
        \hspace{\hspaceCols}
        \includegraphics[height=\heightBW, width=\widthBW]{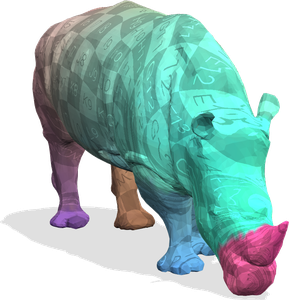}&
        \hspace{\hspaceCols}
        \includegraphics[height=\heightBW, width=\widthBW]{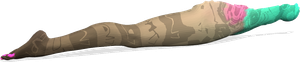}&
        \hspace{\hspaceCols}
        \includegraphics[height=\heightBW, width=\widthBW]{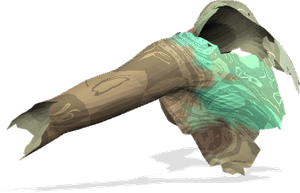}&
        \hspace{\hspaceCols}
        \includegraphics[height=\heightBW, width=\widthBW]{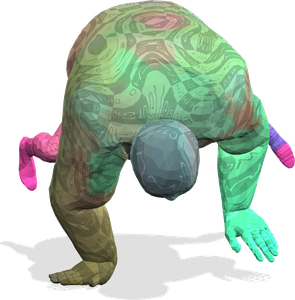}&
        \hspace{\hspaceCols}
        \includegraphics[height=\heightBW, width=\widthBW]{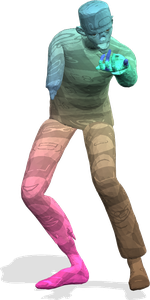}&
        \hspace{\hspaceCols}
        \includegraphics[height=\heightBW, width=\widthBW]{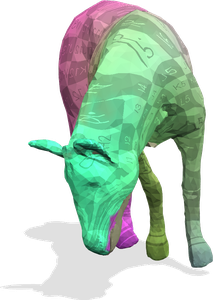}&
        \hspace{\hspaceCols}
        \includegraphics[height=\heightBW, width=\widthBW]{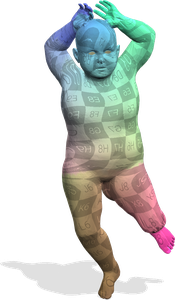}\\
    \end{tabular}
    \caption{{\textbf{Best and worst} pair on each of the evaluated datasets w.r.t. geodesic error of the matchings computed with our approach. We can see that the worst matchings still form reasonable results  in most cases, and that poor geodesic error scores originate from geometrically inconsistent matchings (e.g.~nose of the rhino in the sixth column), or left-right flips (e.g. kid in the last column).}
    }
    \label{fig:bestworst}
\end{figure*}

\section{Ablation Studies}
\label{sec:ablation}
In the following we perform ablative experiments to validate the merits of the individual components of our approach.

\subsection{Coupling Loss, Inference and Test-Time Adaptation}
\label{subsec:ablation}
We evaluate the importance of our introduced loss term  $L_{\mathrm{couple}}$ in Eq.~\ref{eq:couple}, as well as the proposed inference strategy and test-time adaptation described in Sec.~\ref{subsec:test-time-adaptation}. {Additionally, we investigate the advantage of using soft point-wise maps instead of hard ones. To this end, we consider  the Gumbel-trick~\cite{jang2016categorical} to obtain hard point-wise maps during training.} For all ablative experiments, we consider challenging non-isometric shape matching on the SMAL dataset.

\begin{table}[hbt!]
    \setlength{\tabcolsep}{16pt}
    \small
    \centering
    \caption{\textbf{Ablation study on SMAL.}  The first row shows the network trained only with $L_{\mathrm{fmap}}$ in Eq.~\ref{eq:fmaps}. In the second row we obtain the point-wise map based on the functional map computed by functional map solver. The third row represents inference without using test-time adaptation. The fourth row indicates not using $E_{\mathrm{d}}$ in Eq.~\ref{eq:dirichlet} during test-time adaptation. In the fifth row we obtain the hard point-wise maps by using Gumbel trick.}
    \label{tab:ablation}
    \begin{tabular}{@{}ll@{}}
    \toprule
    \multicolumn{1}{l}{\textbf{Ablation Setting}}  & \multicolumn{1}{l}{\textbf{SMAL}}
    \\ \midrule
    \multicolumn{1}{l}{w.o. $L_{\mathrm{couple}}$}  & 10.3 \\
    \multicolumn{1}{l}{w.o. our inference strategy}  & 5.8 \\
    \multicolumn{1}{l}{w.o. test-time adaptation}  & 5.5 \\
    \multicolumn{1}{l}{w.o. $L_{\mathrm{dirichlet}}$ in test-time adaptation}  & 4.3 \\
    \multicolumn{1}{l}{w.o. soft point-wise maps}  & 4.4 \\
    \multicolumn{1}{l}{Ours}  & \textbf{3.9} \\ \hline
    \end{tabular} 
\end{table}

\textbf{Results}. Tab.~\ref{tab:ablation} summarises the quantitative results. By comparing the first row and the last row, we can conclude that $L_{\mathrm{couple}}$ plays an important role for accurate matching. By comparing the second row and the last row, we observe that the point map based on deep feature similarity (last row)  is more accurate than converting the functional map computed by the functional map solver to a point-wise map (second row). Comparing the third and the last row shows that using the test-time adaptation leads to better matching performance. Comparing the fourth and the last row confirms that using the smoothness term $L_{\mathrm{dirichlet}}$ improves matching performance. By comparing the fifth row and the last row, we can conclude that soft point-wise maps provides more flexibility for shape matching, which in turn leads to better matching performance.

\subsection{Robustness Against Spectral Resolution}
Previous functional map methods are sensitive to the choice of spectral resolution. In contrast, we demonstrate that our method is more robust to the choice of the spectral resolution. For this experiment we consider the challenging non-isometric SMAL dataset and compare our method to the state-of-the-art supervised method GeomFMaps~\cite{donati2020deep}.

\textbf{Results.} Fig.~\ref{fig:runtime} (right) shows the  matching performance w.r.t. the choice of the spectral resolution. We observe that GeomFMaps is unstable during the increase of the spectral resolution. In contrast, our method is much more robust to the choice of the spectral resolution.

\subsection{{Our method as an axiomatic approach}}
\label{subsec:axiom}
{Since we observe that the test-time adaptation leads to better matching results (see Tab.~\ref{tab:ablation}), we experimentally analyse whether our method is capable of being used as an axiomatic method by optimising the feature extractor on a single pair of shapes individually. To this end, we compare the individual optimisation strategy (randomly initialised network weights for each shape pair) to our unsupervised training strategy for challenging non-isometric shape matching on the SMAL dataset.} 

\begin{table}[hbt!]
    \setlength{\tabcolsep}{16pt}
    \small
    \centering
    \caption{\textbf{Non-isometric matching on SMAL.} We compare our unsupervised training strategy to the individual optimisation strategy. We observe a large performance drop when using our method as an axiomatic approach.}
    \label{tab:axiom}
    \begin{tabular}{@{}ll@{}}
    \toprule
    \multicolumn{1}{l}{\textbf{Unsupervised vs Axiomatic}}  & \multicolumn{1}{l}{\textbf{SMAL}}
    \\ \midrule
    \multicolumn{1}{l}{Axiomatic (individual optimisation)}  & 43.1 \\
    \multicolumn{1}{l}{Ours (unsupervised training)}  & \textbf{3.9} \\ \hline
    \end{tabular} 
\end{table}

{\textbf{Results}. Tab.~\ref{tab:axiom} summarises the quantitative results, which shows that using our pipeline as an axiomatic method does not work well. We believe that this is because it lacks the collective regularisation from large datasets, so that the severe non-convexity of the optimisation problem may lead to poor local optima.} 
\section{Discussion and Limitations}
\label{sec:limitations}
Our proposed shape matching approach sets the new state of the art on a wide range of benchmark datasets. Yet, there are also some limitations that give rise to interesting future research questions. {Fig.~\ref{fig:bestworst} summarises the best and worst pair on each of the evaluated datasets w.r.t. geodesic error of the matchings computed with our method. We observe that most failure cases stem from geometrically inconsistent matchings and left-right flips.} Fig.~\ref{fig:failures} showcases some hand-picked failure modes of our method. While such cases occur rarely (see our complete results in the supplementary material), these examples confirm the difficulty of matching under extreme conditions, such as large non-isometries between an elephant and a giraffe.

\def\heightFailures{2.6cm}
\def\widthFailures{2.6cm}
\begin{figure}[h]
    \centering
    \begin{tabular}{
    ccc}
    \setlength{\tabcolsep}{0pt}
     \includegraphics[width=\widthFailures,height=\heightFailures]{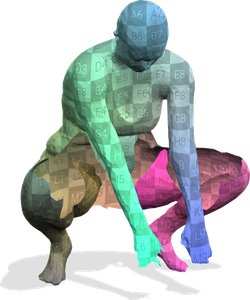} &
     \includegraphics[width=\widthFailures,height=\heightFailures]{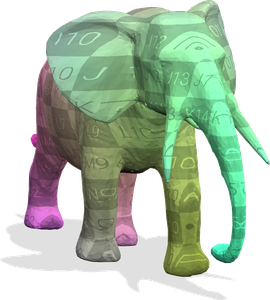} &
     \includegraphics[width=\widthFailures,height=\heightFailures]{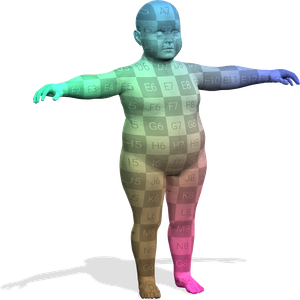}\\
     \includegraphics[width=\widthFailures,height=\heightFailures]{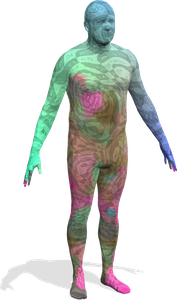}&
     \includegraphics[width=\widthFailures,height=\heightFailures]{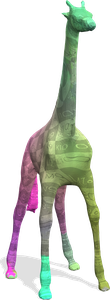}&
     \includegraphics[width=\widthFailures,height=\heightFailures]{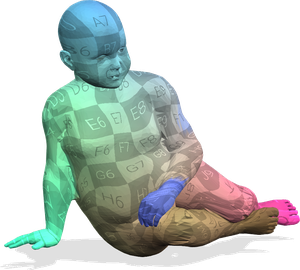}
    \end{tabular}
    \caption{\textbf{Examples of failure modes of our method.} \textbf{Left:} Training our method exclusively on complete shapes (FAUST and SCAPE) and then testing on partial data leads to matching failures (shape 40 of the SHREC'19 dataset, i.e.~the shape on the top, has a large missing part in the upper leg region).   \textbf{Middle:} Extreme non-isometries, such as between an elephant and a giraffe, may lead to erroneous matchings (SHREC'20). \textbf{Right:} Cases with severe topological noise may also lead to local mismatches (TOPKIDS).} 
    \label{fig:failures}
\end{figure}

{Our method is robust to the choice of spectral resolution, see Fig.~\ref{fig:runtime}.
Opposed to previous works for which this choice is critical, our robustness allows us to choose a relatively large number of eigenfunctions. With that, high-frequency features are adequately retained, so that in turn symmetric flips that commonly occur in competing methods (e.g.~of humans) can properly be resolved via the preserved geometric details (e.g.~of the face).
Yet, the number of used eigenfunctions still needs to be chosen carefully in order to balance matching performance and runtime.} Further improving the runtime of our method, as well as finding a reasonable trade-off between matching performance and runtime, are both interesting future directions.

In this work we only consider the case of pairwise shape matching and we leave the investigation of respective multi-shape matching formulations, e.g.~based on cycle-consistency constraints~\cite{huang2013consistent,bernard2019hippi,huang2020consistent,gao2021isometric,cao2022unsupervised}, for future work. {Furthermore, follow-up works might integrate the vertex-to-triangle matching formalism~\cite{ezuz2017deblurring} into our learning framework.} Another interesting direction for future work is a  generalisation of our approach to other types of shape representation, such as point clouds or neural shape representations.

\section{Conclusion}
\label{sec:conclusion}
In this work we propose the first unsupervised learning approach that provides a universal framework for shape matching under different challenging scenarios, including shapes with different discretisation, shapes with topological noise, highly non-isometric shapes, and partial shapes.
The favourable matching quality, robustness and cross-dataset generalisation ability of our method was enabled mainly due to the combination of three aspects: differentiable point-wise map computation, coupling point-wise maps with functional maps, and test-time adaptation. We have demonstrated that by combining these aspects, many previous limitations of functional maps can be addressed, including the sensitivity to the spectral resolution, the need for post-processing to obtain point-wise maps, and the unsatisfactory matching quality in the case of non-isometric shape deformations.
Overall, by bridging the gap between the theoretical advances of deep functional maps and their practical application in real-world settings, we believe that our method will be a valuable contribution to the field of computer graphics and beyond.

\begin{acks}

Funded by the TRA Modellling (University of Bonn) as part of the Excellence Strategy of the federal and state governments and by the Deutsche Forschungsgemeinschaft (DFG, German Research Foundation) – 458610525. This work was supported by the Visual Computing Incubator at the University of Bonn.

\end{acks}

\bibliographystyle{ACM-Reference-Format}
\bibliography{bibliography}

\end{document}